\documentclass[runningheads]{llncs}
\usepackage{hyperref}

\usepackage[dvipsnames]{xcolor}
\usepackage[small]{caption}
\usepackage[utf8]{inputenc} 
\usepackage[T1]{fontenc}    
\usepackage{url}            
\usepackage{booktabs}       
\usepackage{amsfonts}       
\usepackage{nicefrac}       
\usepackage{microtype}      

\usepackage{graphicx}
\usepackage{subfigure}
\usepackage{bm}
\usepackage{xspace}
\usepackage{siunitx} 

\usepackage{amsmath}
\usepackage{amssymb}
\usepackage{stackengine}
\usepackage{algorithm}
\usepackage{algpseudocode}
\usepackage[normalem]{ulem}
\usepackage{wrapfig}
\usepackage{tabularx}
\usepackage{shortcuts_js}

\usepackage{soul}

\newcommand{\name}{BOPrO\xspace}

\newcommand{\param}{\bm{x}}
\newcommand{\Param}{\mathcal{X}}
\newcommand{\priorgood}{P_g(\param)}
\newcommand{\priorbad}{P_b(\param)}
\newcommand{\modelgood}{\mathcal{M}_g(\param)}
\newcommand{\modelbad}{\mathcal{M}_b(\param)}

\newcommand{\topic}[1]{\noindent {\bf #1.}}

\newcommand{\HM}{HyperMapper\xspace}

\newcommand{\edit}[2]{#1\xspace}
\newcommand{\newedit}[2]{#1\xspace}
\newcommand{\postrebuttal}[1]{{\leavevmode\color{black}#1}\xspace}
\newcommand{\spatial}{\texttt{Spatial}\xspace}
\newcolumntype{L}{>{\raggedright\let\newline\\\arraybackslash\hspace{0pt}}X}

\newcommand{\posteriorstr}{pseudo-posterior\xspace}
\newcommand{\posteriorsstr}{pseudo-posteriors\xspace}
\newcommand{\priorstr}{prior\xspace}
\newcommand{\priorsstr}{priors\xspace}
\newcommand{\Priorstr}{Prior\xspace}
\newcommand{\Priorsstr}{Priors\xspace}
\newcommand{\history}{\newedit{\mathcal{D}_{t}}{}\xspace}
\allowdisplaybreaks 

\newcommand{\hide}[1]{}

\usepackage{verbatim}
\usepackage{comment}
\begin{document}
\title{Bayesian Optimization with \\a Prior for the Optimum}
%
%
\author{
Artur Souza\inst{1} \and
Luigi Nardi\inst{2,3} \and
Leonardo B. Oliveira\inst{1} \and \\
Kunle Olukotun\inst{3} \and
Marius Lindauer\inst{4} \and
Frank Hutter\inst{5,6}
}
\authorrunning{Souza \etal}
%
\institute{
Universidade Federal de Minas Gerais \\ \email{\{arturluis, leob\}@dcc.ufmg.br} \and
Lund University \\ \email{luigi.nardi@cs.lth.se} \and
Stanford University \\ \email{\{lnardi, kunle\}@stanford.edu} \and
Leibniz University Hannover \\ \email{lindauer@tnt.uni-hannover.de} \and
University of Freiburg \and
Bosch Center for Artificial Intelligence \\  \email{fh@cs.uni-freiburg.de}
}
\maketitle              
\begin{abstract}
    While Bayesian Optimization (BO) is a very popular method for optimizing expensive black-box functions, it fails to leverage the experience of domain experts. This causes BO to waste function evaluations on bad design choices (e.g., machine learning hyperparameters) that the expert already knows to work poorly. To address this issue, we introduce Bayesian Optimization with a Prior for the Optimum (BOPrO). BOPrO allows users to inject their knowledge into the optimization process in the form of priors about which parts of the input space will yield the best performance, rather than BO’s standard priors over functions, which are much less intuitive for users. BOPrO then combines these priors with BO’s standard probabilistic model to form a pseudo-posterior used to select which points to evaluate next. We show that BOPrO is around $6.67\times$ faster than state-of-the-art methods on a common suite of benchmarks, and achieves a new state-of-the-art performance on a real-world hardware design application. We also show that BOPrO converges faster even if the priors for the optimum are not entirely accurate and that it robustly recovers from misleading priors. 

\end{abstract}
\section{Introduction} \label{sec:intro}

Bayesian Optimization (BO) is a data-efficient method for the joint optimization of design choices
that has gained great popularity in recent years. 
It is impacting
a wide range of areas, including 
hyperparameter optimization~\cite{snoekLA12,falknerKH18}, %
AutoML~\cite{automl_book},  %
robotics~\cite{calandra2016bayesian},
computer vision~\cite{nardi2017algorithmic,bodin2016integrating}, 
Computer Go~\cite{chen_bo_for_alphago}, hardware design~\cite{nardi18hypermapper,koeplinger2018}, and many others. 
It promises greater automation so as to increase both product quality and human productivity. 
As a result, BO is also established in large tech companies, e.g., Google~\cite{google_vizier} and Facebook~\cite{balandat2020botorch}.

Nevertheless, domain experts often have substantial prior knowledge that standard BO cannot easily incorporate so far\newedit{~\cite{wang2019atmseer}}{}. Users can incorporate prior knowledge by narrowing the search space; however, this type of hard prior can lead to poor performance by missing important regions.
BO also supports a prior over functions $p(f)$, e.g., via a kernel function. However, this is not the prior domain experts have: they often know which ranges of hyperparameters tend to work best~\cite{perrone2019learning}, and are able to specify a probability distribution $p_{\text{best}}(\param)$ to quantify these priors; e.g., many users of the Adam optimizer~\cite{kingma_adam} know that its best learning rate is often in the vicinity of \edit{$1\times10^{-3}$}{} (give or take an order of magnitude), yet one may not know the accuracy one may achieve in a new application.
Similarly, Clarke \etal~\cite{clarke20fastai} derived neural network hyperparameter priors for image datasets based on their experience with five datasets. In these cases, users know potentially good values for a new application, but cannot be certain about them.

As a result, many competent users instead revert to manual search, which can fully incorporate their prior knowledge. A recent survey showed that most NeurIPS 2019 and ICLR 2020 papers reported having tuned hyperparameters used manual search, with only a very small fraction using BO~\cite{bouthillier:survey}. In order for BO to be adopted widely, and help facilitate faster progress in the ML community by tuning hyperparameters faster and better, it is therefore crucial to devise a method that fully incorporates expert knowledge about the location of high-performance areas into BO. In this paper, we introduce Bayesian Optimization with a Prior for the Optimum (\name), a novel BO variant that combines priors for the optimum with a probabilistic model of the observations made. Our technical contributions are: 

\begin{itemize}
    \item 
    We introduce \emph{Bayesian Optimization with a Prior over the Optimum}, short \emph{\name}, which allows
    users to inject priors that were previously difficult to inject into BO, such as Gaussian, exponential, multimodal, and multivariate priors for the location of the optimum.
    To ensure robustness against misleading priors, \name gives more importance to the data-driven model as iterations progress, gradually forgetting the prior.
    
    \item \name's model bridges the gap between the well-established Tree-structured Parzen Estimator (TPE) methodology, which is based on Parzen kernel density estimators, 
    and standard
    BO probabilistic models, such as Gaussian Processes (GPs).
    This is made possible by using the Probability of Improvement (PI) criterion to derive from BO's standard posterior over functions $p(f| (\param_i,y_i)_{i=1}^t)$ the probability of an input $\param$ \edit{leading to good function values.}{being of high potential for the optimization}.
    

    \item We demonstrate the effectiveness of \name on a comprehensive set of synthetic benchmarks and real-world applications, showing that  knowledge about the locality of an optimum helps \name to achieve similar performance to current state-of-the-art on average $6.67\times$ faster on synthetic benchmarks and $1.49\times$ faster on a  real-world application. \name also achieves similar or better final performance on all benchmarks. 
\end{itemize}

\name is publicly available as part of the HyperMapper optimization framework\footnote{https://github.com/luinardi/hypermapper/wiki/prior-injection}.

\section{Background} \label{sec:bg}

\subsection{Bayesian Optimization} \label{sec:bg.bo}
\label{bayop}

Bayesian Optimization (BO) is an approach for optimizing an unknown function $f : \Param \rightarrow \mathbb{R}$ that is expensive to evaluate over an input space $\Param$. In this paper, we aim to minimize $f$, i.e., find 
$\param^{*} \in \argmin_{\param \in \Param} f(\param).$
BO approximates $\param^{*}$ with a \edit{}{optimal} sequence of evaluations $\param_{1}, \param_{2}, \ldots \in \Param$ \edit{that maximizes an utility metric}{}, with each new $\param_{t+1}$ depending on the previous function values $y_{1}, y_{2},\ldots, y_t$ at $\param_{1},\ldots, \param_{t}$. BO achieves this by building a posterior on $f$ based on the set of evaluated points. At each iteration, a new point is selected and evaluated based on the posterior, and the posterior is updated to include the new point $(\param_{t+1}, y_{t+1})$.

The points explored by BO are dictated by the acquisition function, which attributes an \edit{utility}{} to each $\param \in \Param$ by balancing the predicted value and uncertainty of the prediction for each $\param$~\cite{shahriari2015taking}.
%
In this work, as the acquisition function we choose
Expected Improvement (EI)~\cite{mockus1978application}, which quantifies the expected improvement over the best function value found so far:
\begin{equation} \label{eq:priorsop.ei0}
    EI_{y_{inc}}(\bm{x}):=\int_{-\infty}^{\infty} \max(y_{inc} - y, 0) p(y|\bm{x}) dy, 
\end{equation}
where $y_{inc}$ is the incumbent function value, i.e., the best objective function value found so far, and $p(y|\bm{x})$ is \newedit{given by}{} a probabilistic model, e.g., a GP. 
%
Alternatives to EI would be Probability of Improvement (PI)~\cite{jones2001taxonomy,kushner1964new}, upper-confidence bounds (UCB)~\cite{srinivas2009gaussian}, entropy-based methods~(e.g. Hernández-Lobato \etal~\cite{NIPS2014_5324}), and knowledge gradient~\cite{DBLP:conf/nips/WuPWF17}.
\subsection{Tree-structured Parzen Estimator} \label{sec:bg.tpe}

\newedit{The Tree-structured Parzen Estimator (TPE) method is a BO approach introduced by Bergstra \etal\cite{bergstra2011algorithms}}{}.
Whereas the standard probabilistic model in BO directly models $p(y|\param)$, the TPE approach models $p(\param|y)$ and $p(y)$ instead.\footnote{Technically, the model does not parameterize $p(y)$, since it is computed based on the observed data points, which are heavily biased towards low values due to the optimization process. Instead, it parameterizes a dynamically changing $p_t(y)$, which helps to constantly challenge the model to yield better observations.}
This is done by constructing two parametric densities, \edit{$g(\param)$}{} and \edit{$b(\param)$}{}, which are computed using the observations with function value below and above a given threshold, respectively. The separating threshold $y^*$ is defined as a quantile of the observed function values. TPE uses the densities \edit{$g(\param)$}{} and \edit{$b(\param)$}{} to define $p(\param|y)$ as: 
\begin{equation}
p(\param|y) = g(\param)I(y < y^*) + b(\param)(1 - I(y < y^*)),
\end{equation}
where $I(y < y^*)$ is $1$ when $y < y^*$ and $0$ otherwise. \edit{Bergstra \etal\cite{bergstra2011algorithms} show that}{} the parametrization of the generative model $p(\param, y) = p(\param|y)p(y)$ facilitates the computation of EI as it leads to \edit{$EI_{y^*}(\param) \propto g(\param)/b(\param)$}{} and, thus, \edit{$\argmax_{\param \in \Param} EI_{y^*}(\param) = \argmax_{\param \in \Param} g(\param)/b(\param)$.}{}

\section{BO with a Prior for the Optimum} \label{sec:priorop}


We now describe our \name approach, which allows domain experts to
inject user knowledge about the locality of an optimum into the optimization in the form of \priorsstr. \name combines this user-defined \priorstr with a probabilistic model that captures the likelihood of the observed data $\history = (\param_i,y_i)_{i=1}^t$.
\name is independent of the probabilistic model being used; it can be freely combined with, e.g., Gaussian processes (GPs), random forests, or Bayesian NNs. 

\subsection{\name \Priorsstr} \label{sec:priorop.priors}

\name allows users to inject prior knowledge w.r.t. promising areas into BO. This is done via a \priorstr distribution that informs where in the input space $\Param$ we expect to find good $f(\bm{x})$ values. A point is considered ``good'' if it leads to low function values, and potentially to a global optimum. 
We denote the \priorstr distribution $\priorgood$, where $g$ denotes that this is a \priorstr on good points and $\param \in \Param$ is a given point. \edit{Examples of priors are shown in Figures~\ref{fig:posterior_breakdown} and~\ref{fig:wrong_prior_exp}, additional examples of continuous and discrete \priorsstr are shown in Appendices~\ref{sec:forgetting.appendix} and~\ref{sec:spatial.appendix}, respectively.}{}  
Similarly, we define a \priorstr on where in the input space we expect to have ``bad'' points.
Although we could have a user-defined probability distribution $\priorbad$, we aim to keep the \newedit{decision-making}{} load on users low and thus, for simplicity, only require the definition of $\priorgood$ and compute $\priorbad = 1 - \priorgood$.\footnote{We note that for continuous spaces, this $\priorbad$ is not a probability distribution, and therefore only a pseudo-prior, as it does not integrate to $1$. For discrete spaces, we normalize $\priorbad$ so that it sums to $1$ and therefore is a proper probability distribution and prior.} 
$\priorgood$ is normalized to $[0,1]$ by min-max scaling before computing $\priorbad$.


In practice, $\bm{x}$ contains several dimensions but it is difficult for domain experts to provide a joint prior distribution $\priorgood$ for all of them. However, users can typically easily specify, e.g., sketch out, a univariate or bivariate prior distribution for continuous dimensions or provide a list of probabilities for
discrete dimensions.
In \name, users are free to define a complex multivariate distribution, but we expect the standard use case to be that users mainly want to specify univariate distributions, implicitly assuming a prior that factors as 
\label{eq:k_prior}
$    \priorgood = \prod_{i=1}^{D}P_g(x_i)$,
where $D$ is the number of dimensions in $\Param$ and $x_i$ is the $i$-th input dimension of $\param$. 
To not assume unrealistically complex priors and to mimic what we expect most users will provide, in our experiments we use factorized \priorsstr; in Appendix~\ref{sec:multivariate.appendix} we show that these factorized \priorsstr can in fact lead to similar BO performance as multivariate \priorsstr.

\subsection{Model} \label{sec:priorop.model}

\begin{figure}[tb]
    \begin{center}
    \includegraphics[width=0.75\linewidth]{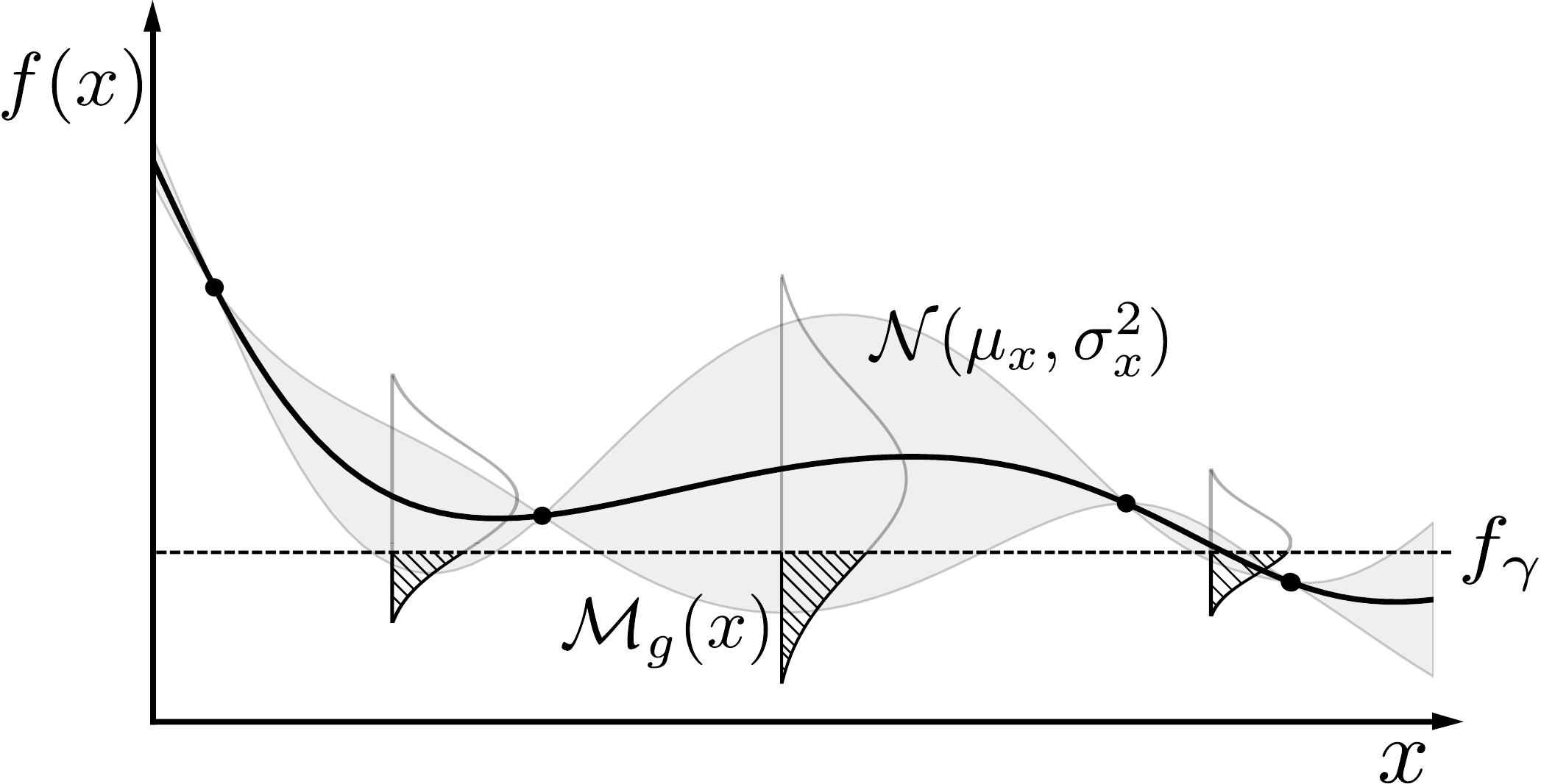}
    \caption{Our model is composed by a probabilistic model and the probability of improving over the threshold $f_{\gamma}$, i.e., right tail of the Gaussian. The black curve is the probabilistic model's mean and the shaded area is the model's variance.}
    \label{fig:priors_model}
    \end{center}
\end{figure}

Whereas the standard probabilistic model in BO, e.g., a GP, quantifies $p(y|\bm{x})$ directly, that model is hard to combine with the \priorstr $\priorgood$. 
We therefore introduce a method to translate the standard probabilistic model $p(y|\bm{x})$ into a model that is easier to combine with this \priorstr. Similar to the TPE work described in Section \ref{sec:bg.tpe},
our generative model combines $p(\bm{x}|y)$ and $p(y)$ instead of directly modeling $p(y|\bm{x})$. 

The computation we perform for this translation is to quantify the probability that a given input $\bm{x}$ is ``good'' under our standard probabilistic model $p(y|\bm{x})$. As in TPE, we define configurations as ``good'' if their observed $y$-value is below a certain quantile $\gamma$ of the observed function values (so that  $p(y < f_{\gamma}) = \gamma$).
We in addition exploit the fact that our standard probabilistic model $p(y|\bm{x})$ has a Gaussian form, and under this Gaussian prediction we can compute the probability $\modelgood$ of the function value lying below a certain quantile using the standard closed-form formula for PI~\cite{kushner1964new}: 

\begin{eqnarray}\label{eq:priorsop.model_good}
    \modelgood = p(f(\param) < f_{\gamma}|\param, \history) = \Phi\left( \dfrac{f_{\gamma} - \mu_{\param}}{\sigma_{\param}} \right),
\end{eqnarray}

where $\history = (\param_i,y_i)_{i=1}^t$ are the evaluated configurations, $\mu_{\param}$ and $\sigma_{\param}$ are the predictive mean and standard deviation of the probabilistic model at $\param$, and $\Phi$ is the standard normal CDF, see Figure~\ref{fig:priors_model}. 
Note that there are two probabilistic models here: 
\begin{enumerate}
    \item The standard probabilistic model of BO, with a \newedit{structural}{} prior over functions $p(f)$, 
    updated by data $\history$ to yield a posterior over functions $p(f | \history )$, allowing us to quantify the probability $\modelgood  = p(f(x) < f_\gamma | \param, \history)$ 
    in Eq.~\eqref{eq:priorsop.model_good}.\footnote{\newedit{We note that the structural prior $p(f)$ and the optimum-prior $\priorgood$ provide orthogonal ways to input prior knowledge. $p(f)$ specifies our expectations about the structure and smoothness of the function, whereas $\priorgood$ specifies knowledge about the location of the optimum.}{}}
    \item The TPE-like generative model that combines $p(y)$ and $p(\param|y)$ instead of directly modelling $p(y|\param)$.
\end{enumerate}
Eq.~\eqref{eq:priorsop.model_good} bridges these two models by using the probability of improvement from BO’s standard probabilistic model as the probability $\modelgood$ in TPE’s model. Ultimately, this is a heuristic since there is no formal connection between the two probabilistic models. However, we believe that the use of BO’s familiar, theoretically sound framework of probabilistic modelling of $p(y|x)$, followed by the computation of the familiar PI formula is an intuitive choice for obtaining the probability of an input achieving at least a given performance threshold -- exactly the term we need for TPE’s $\modelgood$.
\postrebuttal{Similarly, we also define a probability $\modelbad$ of $\bm{x}$ being bad as $\modelbad = 1 - \modelgood$.}

\subsection{Pseudo-posterior} \label{sec:priorop.posterior}

\name combines the \priorstr{} \edit{$\priorgood$}{} in Section~\eqref{eq:k_prior} and the model \edit{$\modelgood$}{} in Eq.~\eqref{eq:priorsop.model_good} into a \posteriorstr on ``good'' points. This \posteriorstr represents the updated beliefs on where we can find good points, based on the \priorstr and data that has been observed. The \posteriorstr is computed as the product:
\begin{equation}\label{eq:posterior}
    g(\param) \propto \priorgood \modelgood^{\tfrac{t}{\beta}},
\end{equation}
where $t$ is the current optimization iteration, $\beta$ is an optimization hyperparameter, $\modelgood$ is defined in Eq.~\eqref{eq:priorsop.model_good}, and $\priorgood$ is the \priorstr defined in Sec~\ref{sec:priorop.priors}, rescaled to [0, 1] \newedit{using min-max scaling}{}. 
We note that 
this \posteriorstr is not normalized, but this suffices for \name to determine the next $\param_t$ as the normalization constant cancels out (c.f. Section~\ref{sec:priorop.ei}). Since $g(\param)$ is not normalized and we include the exponent $t/\beta$ in Eq.~\eqref{eq:posterior}, we refer to $g(\param)$ as a pseudo-posterior, to emphasize that it is not a standard posterior probability distribution. 

The $t/\beta$ fraction in Eq.~\eqref{eq:posterior} controls how much weight is given to \edit{$\modelgood$}{}. As the optimization progresses, more weight is given to \edit{$\modelgood$}{} over $\priorgood$. Intuitively, we put more emphasis on \edit{$\modelgood$}{} as it observes more data and becomes more accurate. We do this under the assumption that the model \edit{$\modelgood$}{} will eventually be better than the user at predicting where to find good points. This also allows to recover from misleading \priorsstr as we show in Section~\ref{sec:experiments.prior}; similar to, and inspired by Bayesian models, the data ultimately washes out the prior. The $\beta$ hyperparameter defines the balance between \priorstr and model, with higher $\beta$ values giving more importance to the \priorstr and requiring more data to overrule it. 

We note that,   
directly computing Eq~\eqref{eq:posterior}
can lead to numerical issues. Namely, the \posteriorstr can reach extremely low values if the  \edit{$\priorgood$ and $\modelgood$}{} probabilities are low, especially as $t/\beta$ grows. To prevent this, in practice, \name uses the logarithm of the \posteriorstr instead: 
\begin{equation}\label{eq:posterior.log}
   \log(g(\param)) \propto \log(\priorgood) + \tfrac{t}{\beta} \cdot \log(\modelgood).
\end{equation}

Once again, we also define an analogous \posteriorstr distribution on bad $\param$: $b(\param) \propto \priorbad \modelbad^{\tfrac{t}{\beta}}$.
We then use these quantities 
to define a density model $p(\bm{x}|y)$ as follows:
\begin{equation}
   p(\bm{x}|y) \propto
    \begin{cases}
      g(\bm{x}) & \textrm{if} \;\; y < f_{\gamma}\\
      b(\bm{x}) & \textrm{if} \;\; y \geq f_{\gamma}.
    \end{cases}       
\end{equation}


\subsection{Model and Pseudo-posterior Visualization} \label{sec:visualization.appendix}

\begin{figure*}[tb]
\centering
\subfigure{\includegraphics[width=0.203\textwidth]{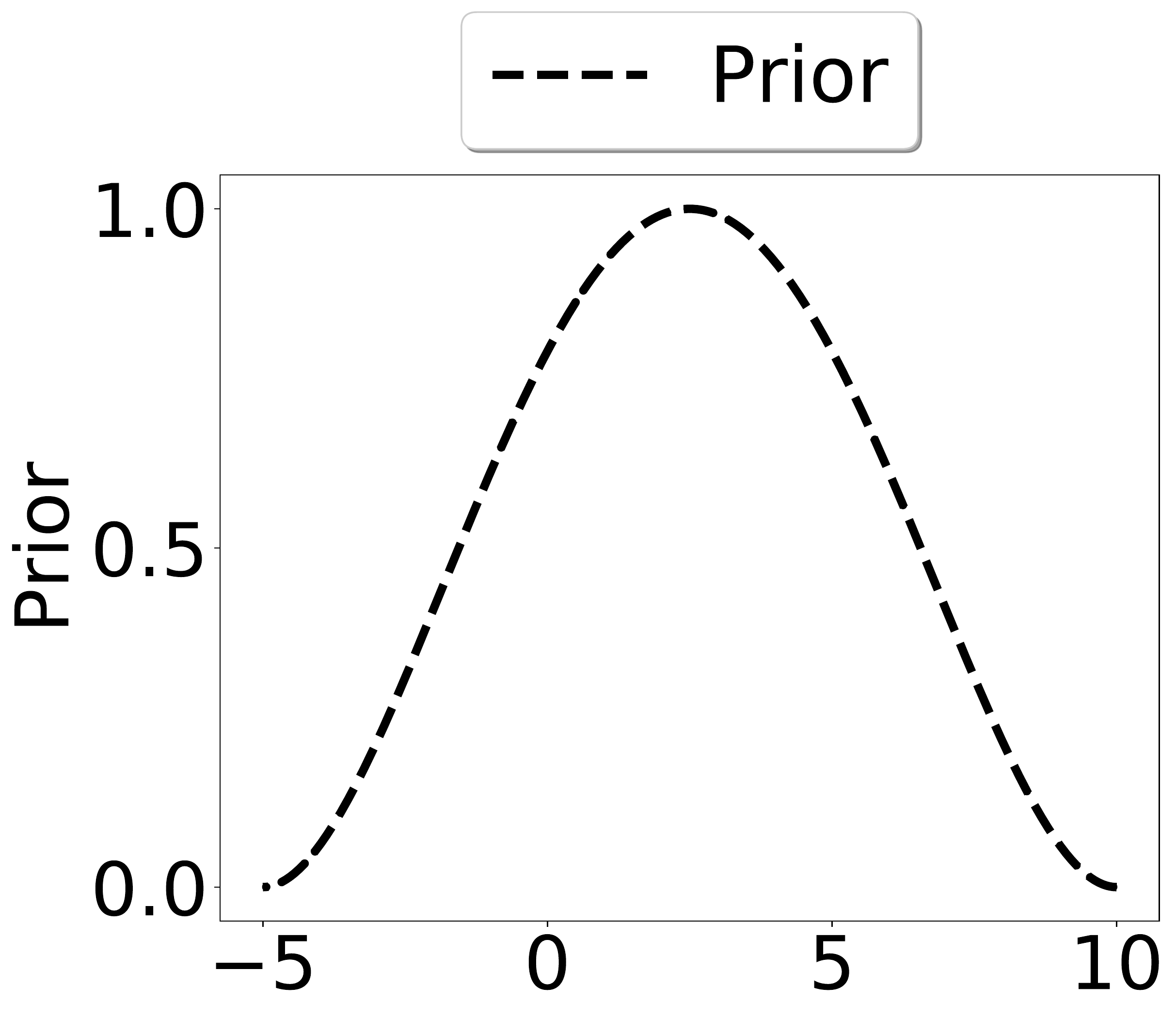}}\hspace{0.3cm}
\subfigure{\includegraphics[width=0.245\textwidth]{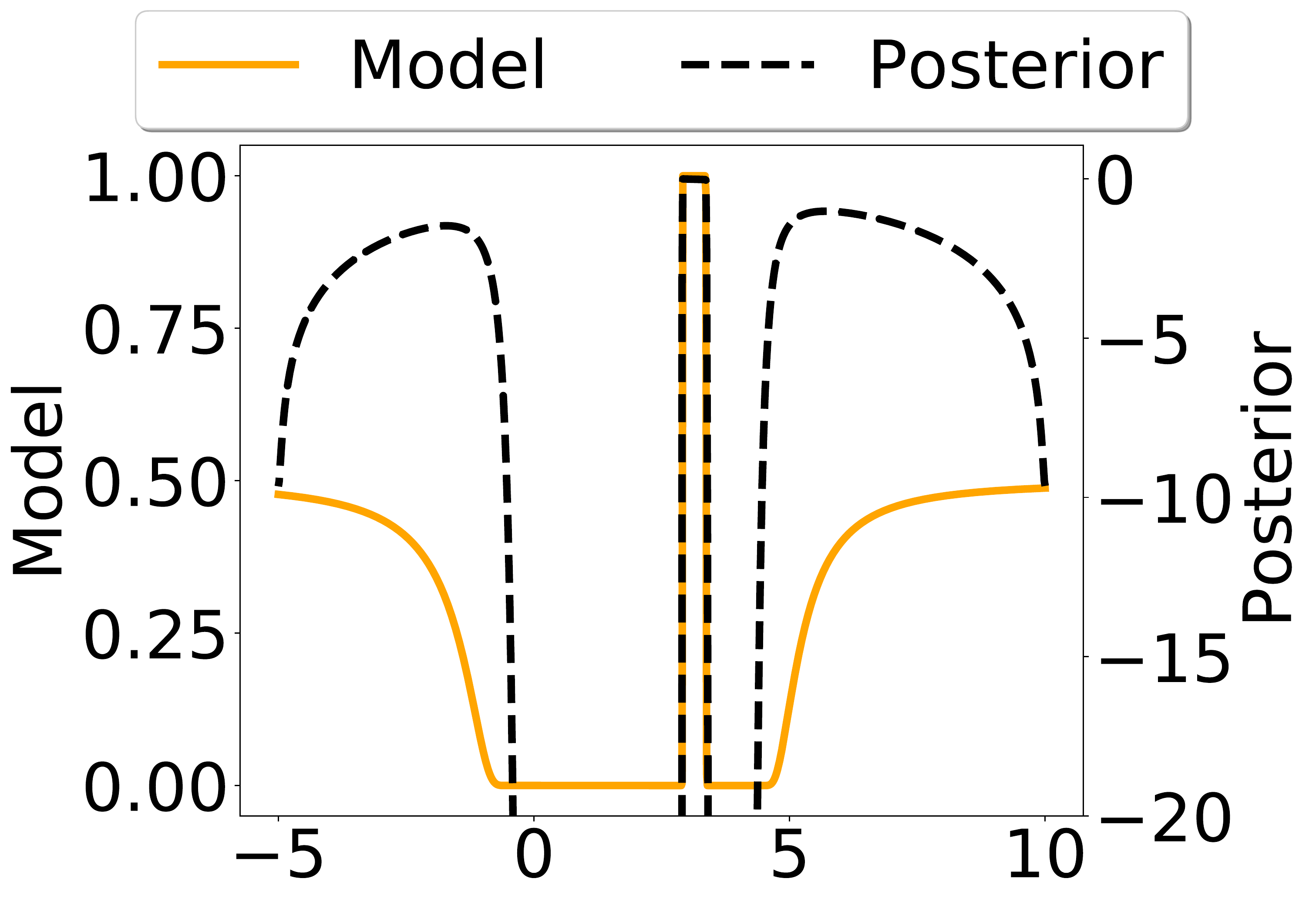}}
\subfigure{\includegraphics[width=0.245\textwidth]{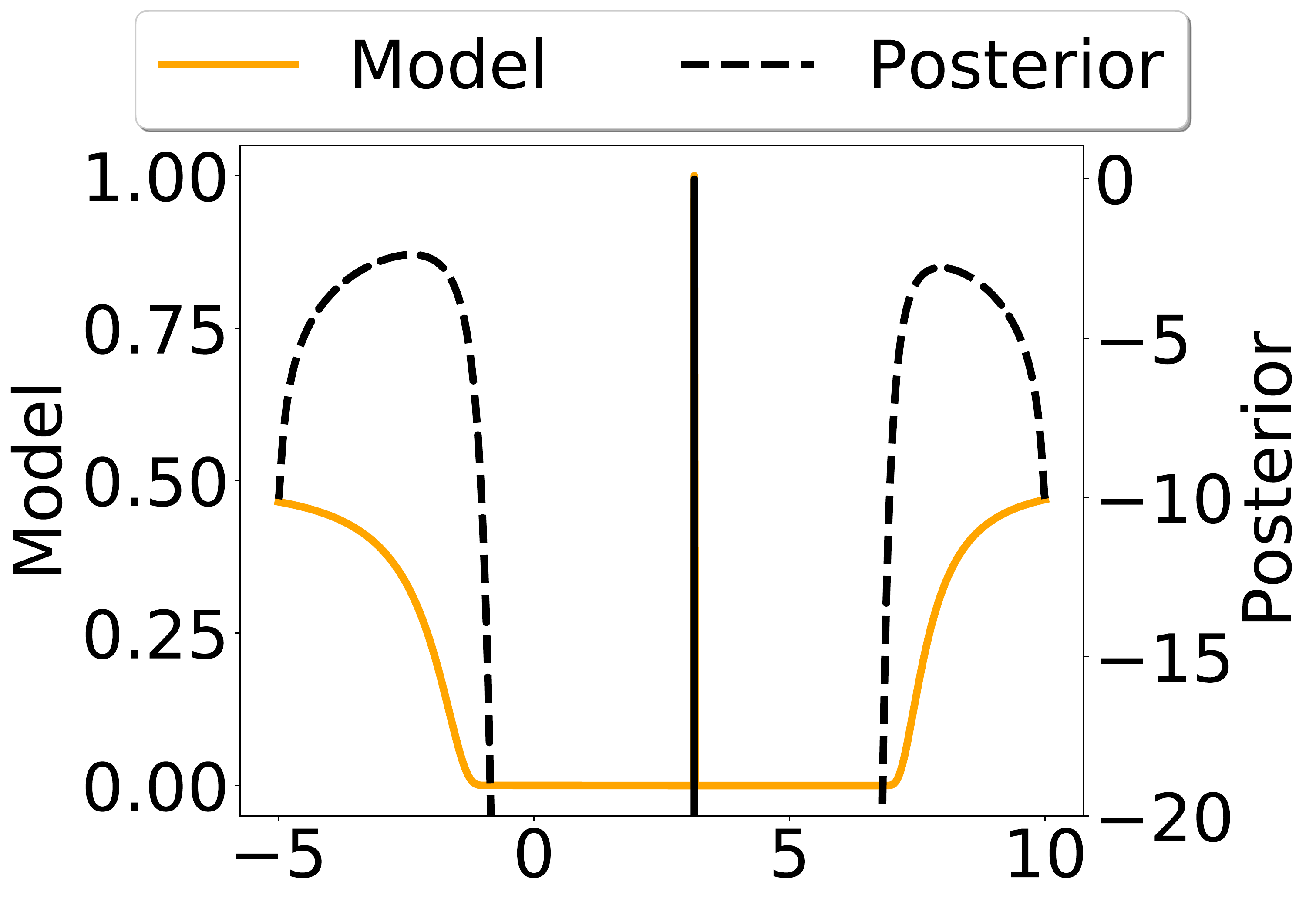}}
\subfigure{\includegraphics[width=0.245\textwidth]{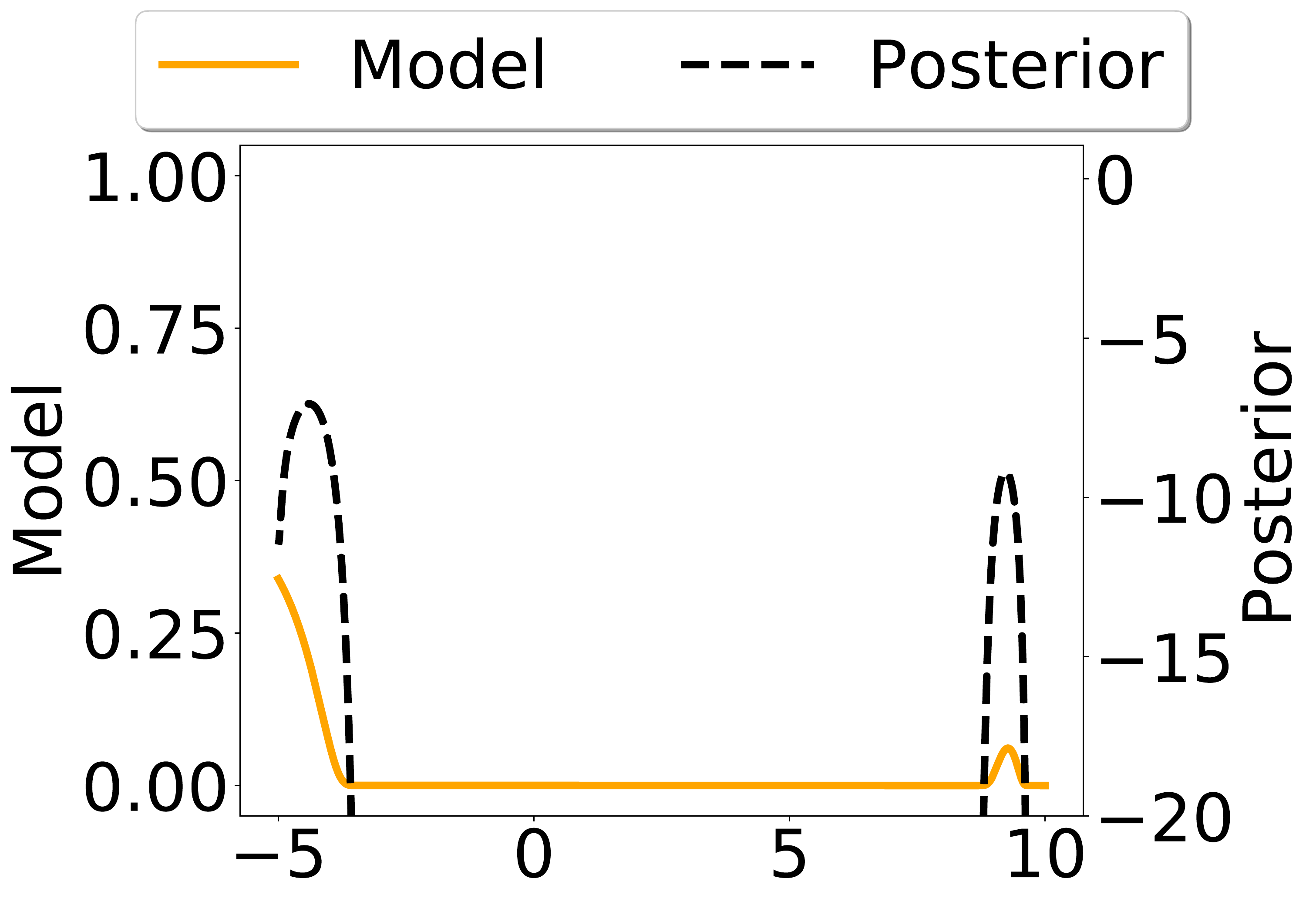}}
\centering\subfigure{\includegraphics[width=0.8\textwidth]{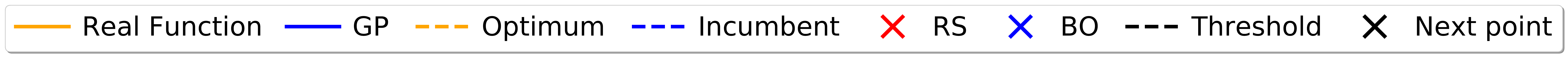}}\\\vspace{-0.3cm}
\setcounter{subfigure}{0}
\subfigure[0 BO iterations]{\includegraphics[width=0.24\textwidth]{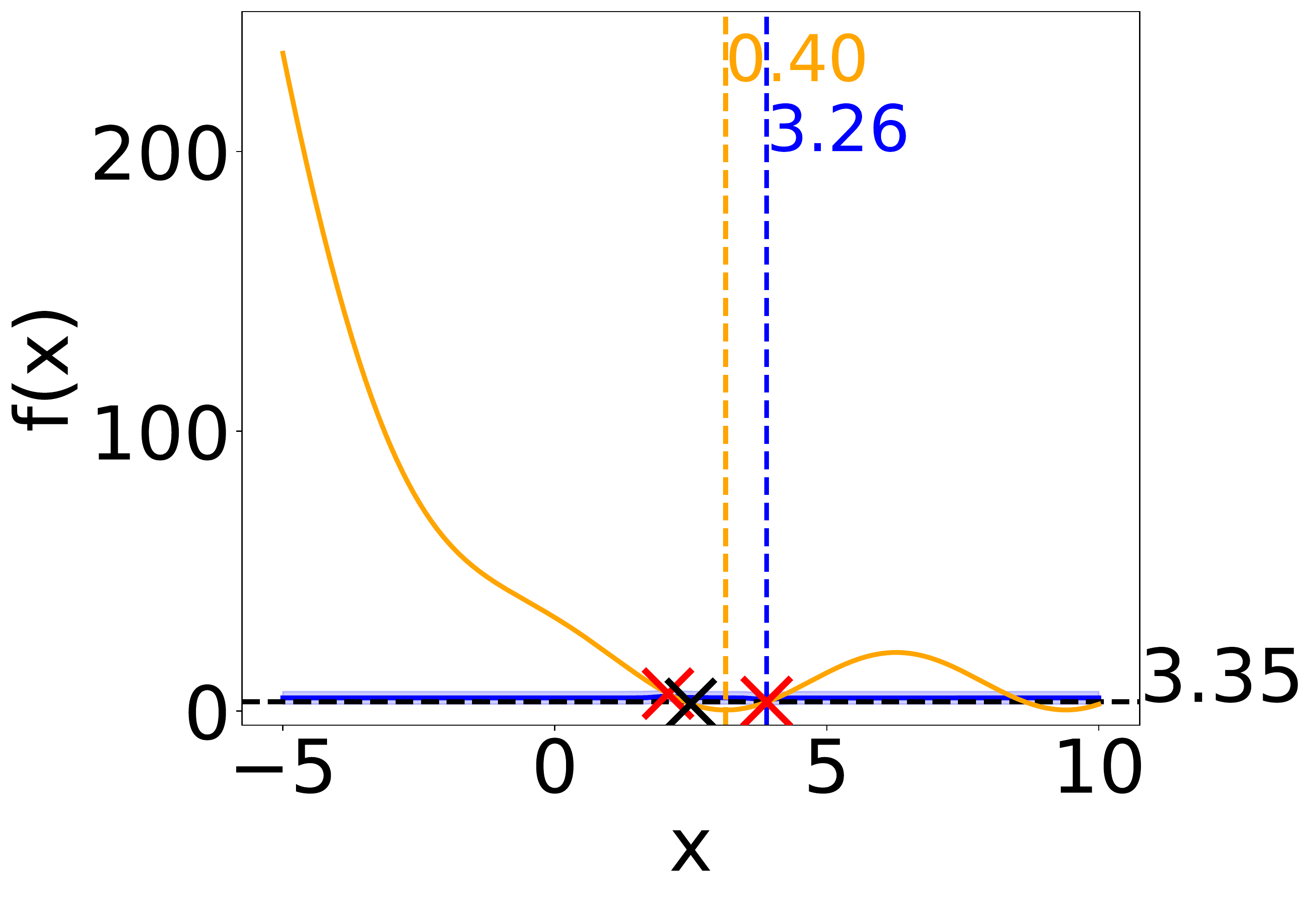}}
\subfigure[5 BO iterations]{\includegraphics[width=0.24\textwidth]{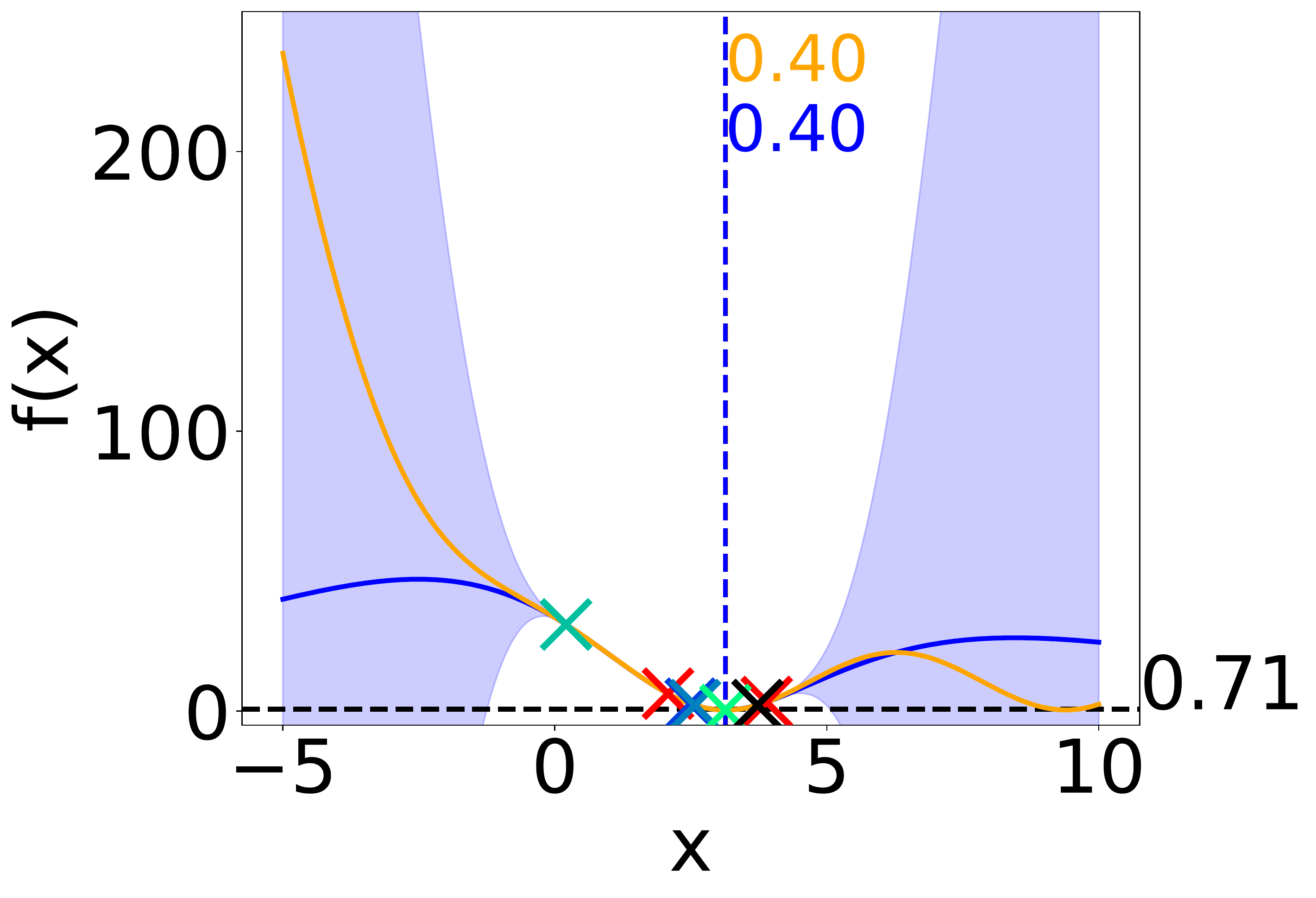}}
\subfigure[10 BO iterations]{\includegraphics[width=0.24\textwidth]{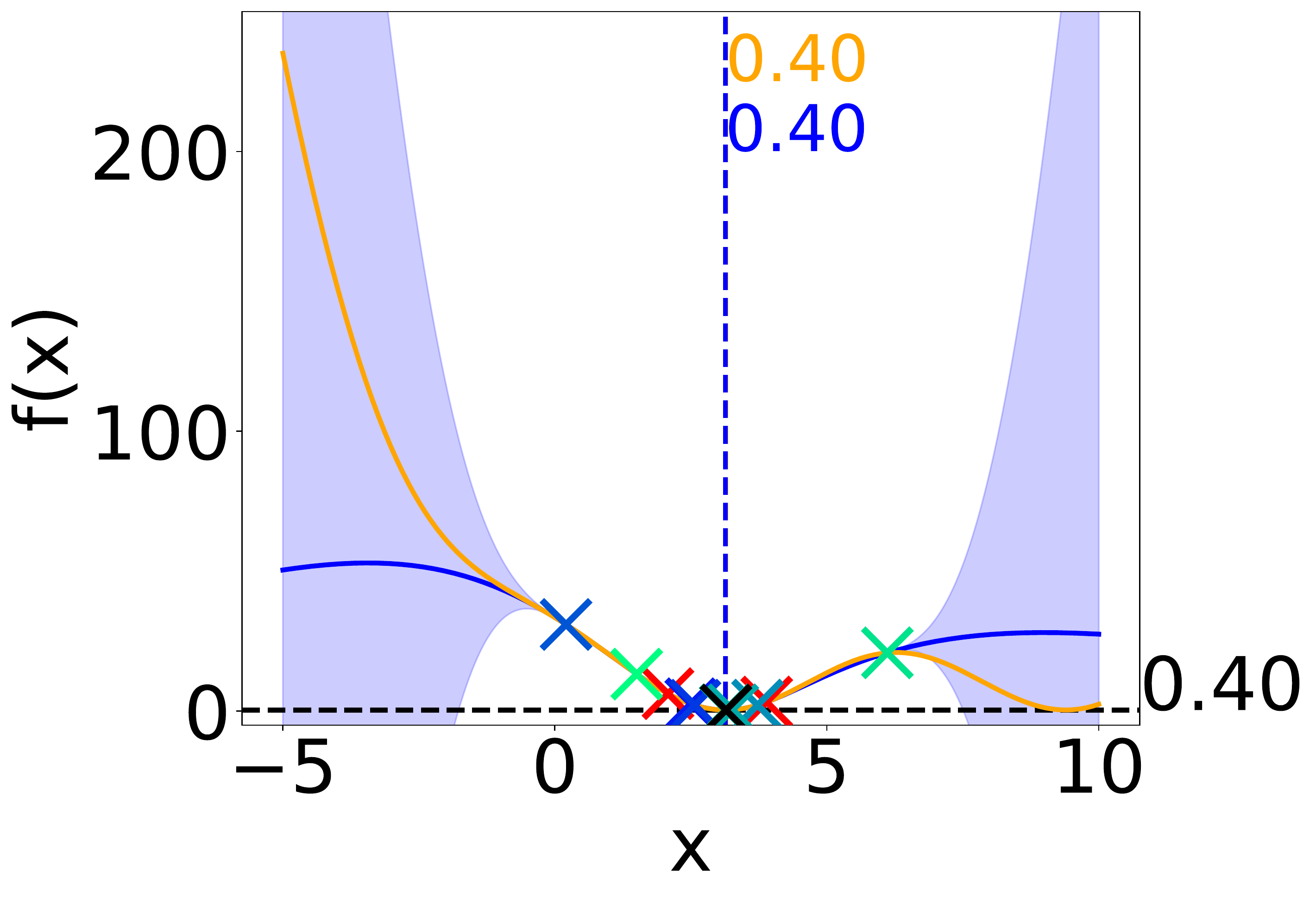}}
\subfigure[20 BO iterations]{\includegraphics[width=0.24\textwidth]{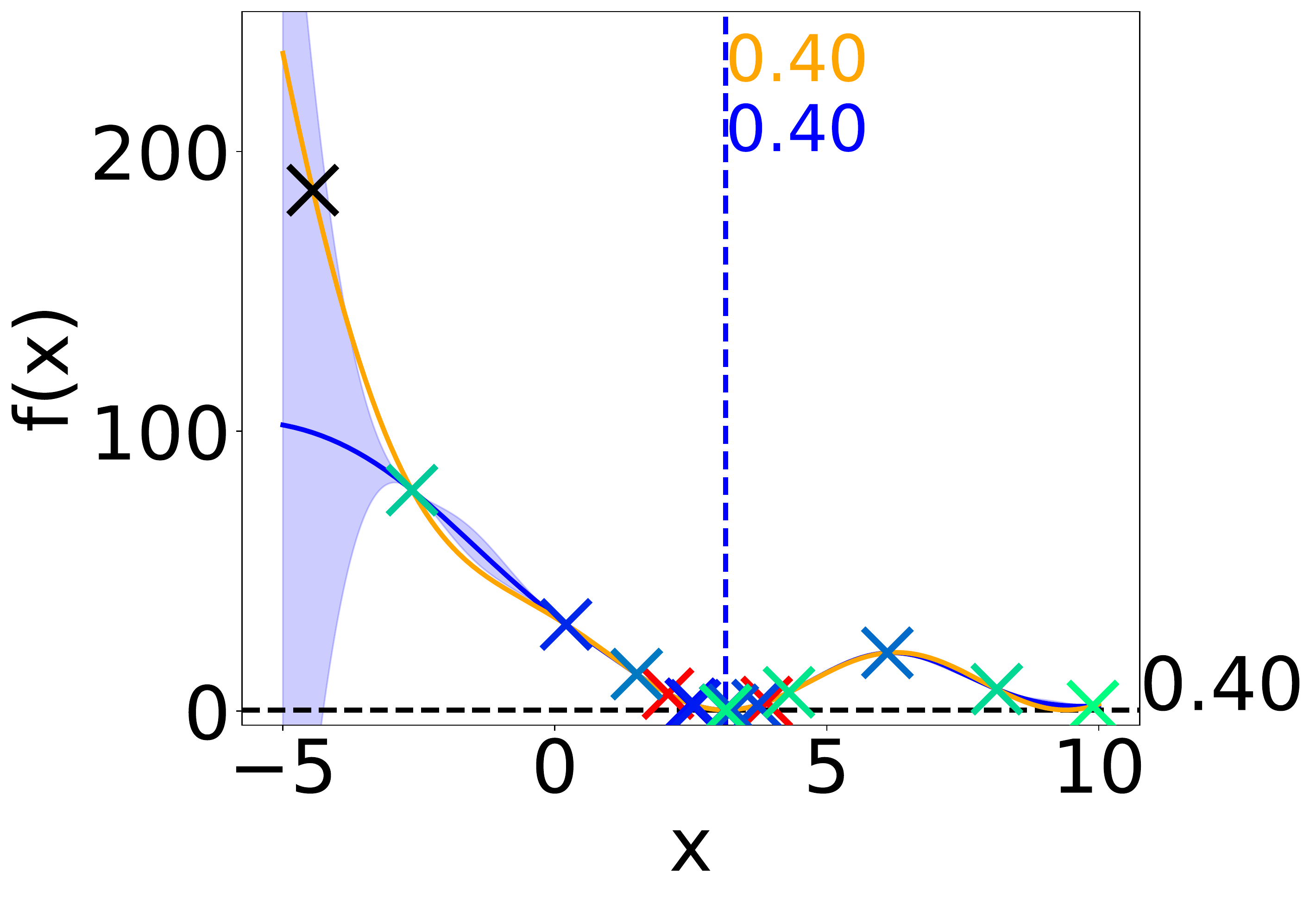}}
\caption{Breakdown of the \priorstr $\priorgood=\mathcal{B}(3, 3)$, the model $M_g(\param)$, and the \posteriorstr $g(\param)$ (top row) on the 1D-Branin function (bottom row) and their evolution over the optimization iterations.
In early iterations, the \posteriorstr is high around the optimum, where both prior and model agree there are good points. In later iterations, it vanishes where the model is certain there will be no improvement and is high where there is uncertainty in the model. 
}
\label{fig:posterior_breakdown}
\end{figure*}

We visualize the \priorstr $P_g(\param)$, the model $M_g(\param)$, and the \posteriorstr $g(\param)$ and their evolution over the optimization iterations for a 1D-Branin function. We define the 1D-Branin by setting the second dimension of the function to the global optimum $x_2 = 2.275$ and optimizing the first dimension.
We use a Beta distribution \priorstr $\priorgood=\mathcal{B}(3, 3)$, which resembles a truncated Gaussian centered close to the global optimum, and a GP as predictive model. 
We perform an initial design of $D+1 = 2$ random points sampled from the prior and then run \name for 20 iterations. 

Figure~\ref{fig:posterior_breakdown} shows the optimization at different stages.
Red crosses denote the initial design and blue/green crosses denote \name samples, with green samples denoting later iterations.
Figure~\ref{fig:posterior_breakdown}a shows the initialization phase (bottom) and the Beta \priorstr (top).
After 5 BO iterations, in Figure~\ref{fig:posterior_breakdown}b (top), the \posteriorstr is high near the global minimum, around $x = \pi$, where both the \priorstr{} \edit{$\priorgood$}{} and the model  \edit{$\modelgood$}{} agree there are good points.
After 10 BO iterations in Figure~\ref{fig:posterior_breakdown}c (top), there are three regions with high \posteriorstr. The middle region, where \name is exploiting until the optimum is found, and two regions to the right and left, which will lead to future exploration as shown in Figure~\ref{fig:posterior_breakdown}d (bottom) on the right and left of the global optimum in light green crosses.  
After 20 iterations, see Figure~ \ref{fig:posterior_breakdown}d (top), the \posteriorstr vanishes where the model $\modelgood$ is certain there will be no improvement, but it is high wherever there is uncertainty in the GP. 


\subsection{Acquisition Function} \label{sec:priorop.ei}


We adopt the EI formulation used in Bergstra \etal\cite{bergstra2011algorithms} by replacing their Adaptive Parzen Estimators with our \posteriorstr from Eq.~\eqref{eq:posterior}, i.e.:
\begin{align} \label{eq:priorsop.ei2}
    EI_{f_{\gamma}}(\bm{x}) &:= \int_{-\infty}^{\infty} \max(f_{\gamma} - y, 0) p(y|\bm{x}) dy \nonumber  
    = \int_{-\infty}^{f_{\gamma}} (f_{\gamma} - y)\frac{p(\bm{x}|y) p(y)}{p(\bm{x})} dy \nonumber \\
    & \propto \left( \gamma + \dfrac{b(\param)}{g(\param)}(1 - \gamma) \right)^{-1}.
\end{align}




The full derivation of Eq.~\eqref{eq:priorsop.ei2} is shown in Appendix~\ref{sec:proofs}. Eq.~\eqref{eq:priorsop.ei2} shows that to maximize improvement we would like points $\bm{x}$ with high probability under $g(\bm{x})$ and low probability under $b(\bm{x})$,
i.e., 
minimizing the ratio $b(\param)/g(\param)$. We note that the point that minimizes the ratio for our unnormalized \posteriorsstr will be the same that minimizes the ratio for the normalized \posteriorstr and, thus, computing the normalized \posteriorsstr is unnecessary.

The dynamics of the \name algorithm 
can be understood in terms of the
following proposition (proof in Appendix~\ref{sec:proofs}): 

\begin{proposition} \label{prop:priorsop1}
Given $f_{\gamma}$, $\priorgood$, $\priorbad$, $\modelgood$, $\modelbad$, $g(\param)$, $b(\param)$, $p(\bm{x}|y)$, and $\beta$ as above, then
\begin{equation*}
\lim_{t\to\infty} \argmax_{\param \in \Param} EI_{f_{\gamma}}(\bm{x}) = \lim_{t\to\infty} \argmax_{\param \in \Param}  
\modelgood,
\end{equation*}
where $EI_{f_{\gamma}}$ is the Expected Improvement acquisition function as defined in Eq.~\eqref{eq:priorsop.ei2} and 
$\modelgood$ is as defined in Eq.~\eqref{eq:priorsop.model_good}.
\end{proposition}

In early BO iterations the prior for the optimum will have a predominant role, but in later BO iterations the model will grow more important, and as Proposition~\ref{prop:priorsop1} shows, if \name is run long enough the prior washes out and \name \emph{only} trusts the model $\modelgood$ informed by the data.
Since $\modelgood$ is the Probability of Improvement (PI) on the probabilistic model $p(y|\param)$ then, in the limit, maximizing the acquisition function $EI_{f_{\gamma}}(\param)$ is equivalent to maximizing the PI acquisition function on the probabilistic model $p(y|\param)$. 
In other words, for high values of $t$, \name converges to standard BO with a PI acquisition function.

\begin{algorithm}[tb]
    \caption{\name Algorithm.
   $\history$ keeps track of all function evaluations so far: $(\param_i, y_i)_{i=1}^t$.}
   \label{alg:bo}
\begin{algorithmic}[1]
   \State {\bfseries Input:} Input space $\Param$, user-defined \priorstr distributions $\priorgood$ and $\priorbad$, quantile $\gamma$ and BO budget $B$. 
   \State {\bfseries Output:} Optimized point $\param_{inc}$.
   \State $\edit{\mathcal{D}_1}{} \leftarrow Initialize(\Param)$ \label{alg:doe}
   \For{$t=1$ {\bfseries to} $B$} \label{alg:BOloop}
   \State $\modelgood \leftarrow fit\_model\_good(\history)$ \label{alg:prob_model_training1} \algorithmiccomment{\newedit{see Eq.~\eqref{eq:priorsop.model_good}}{}}
   \State $\modelbad \leftarrow fit\_model\_bad(\history)$ \label{alg:prob_model_training2}
   \State $g(\param) \leftarrow \priorgood \cdot \modelgood^{\tfrac{t}{\beta}}$ \label{alg:modelgood}  \algorithmiccomment{\newedit{see Eq.~\eqref{eq:posterior}}{}}
   \State $b(\param) \leftarrow \priorbad \cdot \modelbad^{\tfrac{t}{\beta}}$ \label{alg:modelbad}
   \State $\param_t \in \argmax_{\param \in \Param} EI_{f_{\gamma}}(\param)$ \algorithmiccomment{\newedit{see Eq.~\eqref{eq:priorsop.ei2}}{}}
   \State $y_t \leftarrow f(\param_t)$ \label{alg:bbf_evaluation}
   \State $\edit{\mathcal{D}_{t+1}}{}  \leftarrow \history \cup (\param_t, y_t)$
   \EndFor
   \State $\param_{inc} \leftarrow ComputeBest(\edit{\mathcal{D}_{t+1}}{})$
   \State \textbf{return} $\param_{inc}$
\end{algorithmic}
\end{algorithm}

\subsection{Putting It All Together} \label{sec:priorop.bo}
Algorithm~\ref{alg:bo} shows the \name algorithm.
In Line~3, \name starts with a design of experiments (DoE) phase, where it randomly samples a number of points from the user-defined \priorstr $\priorgood$. After initialization, the BO loop starts at Line~4. In each loop iteration, \name fits the models \edit{$\modelgood$ and $\modelbad$}{} on the previously evaluated points (Lines 5 and 6)
and computes the \posteriorsstr $g(\param)$ and $b(\param)$ (Lines 7 and 8 respectively). The EI acquisition function is computed next, using the \posteriorsstr, and the point that maximizes EI is selected as the next point to evaluate at Line~9. The black-box function evaluation is performed at Line~10. This BO loop is repeated for a predefined number of iterations, according to the user-defined budget $B$. 

\section{Experiments} \label{sec:experiments}

We implement both Gaussian processes (GPs) and random forests (RFs) as predictive models and use GPs in all experiments, except for our real-world experiments (Section~\ref{sec:experiments.spatial}), where we use RFs for a fair comparison.  
We set the model weight $\beta = 10$ and the model quantile to $\gamma = 0.05$, see our sensitivity hyperparameter study in Appendices~\ref{sec:gamma.appendix} and~\ref{sec:beta.appendix}. Before starting the main BO loop in \name, we randomly sample $D+1$ points from the \priorstr as an initial design consistently on all benchmarks. 
We optimize our EI acquisition function using a combination of multi-start local search~\cite{hutter2011sequential} and CMA-ES~\cite{hansen96cmaes}. 
We consider four synthetic benchmarks: Branin, SVM, FC-Net, and XGBoost, which are 2, 2, 6, and 8 dimensional, respectively. The last three are part of the Profet benchmarks~\cite{klein2019meta}, generated by a generative model built using performance data on OpenML or UCI datasets. See Appendix~\ref{sec:experimental_setup.appendix} for more details. 

\subsection{\Priorstr Forgetting} 
\label{sec:experiments.prior}

\begin{figure*}[tb]
\centering
\subfigure{\includegraphics[width=0.203\textwidth]{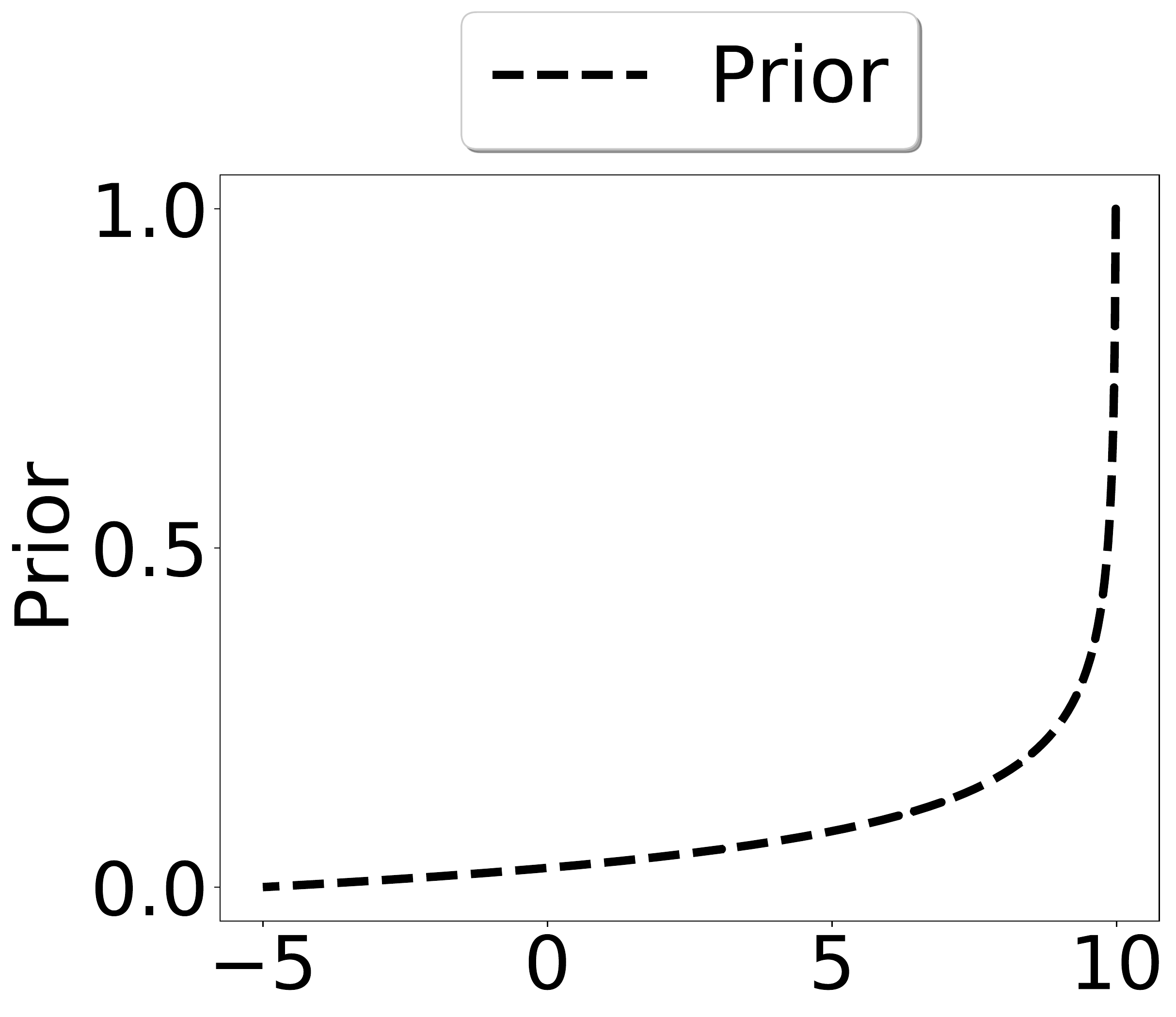}}
\subfigure{\includegraphics[width=0.245\textwidth]{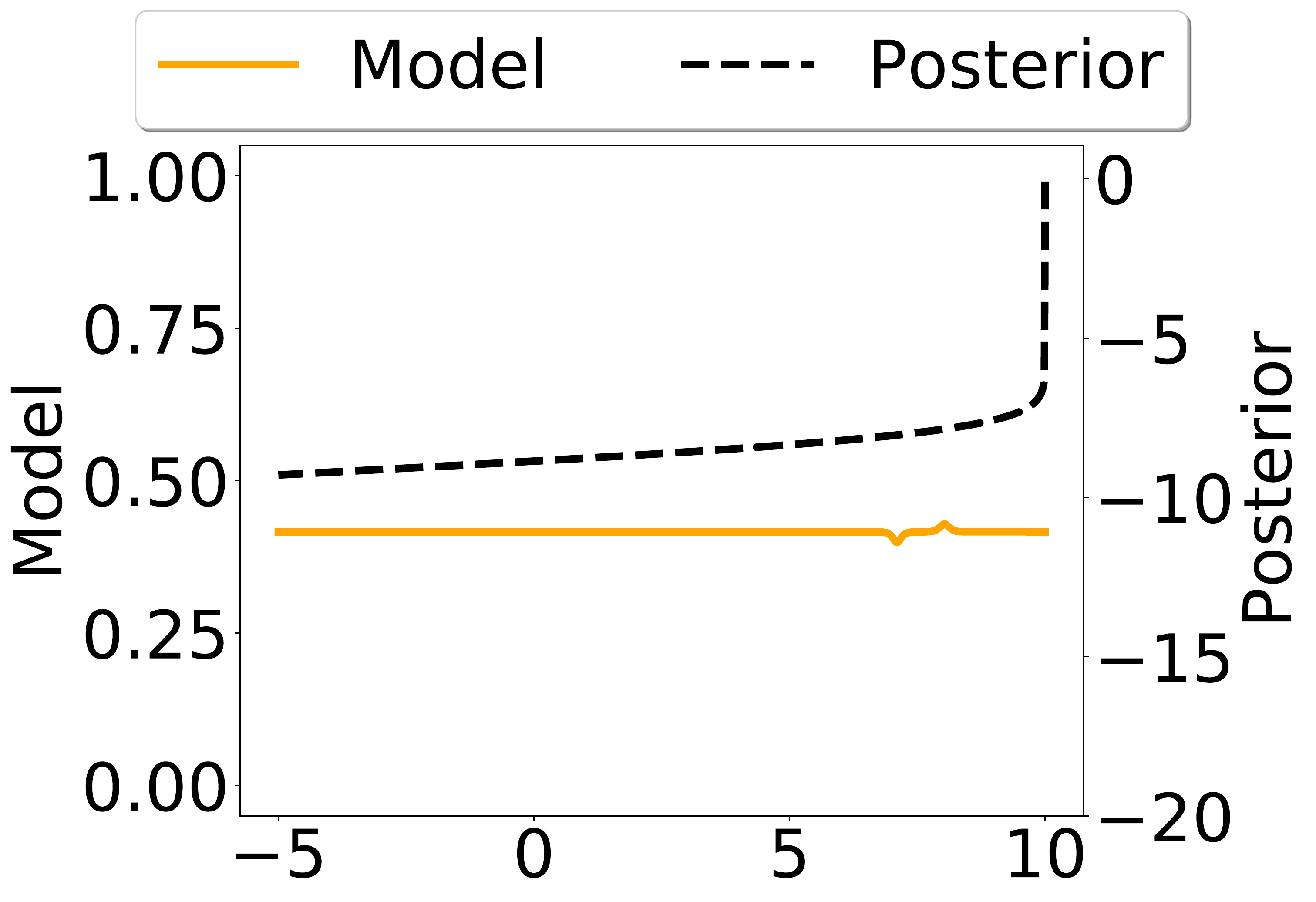}}
\subfigure{\includegraphics[width=0.245\textwidth]{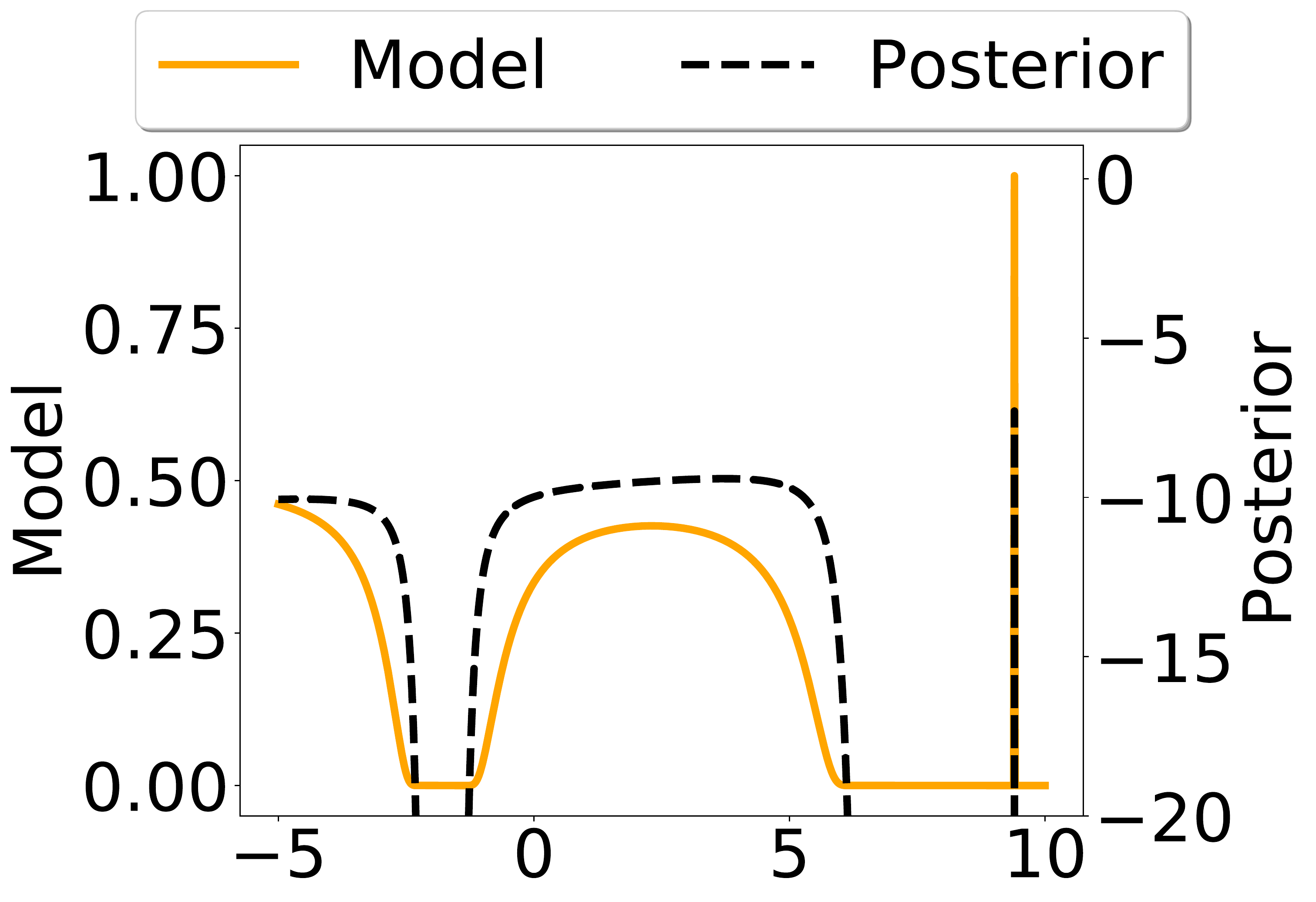}}
\subfigure{\includegraphics[width=0.245\textwidth]{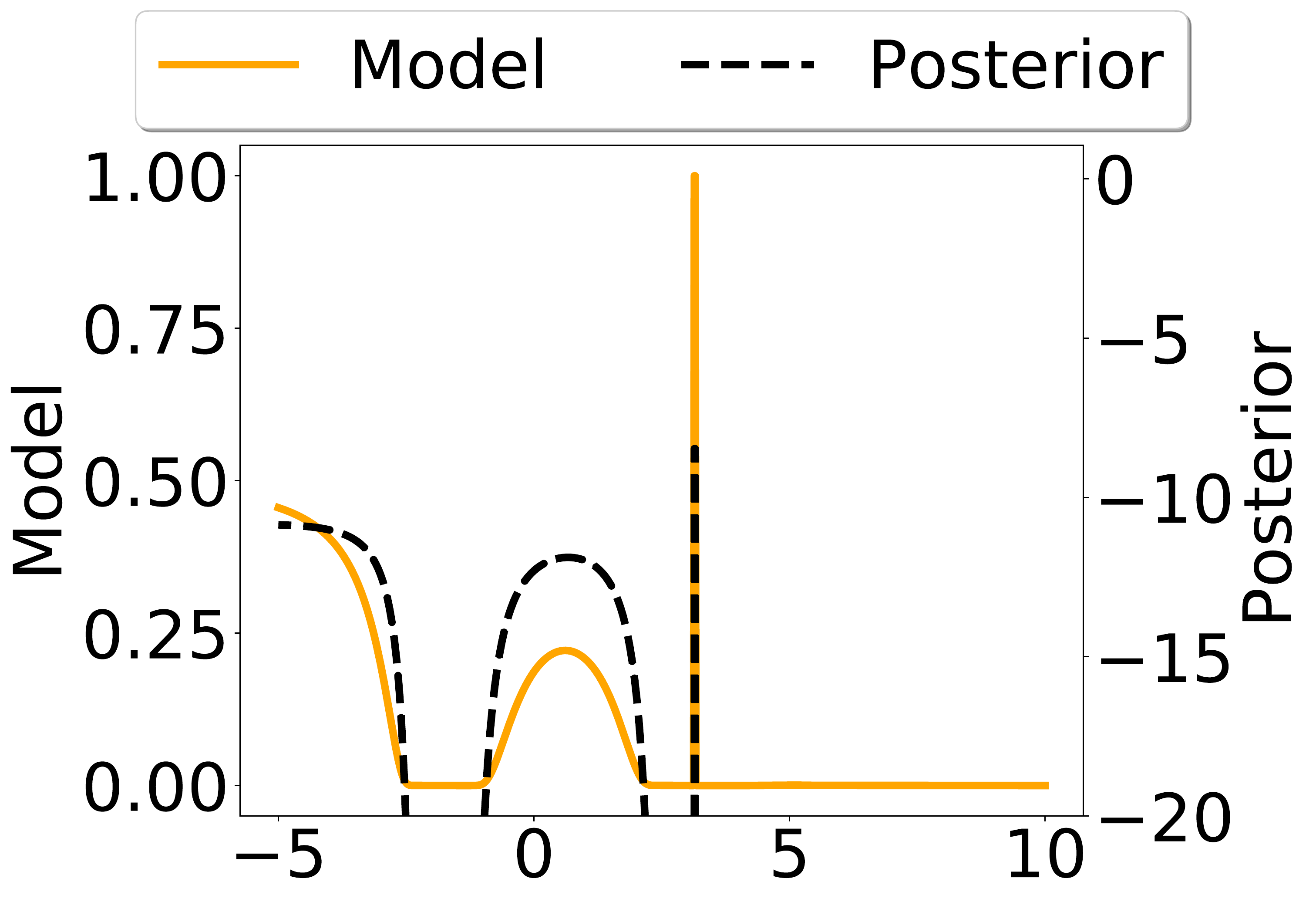}}
\centering\subfigure{\includegraphics[width=0.8\textwidth]{fig/nips_exponential_forgetting/legend.pdf}}\\\vspace{-0.3cm}
\setcounter{subfigure}{0}

\subfigure[No samples]{\includegraphics[width=0.215\textwidth]{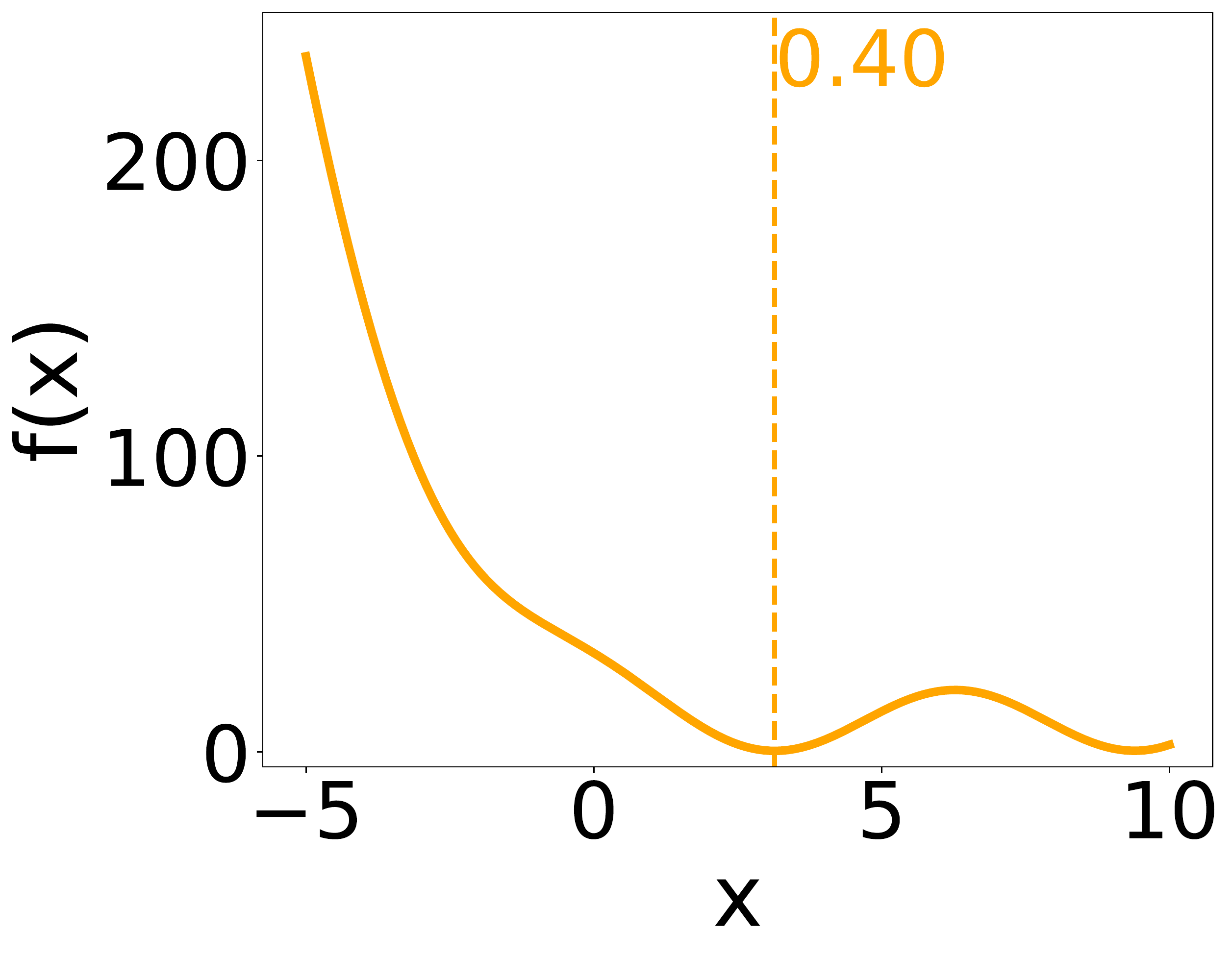}}
\subfigure[0 BO iterations]{\includegraphics[width=0.24\textwidth]{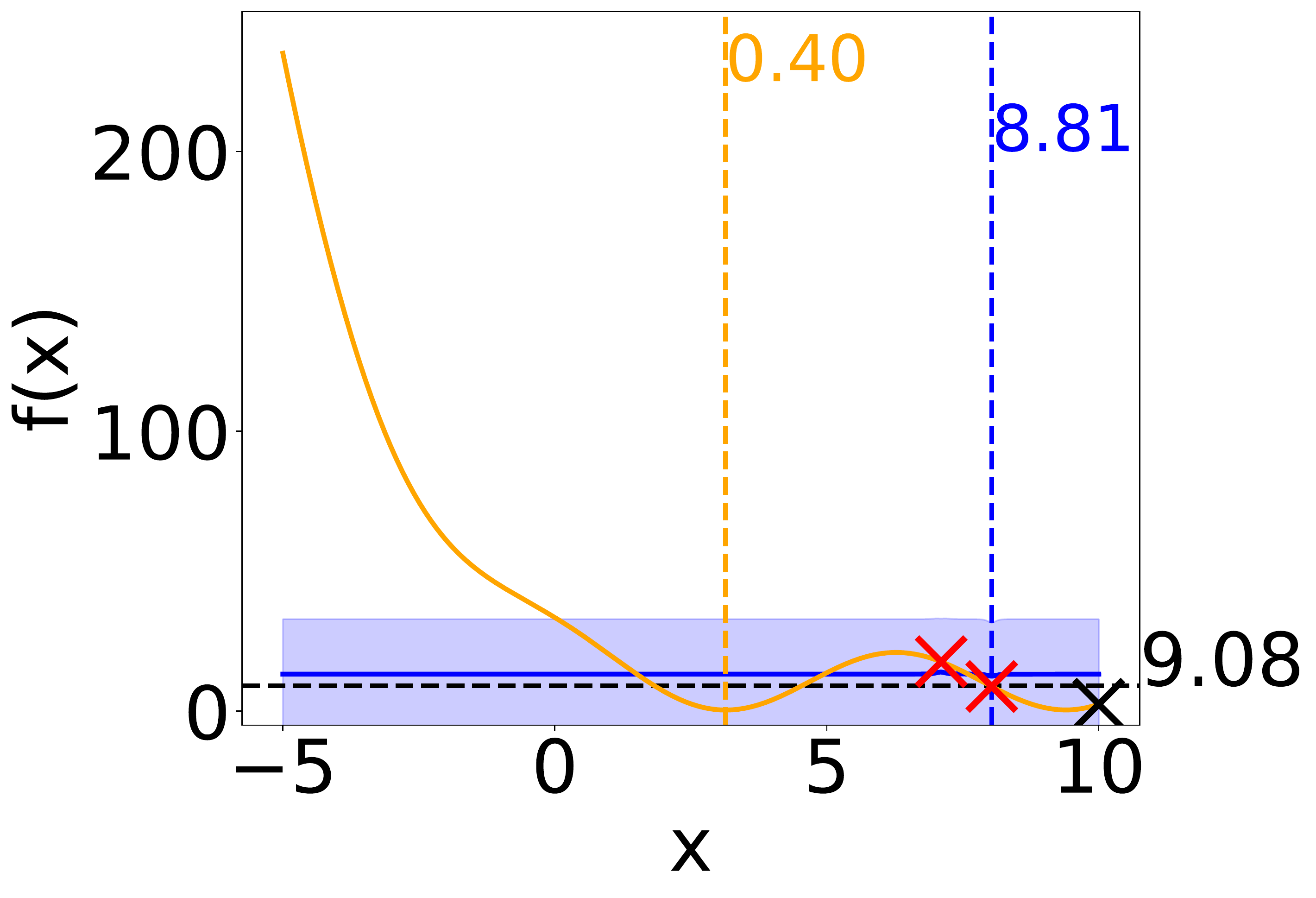}}
\subfigure[10 BO iterations]{\includegraphics[width=0.24\textwidth]{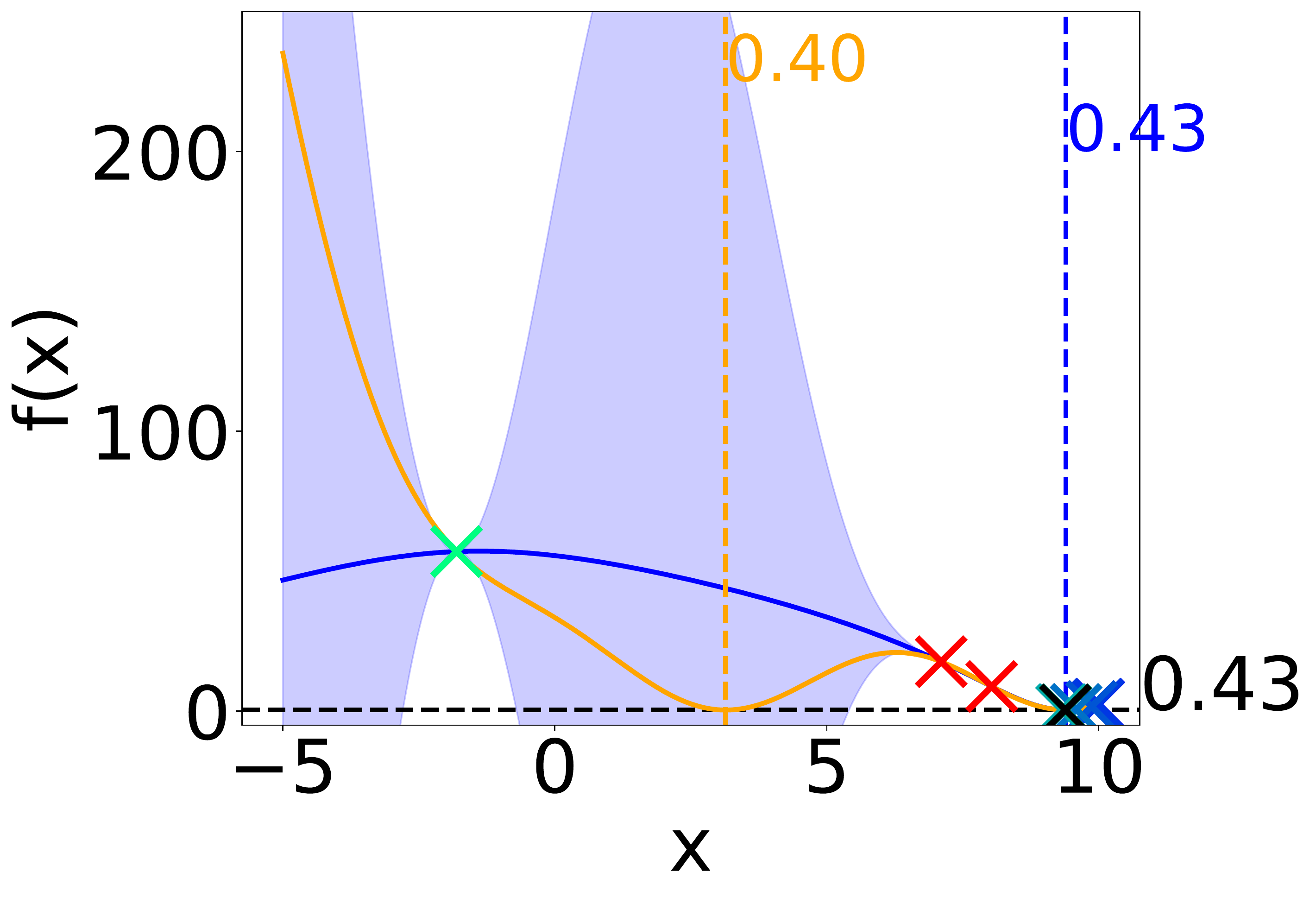}}
\subfigure[20 BO iterations]{\includegraphics[width=0.24\textwidth]{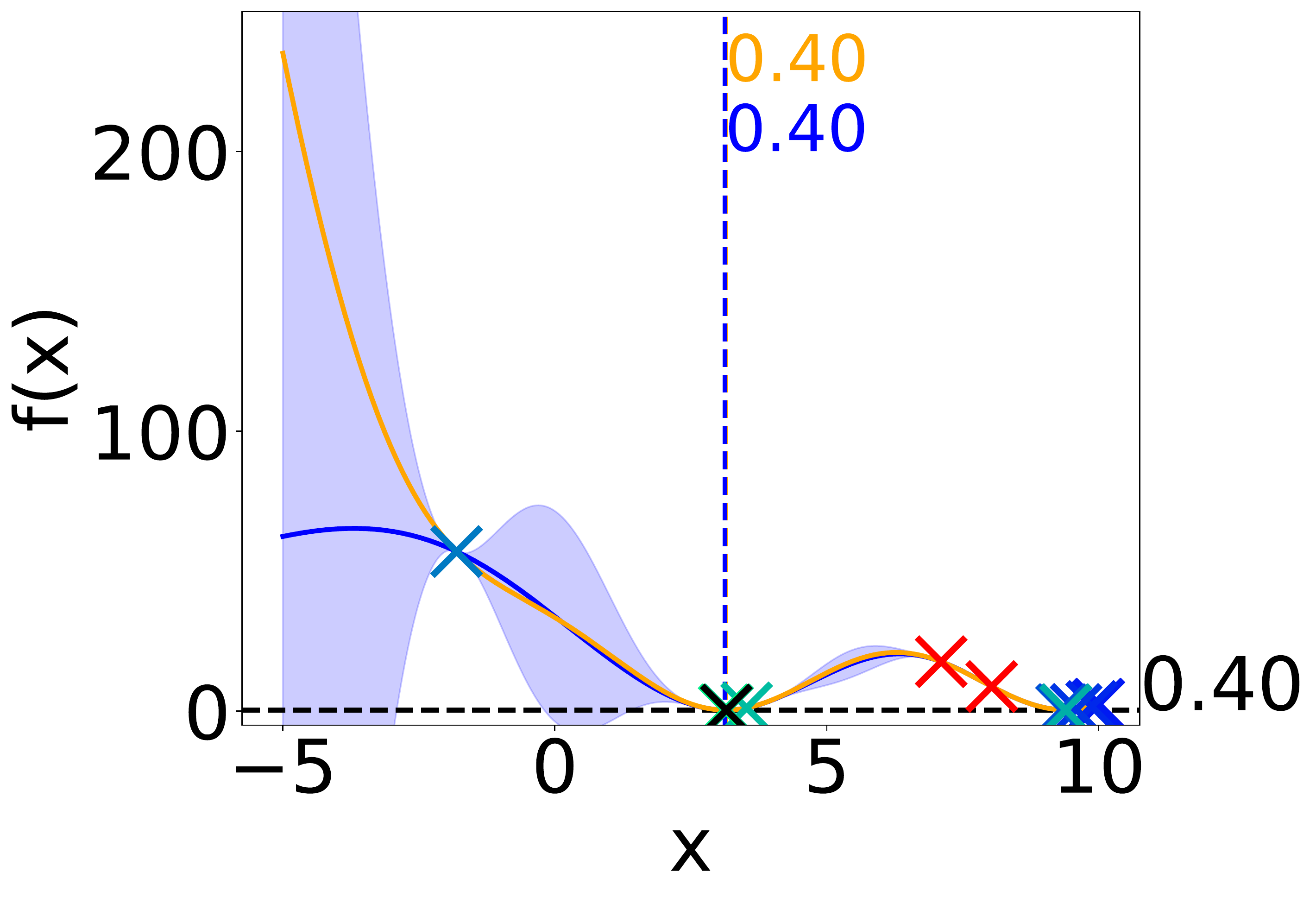}}
\caption{\name on the 1D Branin function. The leftmost column shows 
the exponential \priorstr. The other columns show the model and the log \posteriorstr after 0 (\edit{initialization only}{}), 10, and 20 BO iterations. 
\name forgets the wrong \priorstr on the local optimum and converges to the global optimum.}
\label{fig:wrong_prior_exp}
\end{figure*}

We first show that \name can recover from a misleading \priorstr, 
thanks to our model \edit{$\modelgood$}{} and the $t/\beta$ parameter in the \posteriorstr computation in Eq.~\eqref{eq:posterior}. As BO progresses, the model \edit{$\modelgood$}{} becomes more accurate and receives more weight, guiding optimization away from the wrong \priorstr and towards better values of the function. 

Figure~\ref{fig:wrong_prior_exp} shows \name on the 1D Branin function with an exponential \priorstr. 
Columns (b), (c), and (d) show \name after $D+1=2$ initial samples and $0$, $10$, $20$ BO iterations, respectively. After initialization, as shown in Column (b), the \posteriorstr is nearly identical to the exponential \priorstr and guides \name towards the region of the space on the right, which is towards the local optimum. 
This happens until the model \edit{$\modelgood$}{} becomes certain there will be no more improvement from sampling that region (Columns (c) and (d)). After that, \edit{$\modelgood$}{} guides the \posteriorstr towards exploring regions with high uncertainty. Once the global minimum region is found, the \posteriorstr starts balancing exploiting the global minimum and exploring regions with high uncertainty, as shown in Figure~\ref{fig:wrong_prior_exp}d (bottom). 
Notably, the \posteriorstr after $x > 4$ falls to $0$ in Figure~\ref{fig:wrong_prior_exp}d (top), as the model \edit{$\modelgood$}{} is certain there will be no improvement from sampling the region of the local optimum. 
We provide additional examples of forgetting 
in Appendix~\ref{sec:forgetting.appendix}, and a comparison of \name with misleading \priorsstr, no \priorstr, and correct \priorsstr in Appendix~\ref{sec:misleading}.

\subsection{Comparison Against Strong Baselines} \label{sec:experiments.regret}



We build two
\priorsstr for the optimum in a controlled way and evaluate \name's performance with these different \priorstr strengths. We emphasize that in practice, manual \priorsstr would be based on the domain experts' expertise on their applications;
here, we only use artificial priors to guarantee that our prior is not biased by our own expertise for the benchmarks we used. 
In practice, users will manually define these \priorsstr like in our real-world experiments (Section~\ref{sec:experiments.spatial}).

Our synthetic priors take the form of Gaussian distributions centered near the optimum. 
For each input $x \in \Param$, we inject a prior of the form $\mathcal{N}(\mu_x, \sigma^2_x)$, where $\mu_x$ is sampled from a Gaussian centered at the optimum value $x_{opt}$\footnote{If the optimum for a benchmark is not known, we approximate it using the best value found during previous BO experiments.} for that parameter $\mu_x \sim \mathcal{N}(x_{opt}, \sigma^2_x)$, and $\sigma_x$ is a hyperparameter of our experimental setup determining the prior's strength. For each run of \name, we sample new $\mu_x$'s. This setup provides us with a synthetic prior that is close to the optimum, but not exactly centered at it, and allows us to control the strength of the prior by $\sigma_x$. We use two prior strengths in our experiments: a strong prior, computed with $\sigma_x = 0.01$, and a weak prior, computed with $\sigma_x = 0.1$.

\begin{figure*}[tb]
        \begin{center}
        \subfigure{\includegraphics[width=0.6\linewidth]{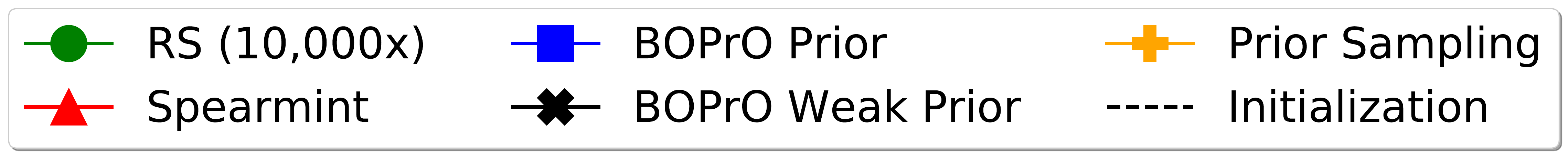}}\\\vspace{-0.3cm}
        \subfigure{\includegraphics[width=0.255\linewidth]{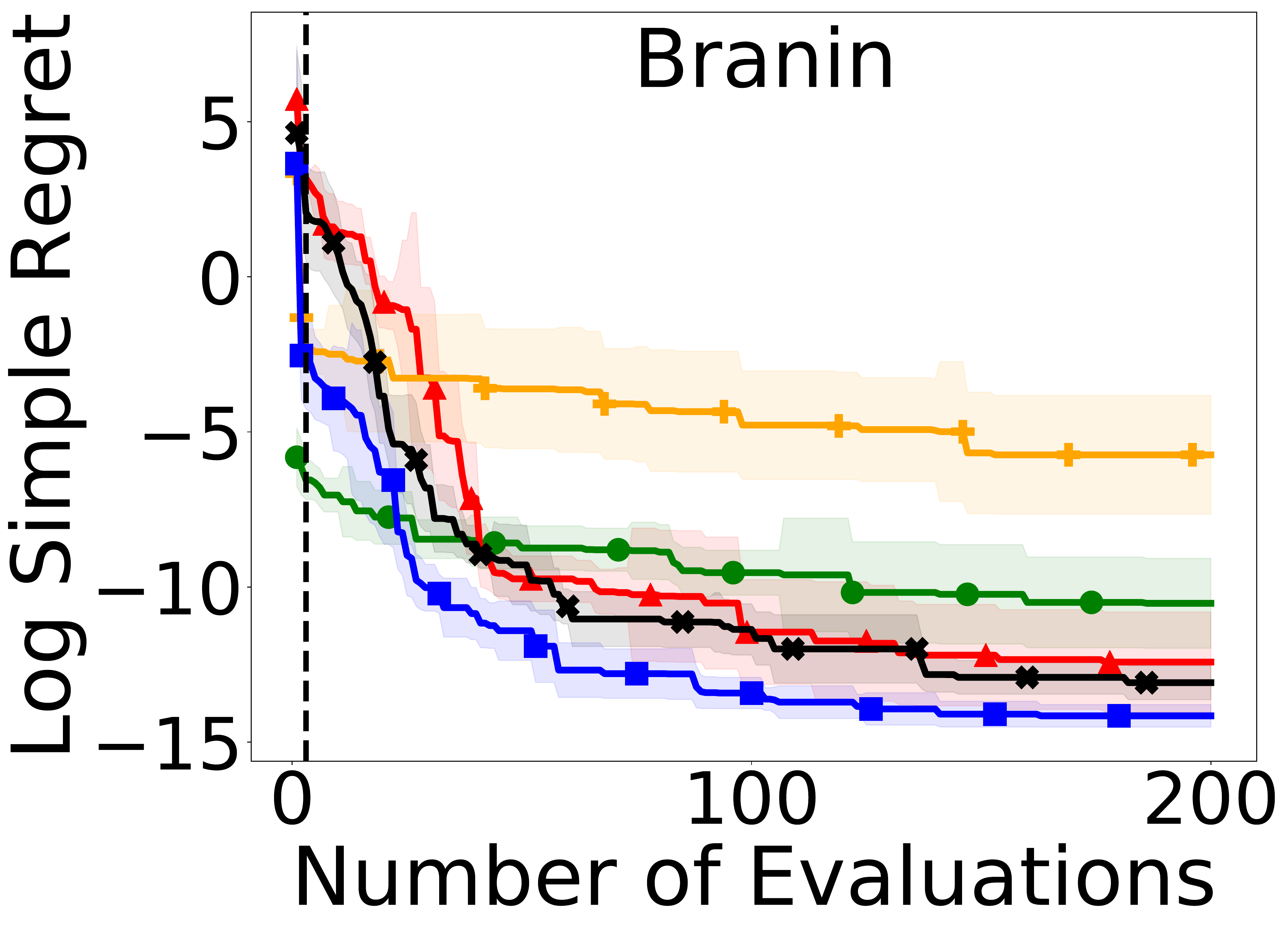}}
        \subfigure{\includegraphics[width=0.238\linewidth]{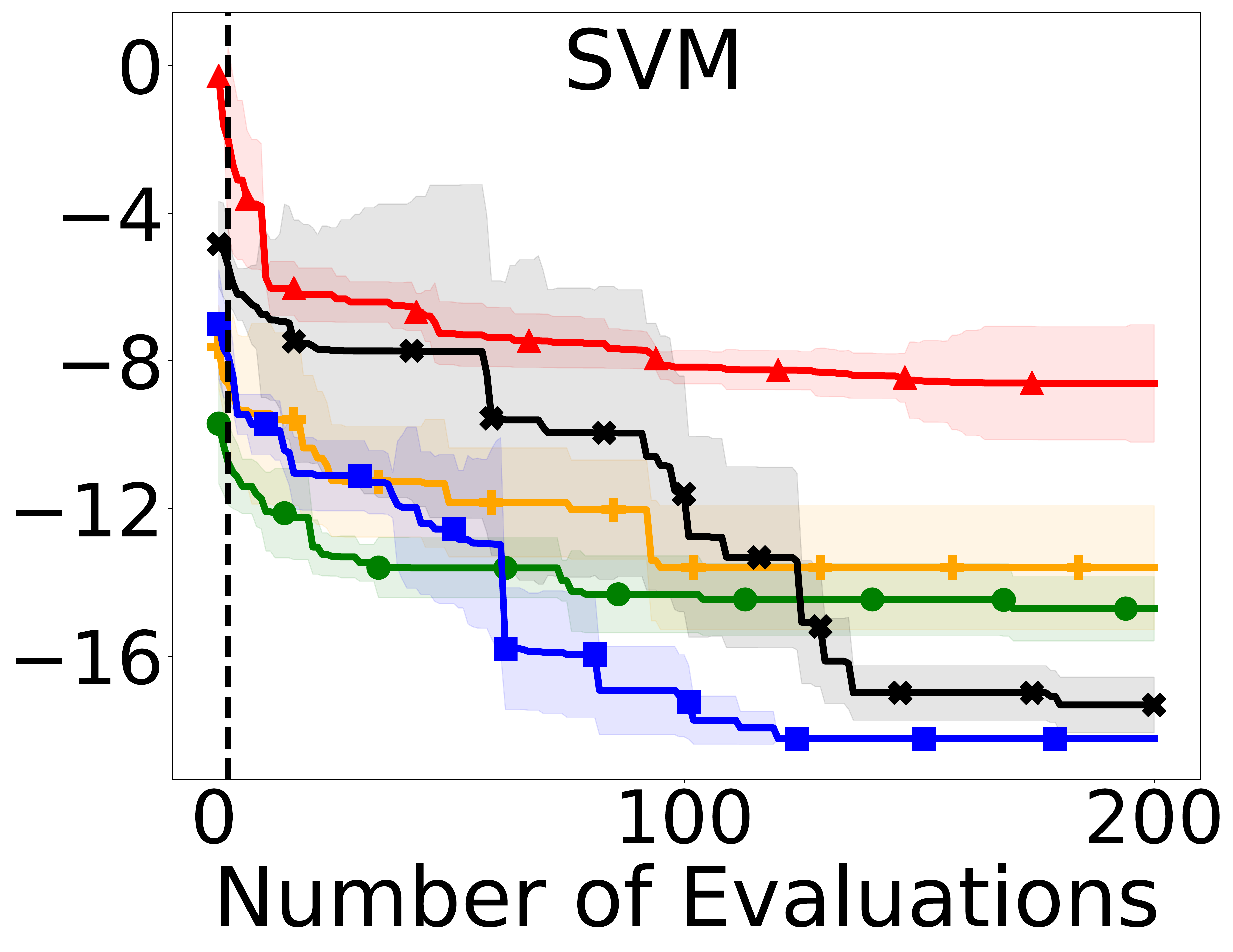}}
        \subfigure{\includegraphics[width=0.226\linewidth]{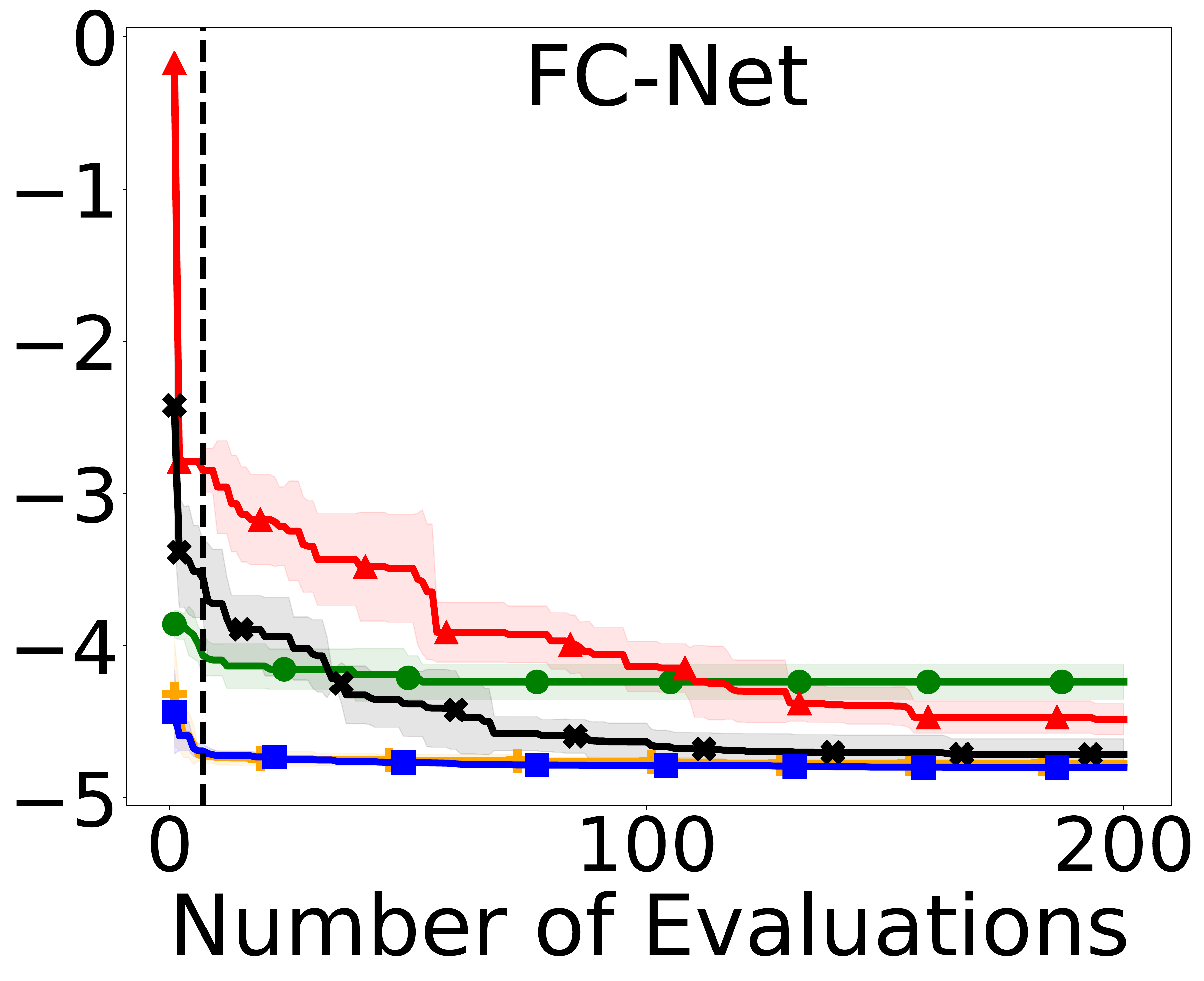}}
        \subfigure{\includegraphics[width=0.237\linewidth]{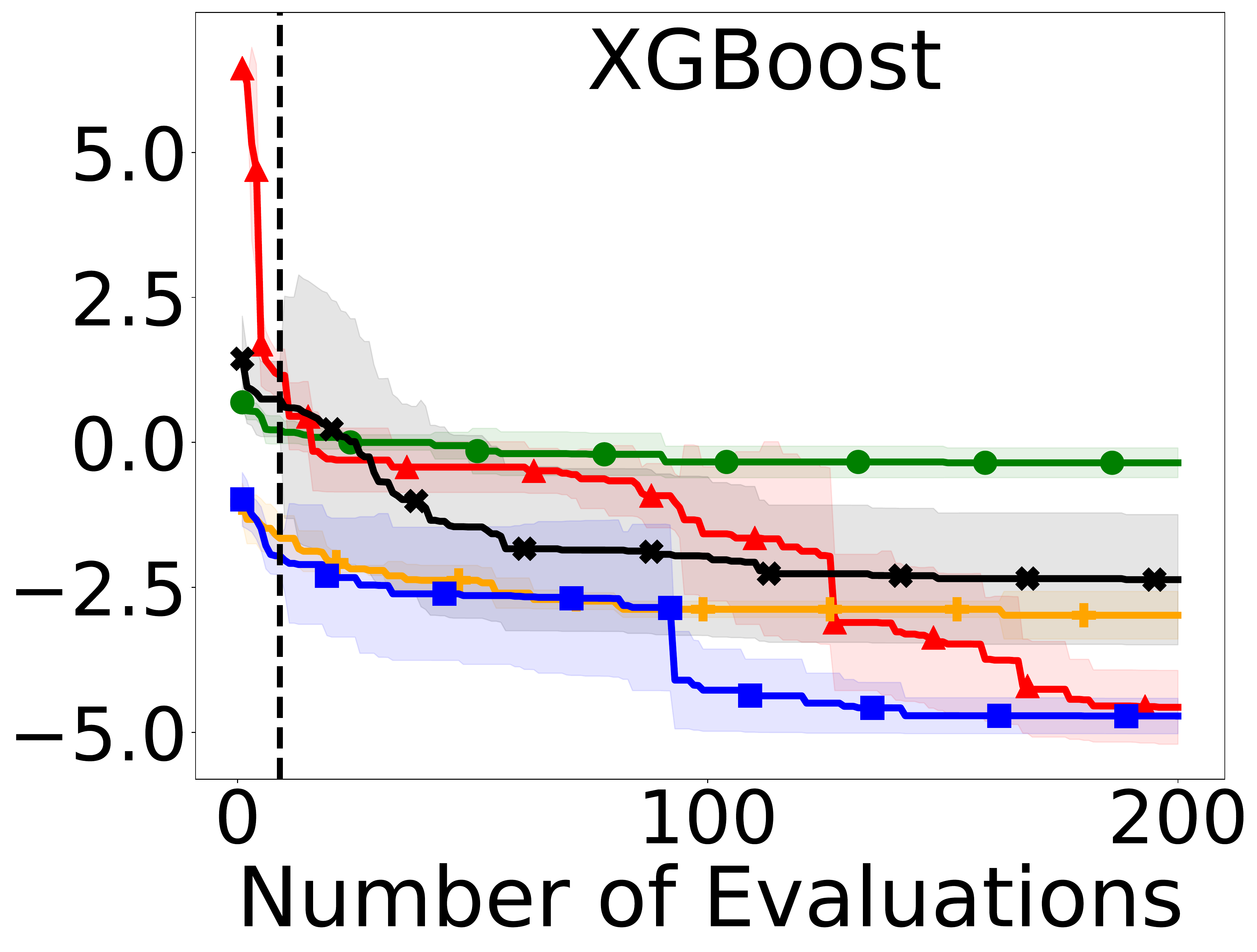}}
    \caption{Log regret comparison of \name with weak and strong \priorsstr, sampling from the strong prior, 10,000$\times$ random search (RS), and Spearmint (mean +/- one std on 5 repetitions). 
    We run each benchmark for 200 iterations.}
    \label{fig:synthetic.regret}
    \end{center}
\end{figure*}

Figure~\ref{fig:synthetic.regret} compares \name to other optimizers using the log simple regret on 5 runs (mean and std error reported) on the synthetic benchmarks.
We compare the results of \name with weak and strong  \priorsstr to $10$,$000\times$ random search (RS, i.e., for each BO sample we draw $10$,$000$ uniform random samples), sampling from the strong \priorstr only, and Spearmint~\cite{snoekLA12}, a well-adopted BO approach using GPs and EI. \postrebuttal{In Appendix~\ref{sec:tpe.appendix}, we also show a comparison of \name with TPE, SMAC\newedit{, and TuRBO~\cite{hutter2011sequential,smac-2017,eriksson2019scalable}.}{}} 
Also, in Appendix~\ref{sec:prior_init.appendix}, we compare \name to other baselines with the same \priorstr initialization and show that the performance of the baselines remains similar.
\name{} with a strong \priorstr for the optimum beats $10$,$000\times$ RS and \name with a weak \priorstr on all benchmarks. 
It also 
outperforms 
the performance of sampling from the strong \priorstr; this is expected because the \priorstr sampling cannot focus on the real location of the optimum.
The two methods are identical during the initialization phase because they both sample from the same \priorstr in that phase. 

\name{} with a strong \priorstr is also more sample efficient and finds better or similar results to Spearmint 
on all benchmarks. 
Importantly, in all our experiments, \name with a good \priorstr consistently shows tremendous speedups in the early phases of the optimization process, requiring on average only 15 iterations to reach the performance that Spearmint reaches after 100 iterations ($6.67\times$ faster). 
Thus, in comparison to other traditional BO approaches, \name makes use of the best of both worlds, leveraging prior knowledge and efficient optimization based on BO. 





\subsection{The Spatial Use-case} \label{sec:experiments.spatial}

\begin{figure*}[t]
    \begin{center}
        \subfigure{\includegraphics[width=0.9\linewidth]{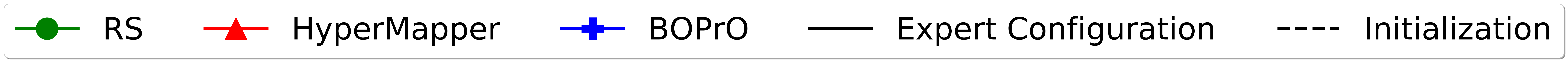}}\\\vspace{-0.3cm}
        \subfigure{\includegraphics[width=0.31\linewidth]{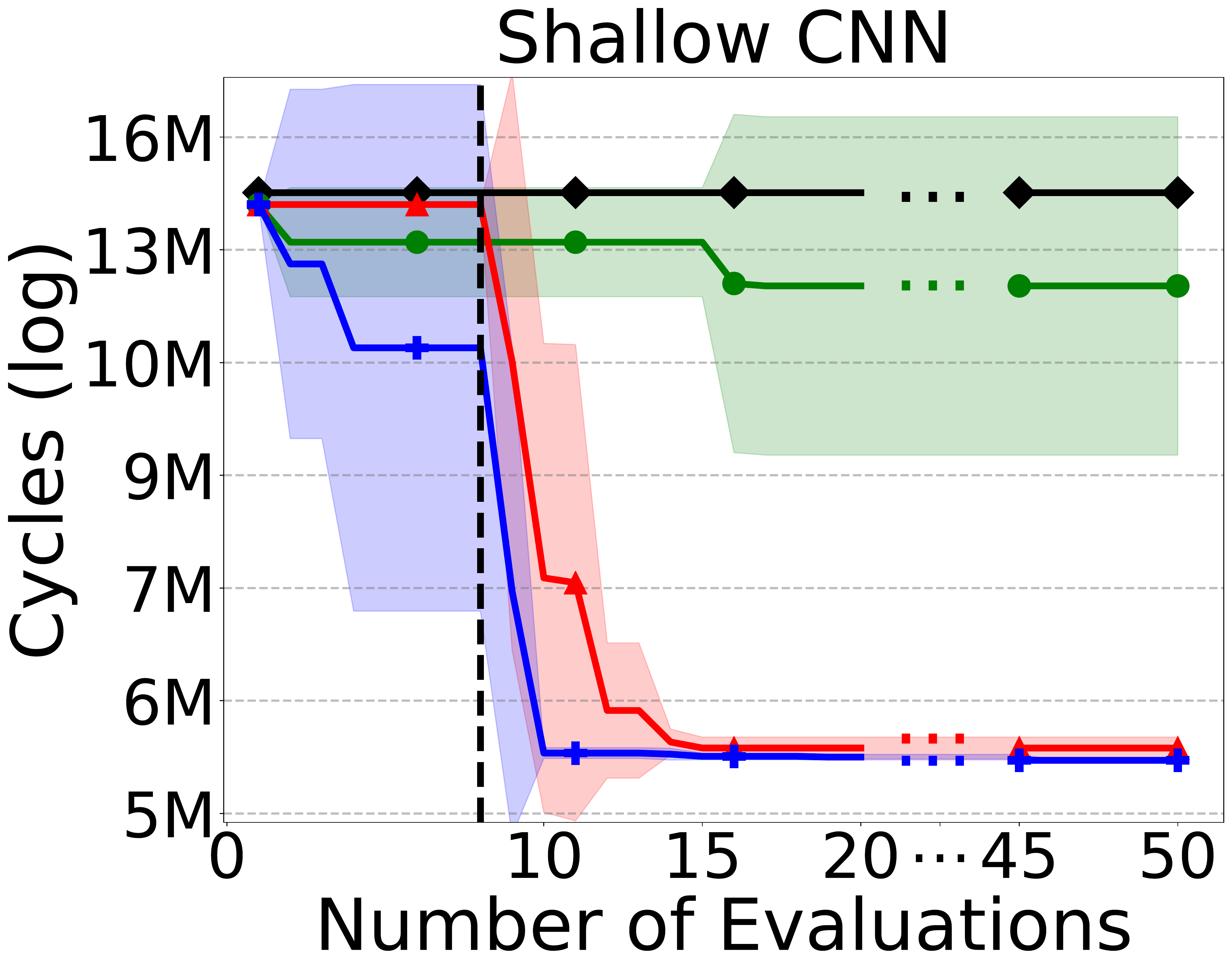}}
        \subfigure{\includegraphics[width=0.3\linewidth]{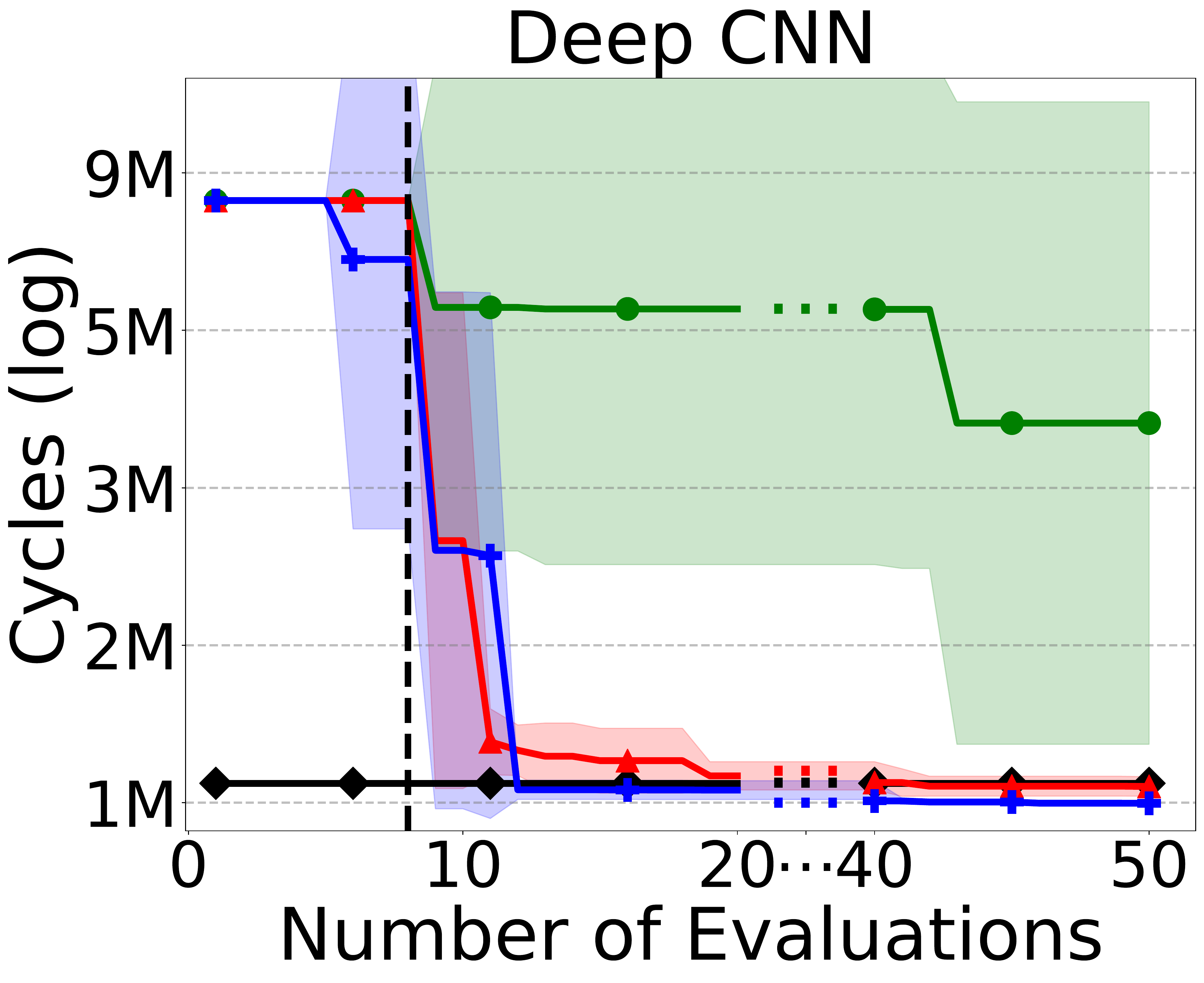}}
        \subfigure{\includegraphics[width=0.363\linewidth]{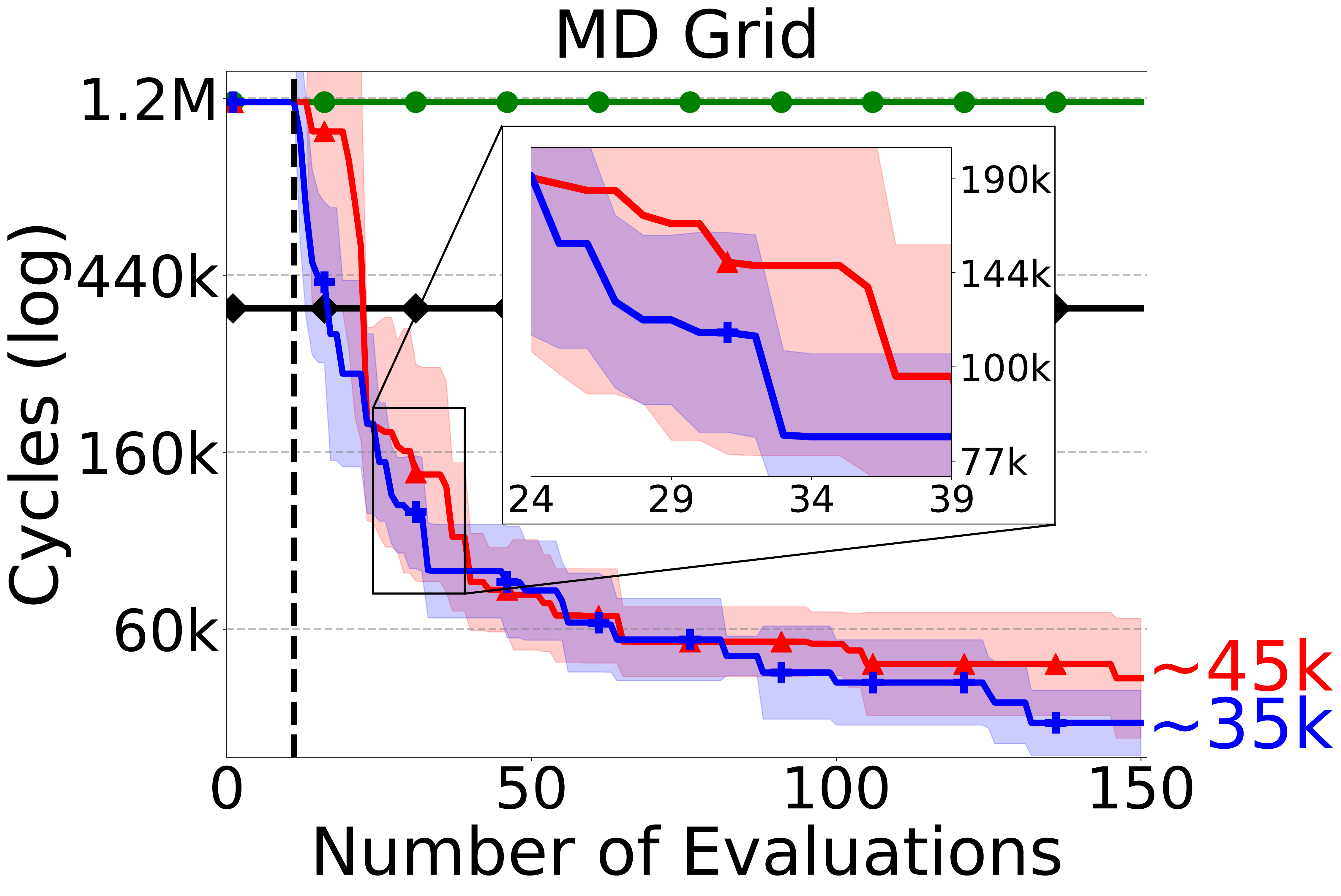}}
    \end{center}
    \caption{Log regret comparison of random search (RS), HyperMapper, \name, and manual optimization on Spatial. The line and shaded regions show mean and std after 5 repetitions. Vertical lines denote the end of the initialization phase. 
    }
    \label{fig:spatial.regret}
\end{figure*}


We next apply \name to the \texttt{Spatial}~\cite{koeplinger2018} real-world application. \texttt{Spatial} is a programming language and corresponding compiler for the design of application accelerators,
i.e., FPGAs. We apply \name to three \texttt{Spatial} benchmarks, namely, 7D shallow and deep CNNs, and a 10D molecular dynamics grid application (MD Grid). We compare the performance of \name to RS, manual optimization, and \HM~\cite{nardi18hypermapper}, the current state-of-the-art BO solution for \texttt{Spatial}. For a fair comparison between \name and \HM, since \HM uses RFs as its surrogate model, here, we also use RFs in \name. The manual optimization and the \priorstr for \name were provided by an unbiased \texttt{Spatial} developer, who is not an author of this paper. 
The \priorsstr were provided once and kept unchanged for the whole project. More details on the setup, including the \priorsstr used, are presented in Appendix~\ref{sec:spatial.appendix}. 

Figure~\ref{fig:spatial.regret} shows the log regret on the  \texttt{Spatial} benchmarks. \name vastly outperforms RS in all benchmarks; notably, RS does not improve over the default configuration in MD Grid. \name is also able to leverage the expert's \priorstr and outperforms the expert's configuration in all benchmarks ($2.68\times$, $1.06\times$, and $10.4\times$ speedup for shallow CNN, deep CNN, and MD Grid, respectively). 
In the MD Grid benchmark, \name achieves better performance than \HM in the early stages of optimization (up to $1.73\times$ speedup between iterations 25 and 40, see the plot inset), and achieves better final performance ($1.28\times$ speedup). For context, this is a significant improvement in the FPGA field, where a 10\% improvement could qualify for acceptance in a top-tier conference. 
In the CNN benchmarks, \name converges to the minima regions faster than \HM ($1.58\times$ and $1.4\times$ faster for shallow and deep, respectively). 
Thus, \name leverages both the expert's prior knowledge and BO to provide a new state of the art for \texttt{Spatial}.


\section{Related Work} \label{sec:rw}

TPE by Bergstra \etal~\cite{bergstra2011algorithms}, the default optimizer in the popular HyperOpt package~\cite{bergstra2013hyperopt}, supports limited hand-designed \priorsstr in the form of normal or log-normal distributions. We make three technical contributions that make \name more flexible than TPE. First, we generalize over the TPE approach by allowing more flexible priors; second, \name is agnostic to the probabilistic model used, allowing the use of more sample-efficient models than TPE's kernel density estimators (e.g., we use GPs and RFs in our experiments); and third, \name is inspired by Bayesian models that give more importance to the data as iterations progress. We also show that \name outperforms HyperOpt's TPE in Appendix~\ref{sec:tpe.appendix}.

In parallel work, Li \etal~\cite{li2020incorporating} allow users to specify \priorsstr via a probability distribution.
Their two-level approach samples a number of configurations by maximizing samples from a GP posterior and then chooses the configuration with the highest \priorstr as the next to evaluate. In contrast, \name leverages the information from the \priorstr more directly; is agnostic to the probabilistic model used, which is important for applications with many discrete variables like our real-world application, where RFs outperform GPs; and provably recovers from misspecified priors{, while in their approach the prior never gets washed out.}

The work of Ramachandran \etal\cite{Ramachandran2020warping} also supports priors in the form of probability distributions. Their work uses the probability integral transform to warp the search space, stretching regions where the prior has high probability, and shrinking others. Once again, compared to their approach, \name is agnostic to the probabilistic model used and directly controls the balance between prior and model via the $\beta$ hyperparameter. Additionally, BOPrO’s probabilistic model is fitted independently from the prior, which ensures it is not biased by the prior, while their approach fits the model to a warped version of the space, transformed by the prior, making it difficult to recover from misleading priors.

Black-box optimization tools, such as SMAC~\cite{hutter2011sequential} or iRace~\cite{lopez2016irace} also support simple hand-designed \priorsstr, e.g. log-transformations. However, these are not properly reflected in the predictive models and both cannot explicitly recover from bad \priorsstr. 

Oh \etal\cite{oh2018bock} and Siivola \etal\cite{siivola2018correcting} propose structural priors for high-dimensional problems. They assume that users always place regions they expect to be good at the center of the search space and then develop BO approaches that favor configurations near the center. However, this is a rigid assumption about optimum locality, which does not allow users to freely specify their priors.
Similarly, Shahriari \etal\cite{shahriari2016unbounded} focus on unbounded search spaces. The priors in their work are not about good regions of the space, but rather a regularization function that penalizes configurations based on their distance to the center of the user-defined search space. The priors are automatically derived from the search space and not provided by users.

Our work also relates to meta-learning for BO~\cite{vanschoren2019automl}, where BO is applied to many similar optimization problems in a sequence such that knowledge about the general problem structure can be exploited in future optimization problems. In contrast to meta-learning, \name allows human experts to explicitly specify their \priorsstr. 
Furthermore, \name does not depend on any meta-features~\cite{feurer2015init}
and incorporates the human's prior instead of information gained from different experiments~\cite{lindauer2018a}.


\section{Conclusions and Future Work} \label{sec:conc}
We have proposed a novel BO variant, \name, that allows users to inject their expert knowledge into the optimization in the form of \priorsstr about which parts of the input space will yield the best performance. 
These are different than standard \priorsstr over functions which are much less intuitive for users. 
So far, BO failed to leverage the experience of domain experts, not only causing inefficiency
but also driving users away from applying BO approaches because they could not exploit their years of knowledge in optimizing their black-box functions.
\name addresses this issue and we therefore expect it to facilitate the adoption of BO. 
We showed that \name is $6.67\times$ more sample efficient than strong BO baselines,
and $10$,$000\times$ faster than random search, on a common suite of benchmarks and achieves a new state-of-the-art performance on a real-world hardware design application. We also showed that \name converges faster 
and robustly recovers from misleading \priorsstr. 

In future work, we will study how our approach can 
be used to leverage prior knowledge from meta-learning. Bringing these two worlds together will likely boost the performance of BO even further.

\section{Acknowledgments}

We thank Matthew Feldman for \spatial support. Luigi Nardi and Kunle Olukotun were supported in part by affiliate members and other supporters of the Stanford DAWN project — Ant Financial, Facebook, Google, Intel, Microsoft, NEC, SAP, Teradata, and VMware. Luigi Nardi was also partially supported by the Wallenberg AI, Autonomous Systems and Software Program (WASP) funded by the Knut and Alice Wallenberg Foundation. Artur Souza and Leonardo B. Oliveira were supported by CAPES, CNPq, and FAPEMIG. Frank Hutter acknowledges support by the European Research Council (ERC) under the European Union Horizon 2020 research and innovation programme through grant no.\ 716721. The computations were also enabled by resources provided by the Swedish National Infrastructure for Computing (SNIC) at LUNARC partially funded by the Swedish Research Council through grant agreement no. 2018-05973.

\bibliographystyle{splncs04}
\bibliography{ref}
\clearpage

\appendix

\section{\Priorstr Forgetting Supplementary Experiments} \label{sec:forgetting.appendix}

\begin{figure*}[t]
\centering
\subfigure{\includegraphics[width=0.203\textwidth]{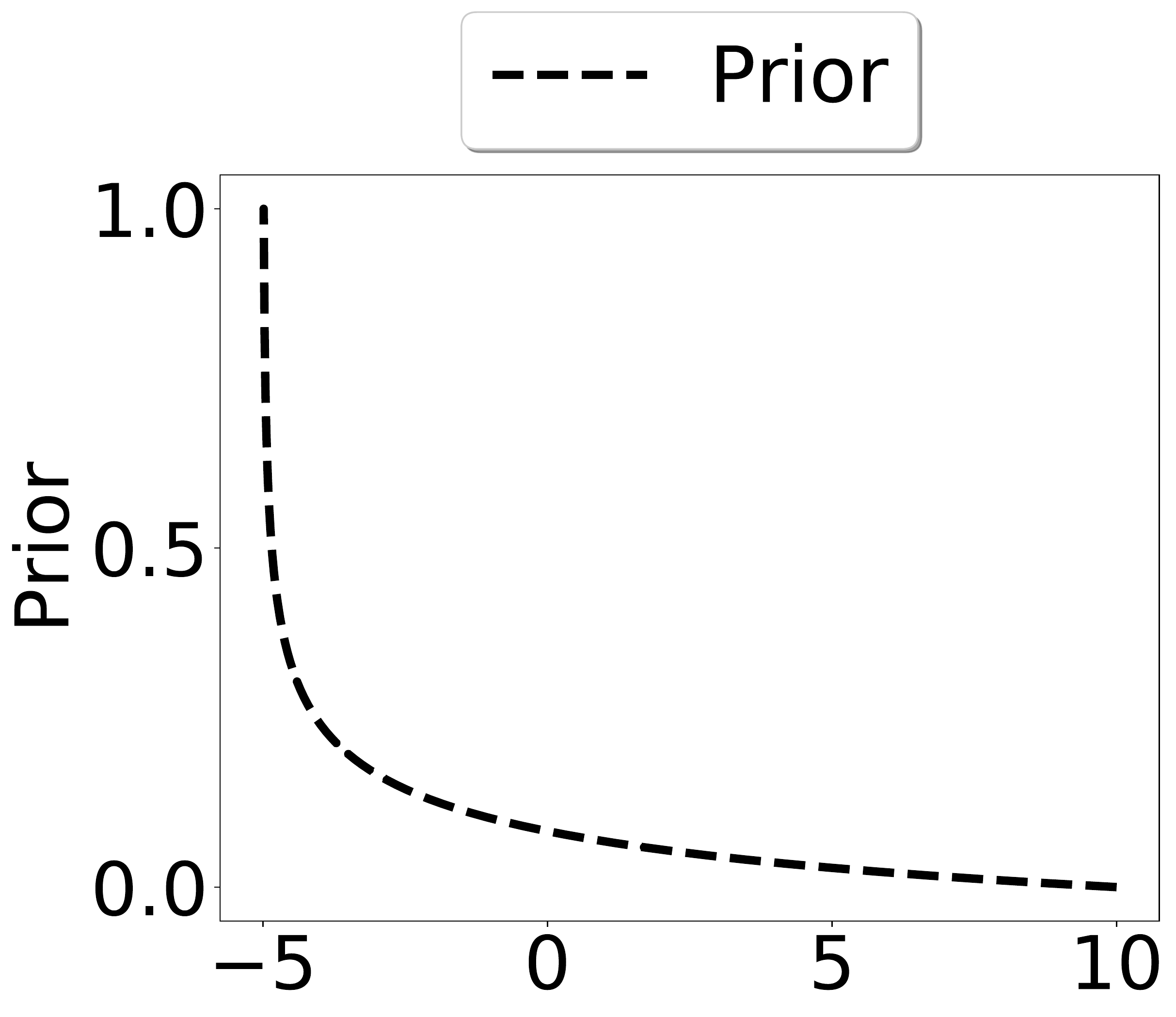}}
\subfigure{\includegraphics[width=0.245\textwidth]{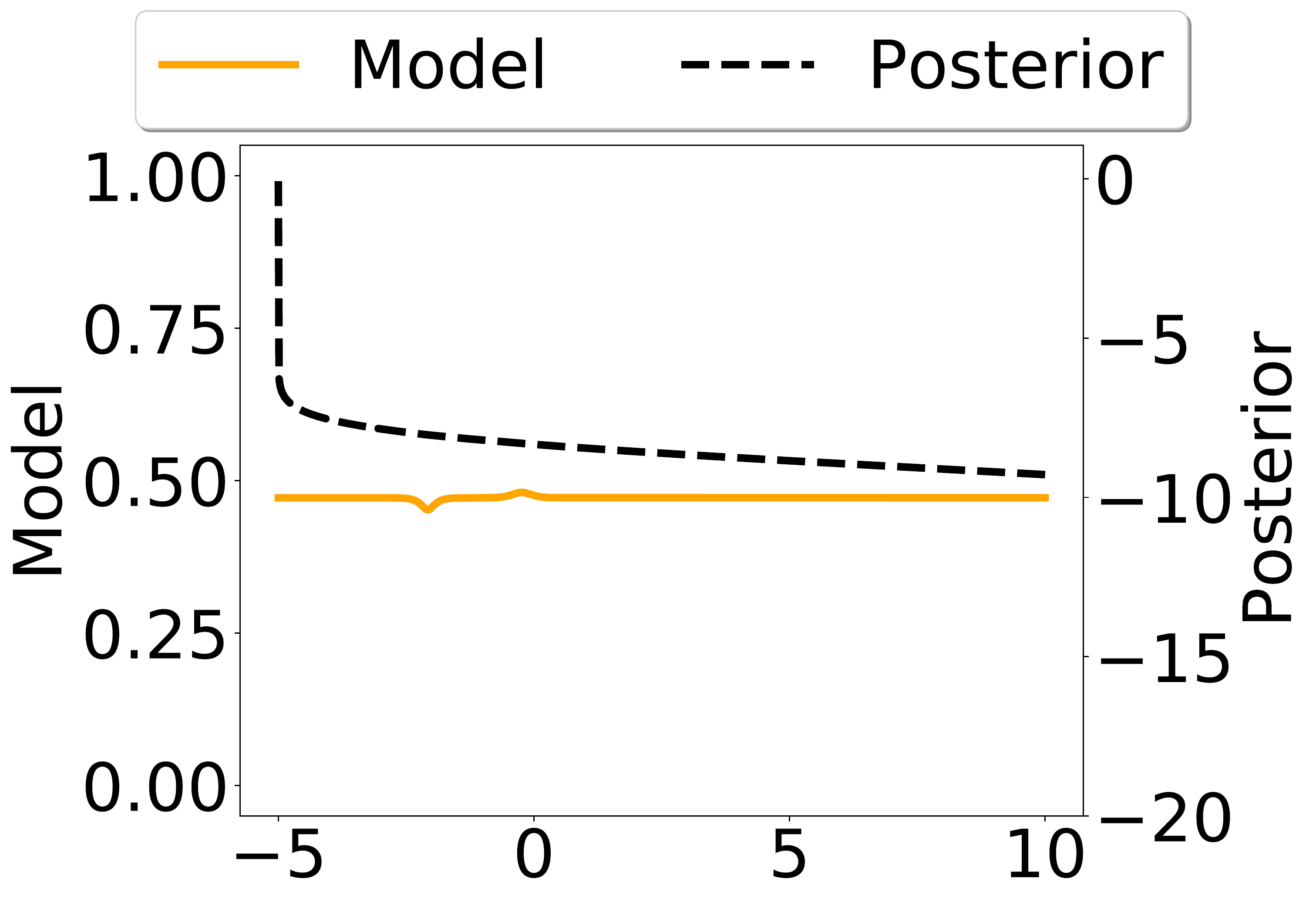}}
\subfigure{\includegraphics[width=0.245\textwidth]{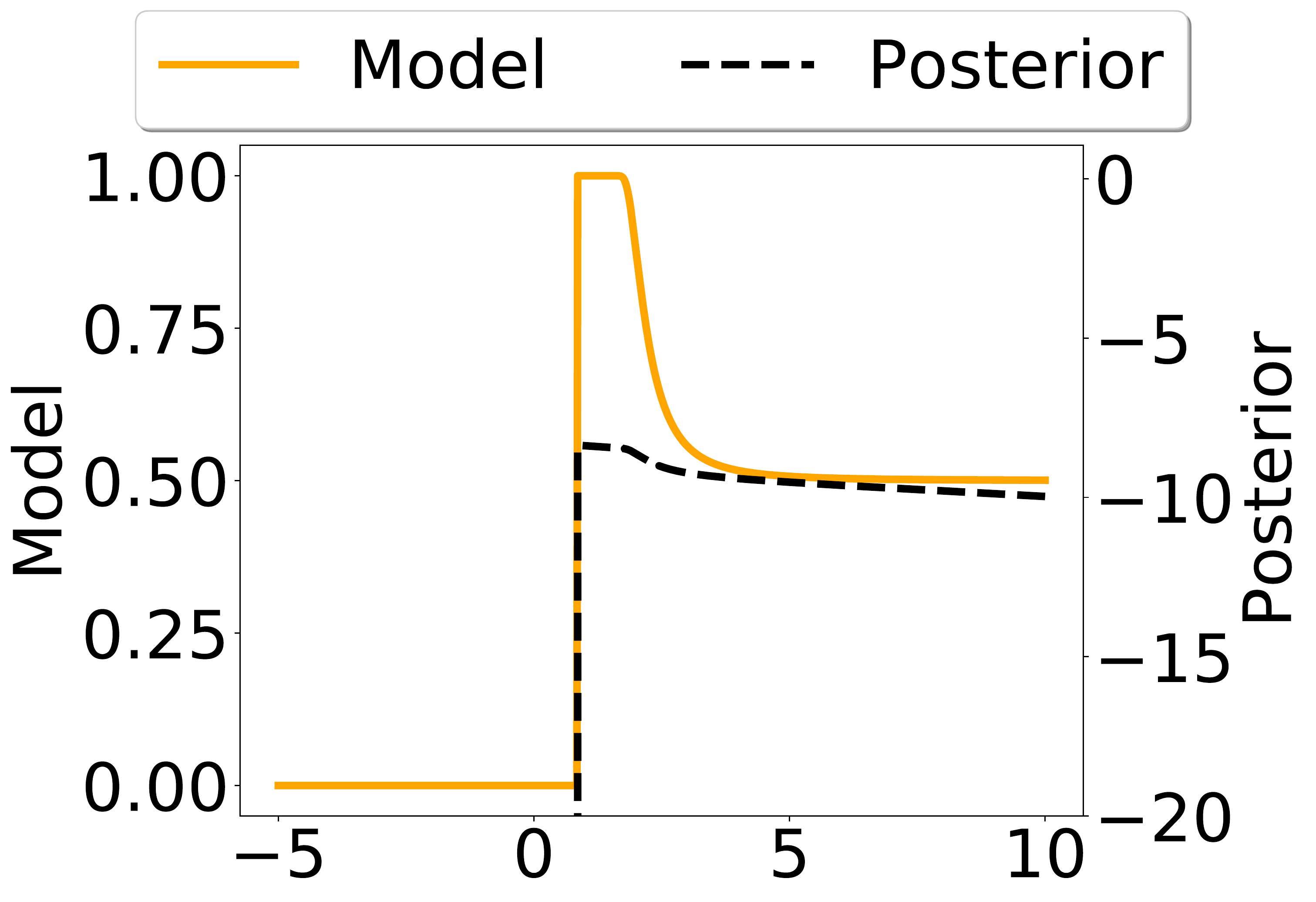}}
\subfigure{\includegraphics[width=0.245\textwidth]{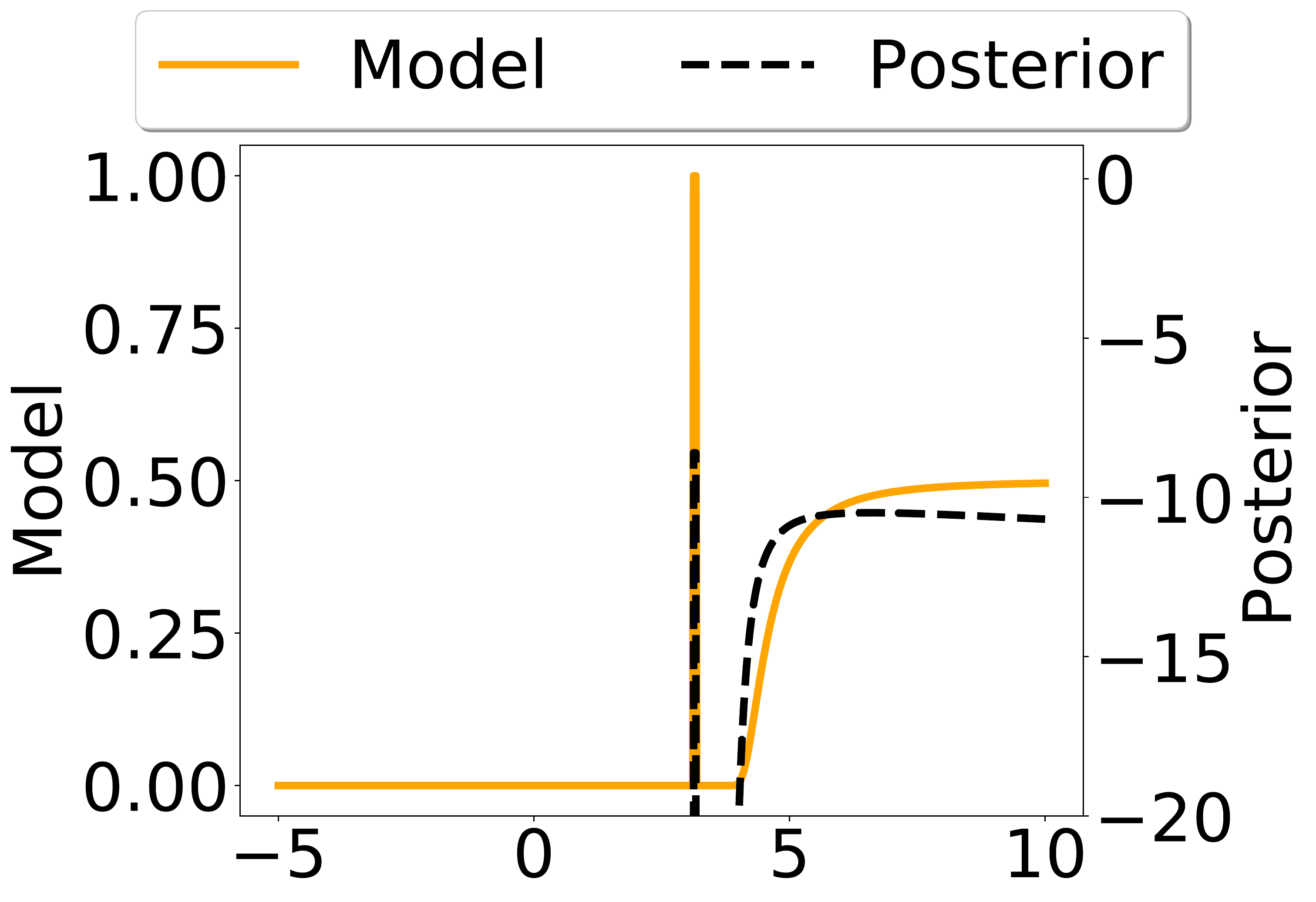}}
\centering\subfigure{\includegraphics[width=0.9\textwidth]{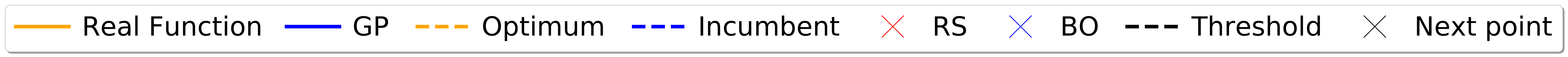}}\\\vspace{-0.3cm}
\setcounter{subfigure}{0}

\subfigure[No samples]{\includegraphics[width=0.22\textwidth]{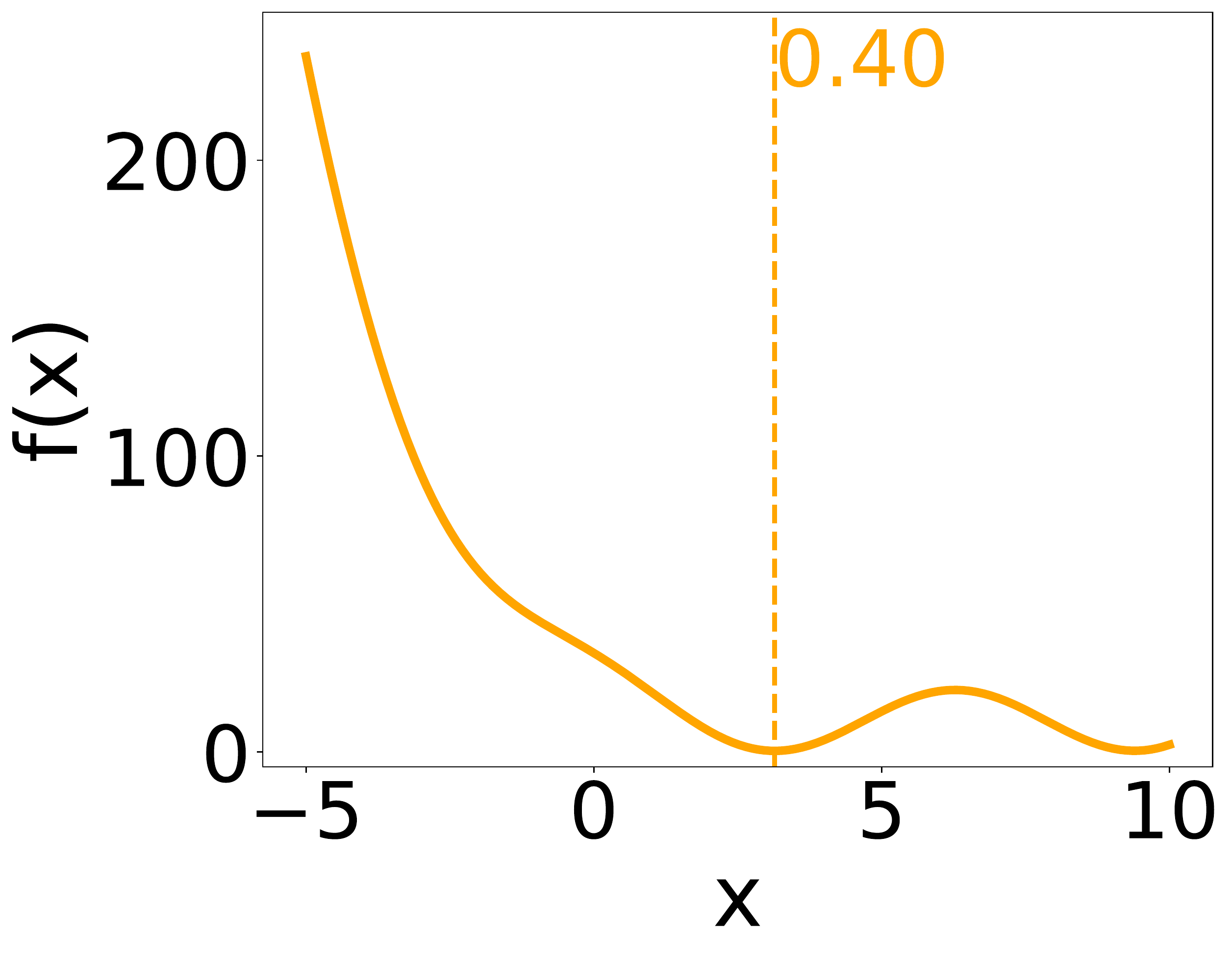}}
\subfigure[0 BO iterations]{\includegraphics[width=0.255\textwidth]{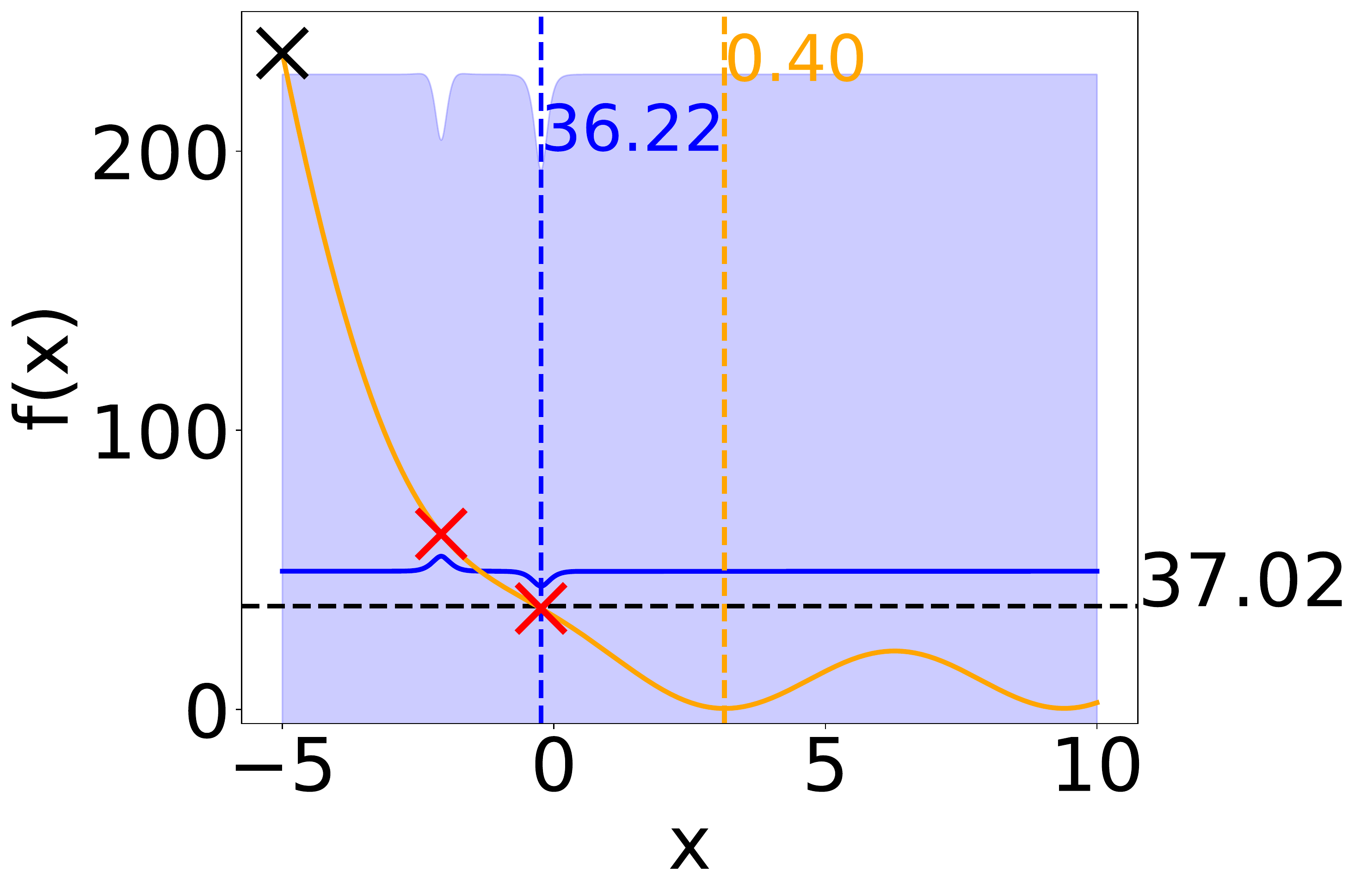}}
\subfigure[10 BO iterations]{\includegraphics[width=0.253\textwidth]{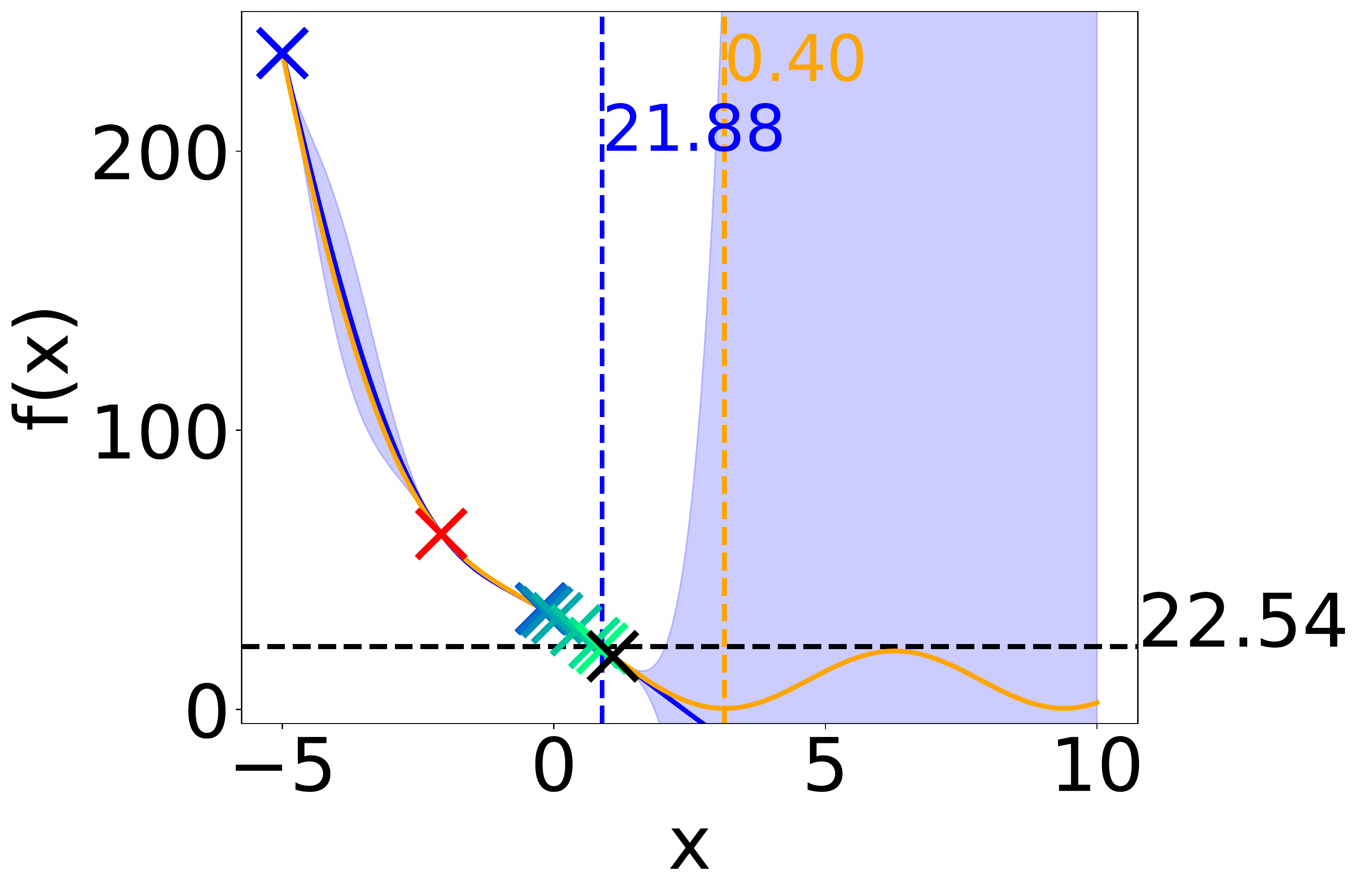}}
\subfigure[20 BO iterations]{\includegraphics[width=0.246\textwidth]{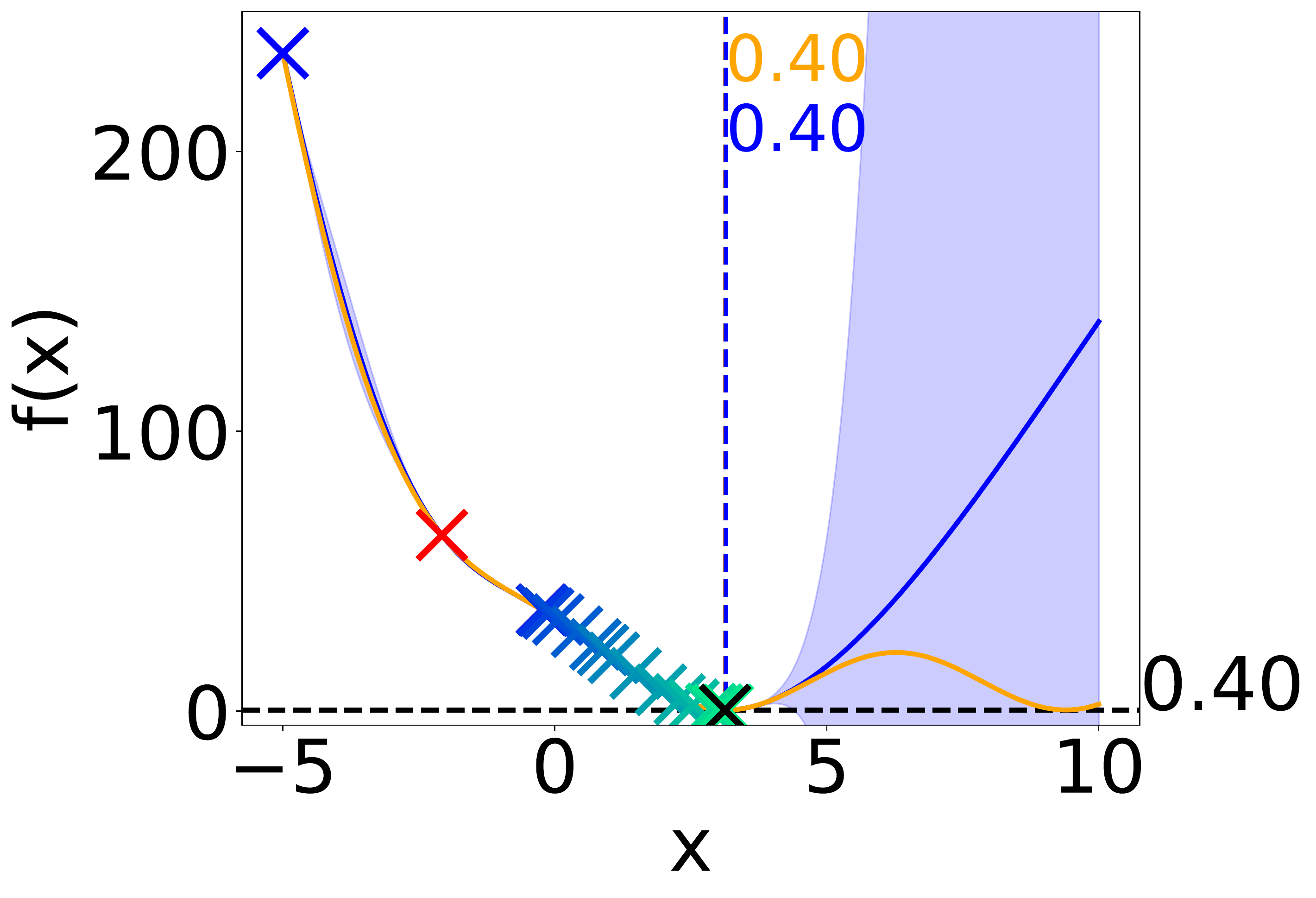}}
\caption{\name on the 1D Branin function with a decay \priorstr. The leftmost column shows the log pseudo-posterior before any samples are evaluated, in this case, the pseudo-posterior is equal to the decay  \priorstr. The other columns show the model and pseudo-posterior after 0 (only random samples), 10, and 20 BO iterations. 2 random samples are used to initialize the GP model.}
\label{fig:wrong_prior_decay}
\end{figure*}

In this section, we show additional evidence that \name can recover from wrongly defined \priorsstr so to complement section~\ref{sec:experiments.prior}.
Figure~\ref{fig:wrong_prior_decay} shows \name on the 1D Branin function as in Figure~\ref{fig:wrong_prior_exp} but with a decay \priorstr. Column (a) of Figure~\ref{fig:wrong_prior_decay} shows the decay \priorstr and the 1D Branin function.
This \priorstr emphasizes the wrong belief that the optimum is likely located on the left side around $\mathrm{x} = -5$ while the optimum is located at the orange dashed line. Columns (b), (c), and (d) of Figure~\ref{fig:wrong_prior_decay} show \name on the 1D Branin after $D+1=2$ initial samples and $0$, $10$, and $20$ BO iterations, respectively. In the beginning of BO, as shown in column (b), the \posteriorstr is nearly identical to the \priorstr and guides \name towards the left region of the space. As more points are sampled, the model becomes more accurate and starts guiding the \posteriorstr away from the wrong \priorstr (column (c)). Notably, the \posteriorstr before $\mathrm{x} = 0$ falls to $0$, as the predictive model is certain there will be no improvement from sampling this region. After 20 iterations, \name finds the optimum region, despite the poor start (column (d)). The peak in the \posteriorstr in column (d) shows \name will continue to exploit the optimum region as it is not certain if the exact optimum has been found. The \posteriorstr is also high in the high uncertainty region after $x = 4$, showing \name will explore that region after it finds the optimum.

\begin{figure}[tb]
\centering
\subfigure{\includegraphics[width=0.8\columnwidth]{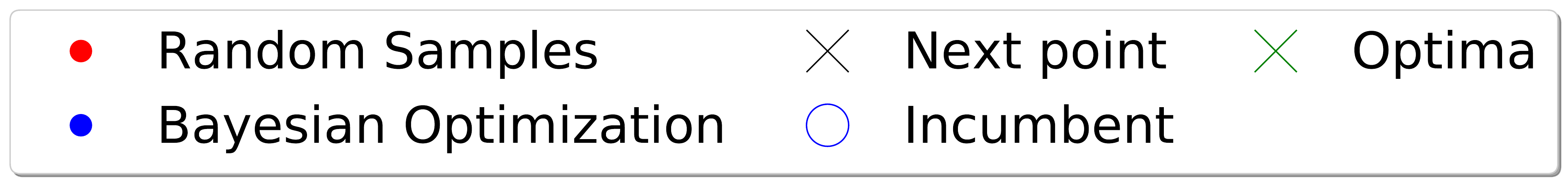}}\\\vspace{-0.3cm}
\subfigure{\includegraphics[width=0.45\columnwidth]{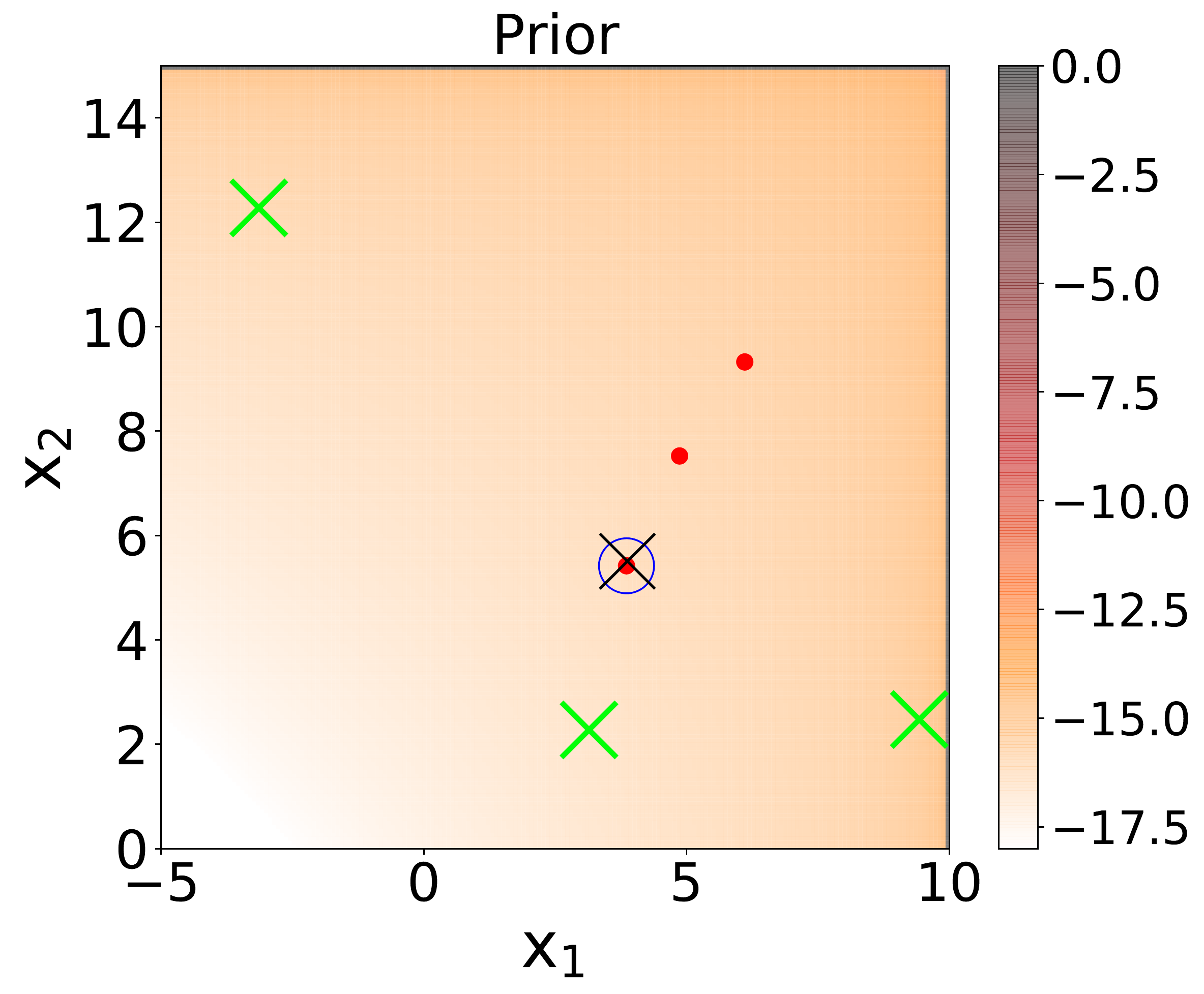}}
    \subfigure{\includegraphics[width=0.45\columnwidth]{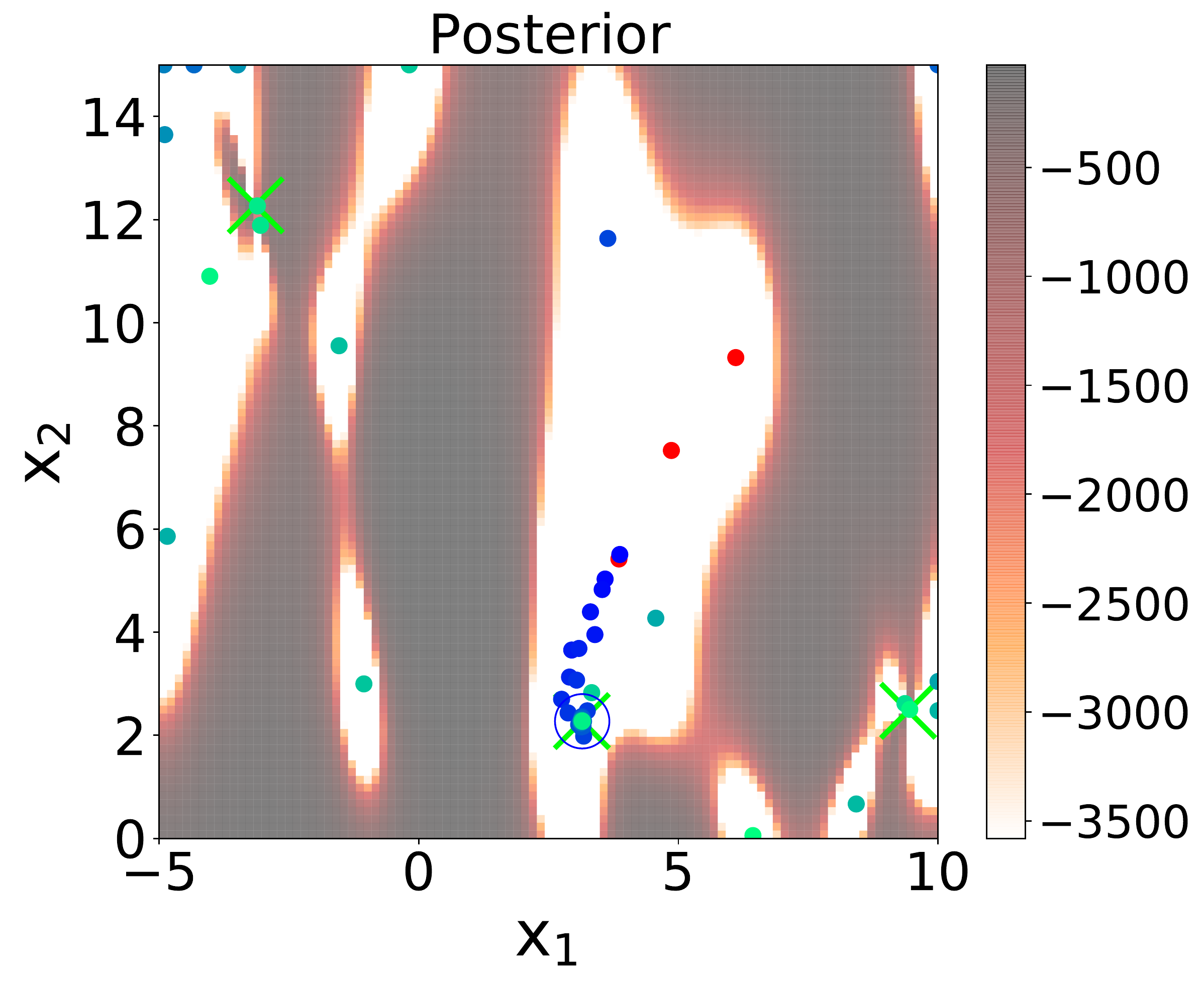}}
\caption{\name on the Branin function with exponential \priorsstr for both dimensions. (a) shows the log \posteriorstr before any samples are evaluated, in this case, the \posteriorstr is equal to the \priorstr; the green crosses are the optima. (b) shows the result of optimization after 3 initialization samples drawn from the \priorstr at random and 50 BO iterations. The dots in (b) show the points explored by \name, with greener points denoting later iterations. The colored heatmap shows the log of the \posteriorstr $g(\param)$.}
\label{fig:wrong_prior_branin}
\end{figure}

Figure~\ref{fig:wrong_prior_branin} shows \name on the standard 2D Branin function. We use exponential \priorsstr for both dimensions, which guides optimization towards a region with only poor performing high function values. \ref{fig:wrong_prior_branin}a  shows the \priorstr and \ref{fig:wrong_prior_branin}b shows optimization results after $D+1=3$ initialization samples and 50 BO iterations. Note that, once again, optimization begins near the region incentivized by the \priorstr, but moves away from the \priorstr and towards the optima as BO progresses. After 50 BO iterations, \name finds all three optima regions of the Branin.

\section{Mathematical Derivations} \label{sec:proofs}

\subsection{EI Derivation} \label{sec:proof.ei}

Here, we provide a full derivation of Eq.~\eqref{eq:priorsop.ei2}:

\begin{align} 
    EI_{f_{\gamma}}(\bm{x}) &:=\int_{-\infty}^{\infty} \max(f_{\gamma} - y, 0) p(y|\bm{x}) dy \nonumber  
    = \int_{-\infty}^{f_{\gamma}} (f_{\gamma} - y)\frac{p(\bm{x}|y) p(y)}{p(\bm{x})} dy.
\end{align}

As defined  in Section~\ref{sec:priorop.model}, $p(y < f_{\gamma}) = \gamma$ and $\gamma$ is a quantile of the observed objective values $\{y^{(i)}\}$. 
Then $p(\bm{x})=
\int_{\mathbb{R}}p(\bm{x}|y)p(y)dy =
\gamma g(\bm{x}) + (1-\gamma) b(\bm{x})$, where $g(\param)$ and $b(\param)$ are the posteriors introduced in Section~\ref{sec:priorop.posterior}. 
Therefore 
\begin{align} 
\int_{-\infty}^{f_{\gamma}} (f_{\gamma} - y) p(\bm{x}|y) p(y) dy &= g(\bm{x}) \int_{-\infty}^{f_{\gamma}} (f_{\gamma} - y) p(y) dy \nonumber\\ 
&= \gamma f_{\gamma} g(\bm{x}) - g(\bm{x}) \int_{-\infty}^{f_{\gamma}} y p(y) dy,
\end{align} 
so that finally
\begin{align} \label{eq:ei.proof.final}
EI_{f_{\gamma}}(\bm{x}) &=\frac{\gamma f_{\gamma} g(\bm{x}) - g(\bm{x}) \int_{-\infty}^{f_{\gamma}} y p(y) dy}{\gamma g(\bm{x}) + (1-\gamma) b(\bm{x})} \propto \left( \gamma + \dfrac{b(\param)}{g(\param)}(1 - \gamma) \right)^{-1}.
\end{align} 

\subsection{Proof of Proposition~\ref{prop:priorsop1}} \label{sec:proof.priorsop1}

Here, we provide the proof of Proposition~\ref{prop:priorsop1}:

\begin{align} \label{eq:priorsop1}
& \lim_{t\to\infty} \argmax_{\param \in \Param} EI_{f_{\gamma}}(\bm{x}) \\
& = \lim_{t\to\infty} \argmax_{\param \in \Param} \int_{-\infty}^{f_{\gamma}} (f_{\gamma} - y) p(\bm{x}|y) p(y) dy \\
& = \lim_{t\to\infty} \argmax_{\param \in \Param} g(\bm{x}) \int_{-\infty}^{f_{\gamma}} (f_{\gamma} - y) p(y) dy \\
& = \lim_{t\to\infty} \argmax_{\param \in \Param} \left(\gamma f_{\gamma} g(\bm{x}) - g(\bm{x}) \int_{-\infty}^{f_{\gamma}} y p(y) dy\right) \\
& = \lim_{t\to\infty} \argmax_{\param \in \Param} \frac{\gamma f_{\gamma} g(\bm{x}) - g(\bm{x}) \int_{-\infty}^{f_{\gamma}} y p(y) dy}{\gamma g(\bm{x}) + (1-\gamma) b(\bm{x})} 
\end{align}
\edit{which, from Eq.~\eqref{eq:ei.proof.final}, is equal to:}{}
\begin{align} \label{eq:priorsop1.2}
& = \lim_{t\to\infty} \argmax_{\param \in \Param} \left( \gamma + \dfrac{b(\param)}{g(\param)}(1 - \gamma) \right)^{-1}
\end{align}
\edit{we can take Eq.~\eqref{eq:priorsop1.2} to the power of $\dfrac{1}{t}$ without changing the expression, since the argument that maximizes EI does not change:}{}
\begin{align}
& = \lim_{t\to\infty} \argmax_{\param \in \Param} \left( \gamma + \dfrac{b(\param)}{g(\param)}(1 - \gamma) \right)^{-\frac{1}{t}}
\end{align}
\edit{substituting $g(x)$ and $b(x)$ using their definitions in Section~\ref{sec:priorop.posterior}:}{}
\begin{align}
& = \lim_{t\to\infty} \argmax_{\param \in \Param} \left( \gamma + \dfrac{\priorbad \modelbad^{\tfrac{t}{\beta}}}{\priorgood \modelgood^{\tfrac{t}{\beta}}}(1 - \gamma) \right)^{-\frac{1}{t}}\\
& = \lim_{t\to\infty} \argmax_{\param \in \Param} \left(\dfrac{\priorbad \modelbad^{\tfrac{t}{\beta}}}{\priorgood \modelgood^{\tfrac{t}{\beta}}}(1 - \gamma) \right)^{-\frac{1}{t}}\\
& = \lim_{t\to\infty} \argmax_{\param \in \Param} \left(\dfrac{\priorbad}{\priorgood}\right)^{-\frac{1}{t}}  \left(\dfrac{\modelbad^{\frac{t}{\beta}}}{\modelgood^{\frac{t}{\beta}}}\right)^{-\frac{1}{t}}  \left(1- \gamma\right)^{-\frac{1}{t}}\\
& = \lim_{t\to\infty} \argmax_{\param \in \Param} \left(\dfrac{\priorbad}{\priorgood}\right)^{-\frac{1}{t}}  \left(\dfrac{\modelbad}{\modelgood}\right)^{-{\frac{1}{\beta}}}  \left(1- \gamma\right)^{-\frac{1}{t}}\\
& = \argmax_{\param \in \Param} \left(\dfrac{\modelbad}{\modelgood} \right)^{-\frac{1}{\beta}}\\
& =  \argmax_{\param \in \Param} \left(\dfrac{1-\modelgood}{\modelgood} \right)^{-\frac{1}{\beta}}\\
& =  \argmax_{\param \in \Param} \left(\dfrac{1}{\modelgood} - 1 \right)^{-\frac{1}{\beta}}\\
& = \argmax_{\param \in \Param} \left(\modelgood\right)^{\frac{1}{\beta}}\\
& = \argmax_{\param \in \Param} \modelgood
\end{align}

This shows that as iterations progress, the model grows more important. If \name is run long enough, the prior washes out and \name only trusts the probabilistic model. Since $\modelgood$ is the Probability of Improvement (PI) on the probabilistic model $p(y|\param)$ then, in the limit, maximizing the acquisition function $EI_{f_{\gamma}}(\param)$ is equivalent to maximizing the PI acquisition function on the probabilistic model $p(y|\param)$. 
In other words, for high values of $t$, \name converges to standard BO with a PI acquisition function.


\section{Experimental Setup} \label{sec:experimental_setup.appendix}

\begin{table}[tb]
    \begin{center}
        \caption{Search spaces for our synthetic benchmarks. For the Profet benchmarks, we report the original ranges and whether or not a log scale was used. 
        }
        \label{tab:benchmarks.space}
        \begin{tabular}{l|lll}
        \textbf{Benchmark} & \textbf{Parameter name} & \textbf{Parameter values} & \textbf{Log scale} \\ \hline
        Branin             & $x_1$                   & $[-5, 10]$                & -                  \\
                           & $x_2$                   & $[0, 15]$                 & -                  \\ \hline
        SVM                & C                       & $[e^{-10}, e^{10}]$       & \checkmark         \\
                           & $\gamma$                & $[e^{-10}, e^{10}]$       & \checkmark         \\ \hline
        FCNet              & learning rate           & $[10^{-6}, 10^{-1}]$      & \checkmark         \\
                           & batch size              & $[8, 128]$                & \checkmark         \\
                           & units layer 1           & $[16, 512]$               & \checkmark         \\
                           & units layer 2           & $[16, 512]$               & \checkmark         \\
                           & dropout rate l1         & $[0.0, 0.99]$             & -                  \\
                           & dropout rate l2         & $[0.0, 0.99]$             & -                  \\ \hline
        XGBoost            & learning rate           & $[10^{-6}, 10^{-1}]$      & \checkmark         \\
                           & gamma                   & $[0,2]$                   & -                  \\
                           & L1 regularization       & $[10^{-5}, 10^{3}]$       & \checkmark         \\
                           & L2 regularization       & $[10^{-5}, 10^{3}]$       & \checkmark         \\
                           & number of estimators    & $[10, 500]$               & -                  \\
                           & subsampling             & $[0.1, 1]$                & -                  \\
                           & maximum depth           & $[1, 15]$                 & -                  \\
                           & minimum child weight    & $[0, 20]$                 & -                  \\ \hline
        \end{tabular}
    \end{center}
\end{table}

We use a combination of publicly available implementations for our predictive models. For our Gaussian Process (GP) model, we use GPy's~\cite{gpy2014} GP implementation with the Matérn5/2 kernel. We use different length-scales for each input dimensions, learned via Automatic Relevance Determination (ARD)~\cite{neal2012bayesian}. For our Random Forests (RF), we use scikit-learn's RF implementation~\cite{scikit-learn}. We set the fraction of features per split to $0.5$, the minimum number of samples for a split to $5$ and disable bagging. We also adapt our RF implementation to use the same split selection approach as Hutter \etal\cite{hutter13rf}. 

For our constrained Bayesian Optimization (cBO) approach, we use scikit-learn's RF classifier, trained on previously explored configurations, to predict the probability of a configuration being feasible. We then weight our EI acquisition function by this probability of feasibility, as proposed by Gardner \etal\cite{gardner2014bayesian}. We normalize our EI acquisition function before considering the probability of feasibility, to ensure both values are in the same range. This cBO implementation is used in the Spatial use-case as in Nardi \etal\cite{nardi18hypermapper}.

For all experiments, we set the model weight hyperparameter to $\beta = 10$ and the model quantile to $\gamma = 0.05$, see Appendices~\ref{sec:beta.appendix} and \ref{sec:gamma.appendix}. Before starting the main BO loop, \name is initialized by random sampling $D+1$ points from the \priorstr, where $D$ is the number of input variables. We use the public implementation of Spearmint\footnote{https://github.com/HIPS/Spearmint}, which by default uses 2 random samples for initialization.
We normalize our synthetic \priorsstr before computing the \posteriorstr, to ensure they are in the same range as our model. We also implement interleaving which randomly samples a point  to explore during BO with a $10\%$ chance.

\begin{table}[tb]
    \begin{center}
        \caption{Search space, \priorsstr, and expert configuration for the Shallow CNN application. The default value for each parameter is shown in bold.}
        \label{tab:spatial.shallow}
        \begin{tabularx}{\linewidth}{|l|l|l|l|L|}
            \hline
            \textbf{Parameter name} & \textbf{Type}           & \textbf{Values}                & \textbf{Expert} & \textbf{\Priorstr} \\ \hline
            LP                 & Ordinal                 & [\textbf{1}, 4, 8, 16, 32]   & 16      & [0.4, 0.065, 0.07, 0.065, 0.4] \\ \hline
            P1                 & Ordinal                 & [\textbf{1}, 2, 3, 4]        & 1       & [0.1, 0.3, 0.3, 0.3]           \\ \hline
            SP                 & Ordinal                 & [\textbf{1}, 4, 8, 16, 32]   & 16      & [0.4, 0.065, 0.07, 0.065, 0.4] \\ \hline
            P2                 & Ordinal                 & [\textbf{1}, 2, 3, 4]        & 4       & [0.1, 0.3, 0.3, 0.3]           \\ \hline
            P3                 & Ordinal                 & [\textbf{1}, 2, ..., 31, 32] & 1       & [0.1, 0.1, 0.033, 0.1, 0.021, 0.021, 
                                                                                                  0.021, 0.1, 0.021, 0.021, 0.021, 0.021, 
                                                                                                  0.021, 0.021, 0.021, 0.021, 0.021, 
                                                                                                  0.021, 0.021, 0.021, 0.021, 0.021,  
                                                                                                  0.021, 0.021, 0.021, 0.021, 0.021, 
                                                                                                  0.021, 0.021, 0.021, 0.021, 0.021] \\ \hline
            P4                 & Ordinal                 & [\textbf{1}, 2, ..., 47, 48] & 4       & [0.08, 0.0809, 0.0137, 0.1, 0.0137, 
                                                                                                0.0137, 0.0137, 0.1, 0.0137, 0.0137,
                                                                                                0.0137, 0.05, 0.0137, 0.0137, 0.0137,
                                                                                                0.0137, 0.0137, 0.0137, 0.0137, 0.0137,
                                                                                                0.0137, 0.0137, 0.0137, 0.0137, 0.0137,
                                                                                                0.0137, 0.0137, 0.0137, 0.0137, 0.0137,
                                                                                                0.0137, 0.0137, 0.0137, 0.0137, 0.0137,
                                                                                                0.0137, 0.0137, 0.0137, 0.0137, 0.0137,
                                                                                                0.0137, 0.0137, 0.0137, 0.0137, 0.0137,
                                                                                                0.0137, 0.0137, 0.0137]  \\ \hline
            x276               & Categorical             & [false, \textbf{true}]       & true    & [0.1, 0.9]     \\\hline
        \end{tabularx}
    \end{center}
\end{table}

We optimize our EI acquisition function using a combination of a multi-start local search and CMA-ES~\cite{hansen96cmaes}. Our multi-start local search is similar to the one used in SMAC~\cite{hutter2011sequential}. Namely, we start local searches on the 10 best points evaluated in previous BO iterations, on the 10 best performing points from a set of $10$,$000$ random samples, on the 10 best performing points from $10$,$000$ random samples drawn from the \priorstr, and on the mode of the prior. To compute the neighbors of each of these 31 total points, we normalize the range of each parameter to $[0, 1]$ and randomly sample four neighbors from a truncated Gaussian centered at the original value and with standard deviation $\sigma = 0.1$. For CMA-ES, we use the public implementation of pycma~\cite{hansen2019pycma}. We run pycma with two starting points, one at the incumbent and one at the mode of the prior. For both initializations we set $\sigma_0 = 0.2$. We only use CMA-ES for our continuous search space benchmarks.


We use four synthetic benchmarks in our experiments.

\topic{Branin} The Branin function is a well-known synthetic benchmark for optimization problems~\cite{dixon1978global}. The Branin function has two input dimensions and three global minima. 

\topic{SVM} A hyperparameter-optimization benchmark in 2D based on Profet~\cite{klein2019meta}. This benchmark is generated by a generative meta-model built using a set of SVM classification models trained on 16 OpenML tasks. The benchmark has two input parameters, corresponding to SVM hyperparameters.

\topic{FC-Net} A hyperparameter and architecture optimization benchmark in 6D based on Profet. The FC-Net benchmark is generated by a generative meta-model built using a set of feed-forward neural networks trained on the same 16 OpenML tasks as the SVM benchmark. The benchmark has six input parameters corresponding to network hyperparameters.

\topic{XGBoost} A hyperparameter-optimization benchmark in 8D based on Profet. The XGBoost benchmark is generated by a generative meta-model built using a set of XGBoost regression models in 11 UCI datasets. The benchmark has eight input parameters, corresponding to XGBoost hyperparameters.

The search spaces for each benchmark are summarized in Table~\ref{tab:benchmarks.space}. For the Profet benchmarks, we report the original ranges and whether or not a log scale was used. However, in practice, Profet's generative model transforms the range of all hyperparameters to a linear $[0, 1]$ range. We use Emukit's public implementation for these benchmarks~\cite{emukit2019}.

\begin{table}[tb]
    \begin{center}
       \caption{Search space, \priorsstr, and expert configuration for the Deep CNN application. The default value for each parameter is shown in bold.}
        \label{tab:spatial.deep}
       \begin{tabularx}{\linewidth}{|l|l|l|l|L|}
        \hline
            \textbf{Parameter name} & \textbf{Type}           & \textbf{Values}                & \textbf{Expert} & \textbf{\Priorstr} \\ \hline
            LP                 & Ordinal                 & [\textbf{1}, 4, 8, 16, 32]   & 8 & [0.4, 0.065, 0.07, 0.065, 0.4] \\ \hline
            P1                 & Ordinal                 & [\textbf{1}, 2, 3, 4]        & 1 & [0.4, 0.3, 0.2, 0.1]           \\ \hline
            SP                 & Ordinal                 & [\textbf{1}, 4, 8, 16, 32]   & 8 & [0.4, 0.065, 0.07, 0.065, 0.4] \\ \hline
            P2                 & Ordinal                 & [\textbf{1}, 2, 3, 4]        & 2 & [0.4, 0.3, 0.2, 0.1]           \\ \hline
            P3                 & Ordinal                 & [\textbf{1}, 2, ..., 31, 32] & 1 & [0.04, 0.01, 0.01, 0.1, 0.01, 
                                                                                               0.01, 0.01, 0.1, 0.01, 0.01,
                                                                                               0.01, 0.01, 0.01, 0.01, 0.01,
                                                                                               0.2, 0.01, 0.01, 0.01, 0.01,
                                                                                               0.01, 0.01, 0.01, 0.1, 0.01, 
                                                                                               0.01, 0.01, 0.01, 0.01, 0.01,
                                                                                               0.01, 0.2] \\ \hline
            P4                 & Ordinal                 & [\textbf{1}, 2, ..., 47, 48] & 4 & [0.05, 0.005, 0.005, 0.005, 
                                                                                               0.005, 0.005, 0.005, 0.13,
                                                                                               0.005, 0.005, 0.005, 0.005,
                                                                                               0.005, 0.005, 0.005, 0.2, 
                                                                                               0.005, 0.005, 0.005, 0.005,
                                                                                               0.005, 0.005, 0.005, 0.11,
                                                                                               0.005, 0.005, 0.005, 0.005,
                                                                                               0.005, 0.005, 0.005, 0.2,
                                                                                               0.005, 0.005, 0.005, 0.005,
                                                                                               0.005, 0.005, 0.005, 0.005,
                                                                                               0.005, 0.005, 0.005, 0.005,
                                                                                               0.005, 0.005, 0.005, 0.1]  \\ \hline
            x276               & Categorical             & [false, \textbf{true}]       & true & [0.1, 0.9]     \\\hline
        \end{tabularx}
    \end{center}
\end{table}

\section{Spatial Real-world Application} \label{sec:spatial.appendix}

\spatial~\cite{koeplinger2018} is a programming language and corresponding compiler for the design of application accelerators on reconfigurable architectures, e.g. field-programmable gate arrays (FPGAs). These reconfigurable architectures are a type of logic chip that can be reconfigured via software to implement different applications. \spatial provides users with a high-level of abstraction for hardware design, so that they can easily design their own applications on FPGAs. It allows users to specify parameters that do not change the behavior of the application, but impact the runtime and resource-usage (e.g., logic units) of the final design. During compilation, the \spatial compiler estimates the ranges of these parameters and estimates the resource-usage and runtime of the application for different parameter values. These parameters can then be optimized during compilation in order to achieve the design with the fastest runtime. We fully integrate \name as a pass in \spatial's compiler, so that \spatial can automatically use \name for the optimization during compilation. This enables \spatial to seamlessly call \name during the compilation of any new application to guide the search towards the best design on an application-specific basis.


\begin{table}[tb]
        \caption{Search space, \priorsstr, and expert configuration for the MD Grid application. The default value for each parameter is shown in bold.}
        \label{tab:spatial.mdgrid}
        \resizebox{\textwidth}{!}{%
        \begin{tabular}{|l|l|l|l|l|}
        \hline
            \textbf{Parameter name} & \textbf{Type} & \textbf{Values}                & \textbf{Expert} & \textbf{\Priorstr} \\ \hline
            loop\_grid0\_z     & Ordinal       & [\textbf{1}, 2, ..., 15, 16]   & 1 & [0.2, 0.1, 0.05, 0.05, 0.05, \\
            &&&&                                                                     0.05, 0.05, 0.05, 0.05, 0.05, \\
            &&&&                                                                     0.05, 0.05, 0.05, 0.05, 0.05, 
                                                                                     0.05] \\ \hline
            loop\_q            & Ordinal       & [\textbf{1}, 2, ..., 31, 32]   & 8 & [0.08, 0.08, 0.02, 0.1, 0.02,   \\
            &&&&                                                                     0.02, 0.02, 0.1, 0.02, 0.02, \\
            &&&&                                                                     0.02, 0.02, 0.02, 0.02, 0.02, \\
            &&&&                                                                     0.1, 0.02, 0.02, 0.02, 0.02,  \\
            &&&&                                                                     0.02, 0.02, 0.02, 0.02, 0.02, \\
            &&&&                                                                     0.02, 0.02, 0.02, 0.02, 0.02, \\
            &&&&                                                                     0.02, 0.02]                   \\ \hline
            par\_load          & Ordinal       & [\textbf{1}, 2, 4]             & 1 & [0.45, 0.1, 0.45]              \\ \hline
            loop\_p            & Ordinal       & [\textbf{1}, 2, ..., 31, 32]   & 2 & [0.1, 0.1, 0.1, 0.1, 0.05, 0.03, \\
            &&&&                                                                     0.02, 0.02, 0.02, 0.02, 0.02, \\
            &&&&                                                                      0.02, 0.02, 0.02, 0.02, 0.02, \\
            &&&&                                                                     0.02, 0.02, 0.02, 0.02, 0.02,  \\ 
            &&&&                                                                     0.02, 0.02, 0.02, 0.02, 0.02, \\ 
            &&&&                                                                     0.02, 0.02, 0.02, 0.02, 0.02,  
                                                                                 0.02] \\ \hline
            loop\_grid0\_x     & Ordinal       & [\textbf{1}, 2, ..., 15, 16]   & 1 & [0.2, 0.1, 0.05, 0.05, 0.05, \\
            &&&&                                                                     0.05, 0.05, 0.05, 0.05, 0.05, \\
            &&&&                                                                     0.05, 0.05, 0.05, 0.05, 0.05, 
                                                                                 0.05] \\ \hline
            loop\_grid1\_z     & Ordinal       & [\textbf{1}, 2, ..., 15, 16]   & 1 & [0.2, 0.2, 0.1, 0.1, 0.07, \\ 
            &&&&                                                                           0.03, 0.03, 0.03, 0.03, 0.03,  \\
            &&&&                                                                           0.03, 0.03, 0.03, 0.03, 0.03, 
                                                                                       0.03]  \\ \hline
            loop\_grid0\_y     & Ordinal       & [\textbf{1}, 2, ..., 15, 16]   & 1 & [0.2, 0.1, 0.05, 0.05, 0.05, \\
            &&&&                                                                     0.05, 0.05, 0.05, 0.05, 0.05, \\
            &&&&                                                                     0.05, 0.05, 0.05, 0.05, 0.05,
                                                                                 0.05] \\ \hline
            ATOM1LOOP          & Categorical   & [false, \textbf{true}]         & true & [0.1, 0.9]     \\ \hline
            ATOM2LOOP          & Categorical   & [false, \textbf{true}]         & true & [0.1, 0.9]     \\ \hline
            PLOOP              & Categorical   & [false, \textbf{true}]         & true & [0.1, 0.9]     \\ \hline
        \end{tabular}}%
\end{table}

In our experiments, we introduce for the first time the automatic optimization of three \spatial real-world applications, namely, 7D shallow and deep CNNs, and a 10D molecular dynamics grid application. Previous work by Nardi \etal\cite{nardi18hypermapper} had applied automatic optimization of \spatial parameters on a set of benchmarks but in our work we focus on real-world applications raising the bar of state-of-the-art automated hardware design optimization. \name is used to optimize the parameters to find a design that leads to the fastest runtime. The search space for these three applications is based on ordinal and categorical parameters; to handle these discrete parameters in the best way we implement and use a Random Forests surrogate instead of a Gaussian Process one as explained in Appendix~\ref{sec:experimental_setup.appendix}. 
These parameters are application specific and control how much of the FPGAs' resources we want to use to parallelize each step of the application's computation. The goal here is to find which steps are more important to parallelize in the final design, in order to achieve the fastest runtime. Some parameters also control whether we want to enable pipeline scheduling or not, which consumes resources but accelerates runtime, and others focus on  memory   management. We refer to Koeplinger \etal\cite{koeplinger2018} and Nardi \etal\cite{nardi18hypermapper} for more details on \spatial's parameters.

The three \spatial benchmarks also have feasibility constraints in the search space, meaning that some parameter configurations are infeasible. A configuration is considered infeasible if the final design requires more logic resources than what the FPGA provides, i.e., it is not possible to perform FPGA synthesis because the design does not fit in the FPGA. To handle these constraints, we use our cBO implementation (Appendix~\ref{sec:experimental_setup.appendix}). Our goal is thus to find the design with the fastest runtime under the constraint that the design fits the FPGA resource budget.

The \priorsstr for these \spatial applications take the form of a list of probabilities, containing the probability of each ordinal or categorical value being good. Each benchmark also has a default configuration, which ensures all methods start with at least one feasible configuration. The \priorsstr and the default configuration for these benchmarks were provided once by an unbiased \spatial developer, who is not an author of this paper, and kept unchanged during the entire project. The search space, \priorsstr, and the expert configuration used in our experiments for each application are presented in Tables~\ref{tab:spatial.shallow}, \ref{tab:spatial.deep}, and \ref{tab:spatial.mdgrid}.

\section{Multivariate \Priorstr Comparison} \label{sec:multivariate.appendix}

\begin{figure*}[tb]        
    \begin{center}
        \subfigure{\includegraphics[scale=0.2]{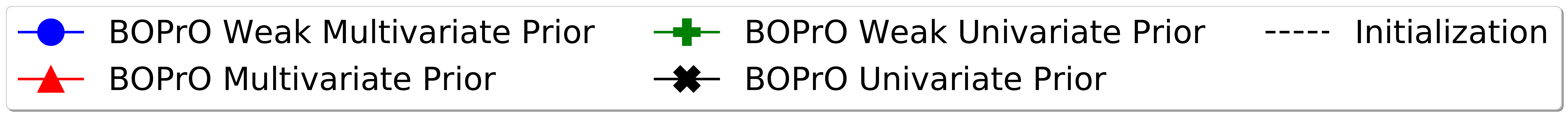}}\\
        \subfigure{\includegraphics[width=0.42\linewidth]{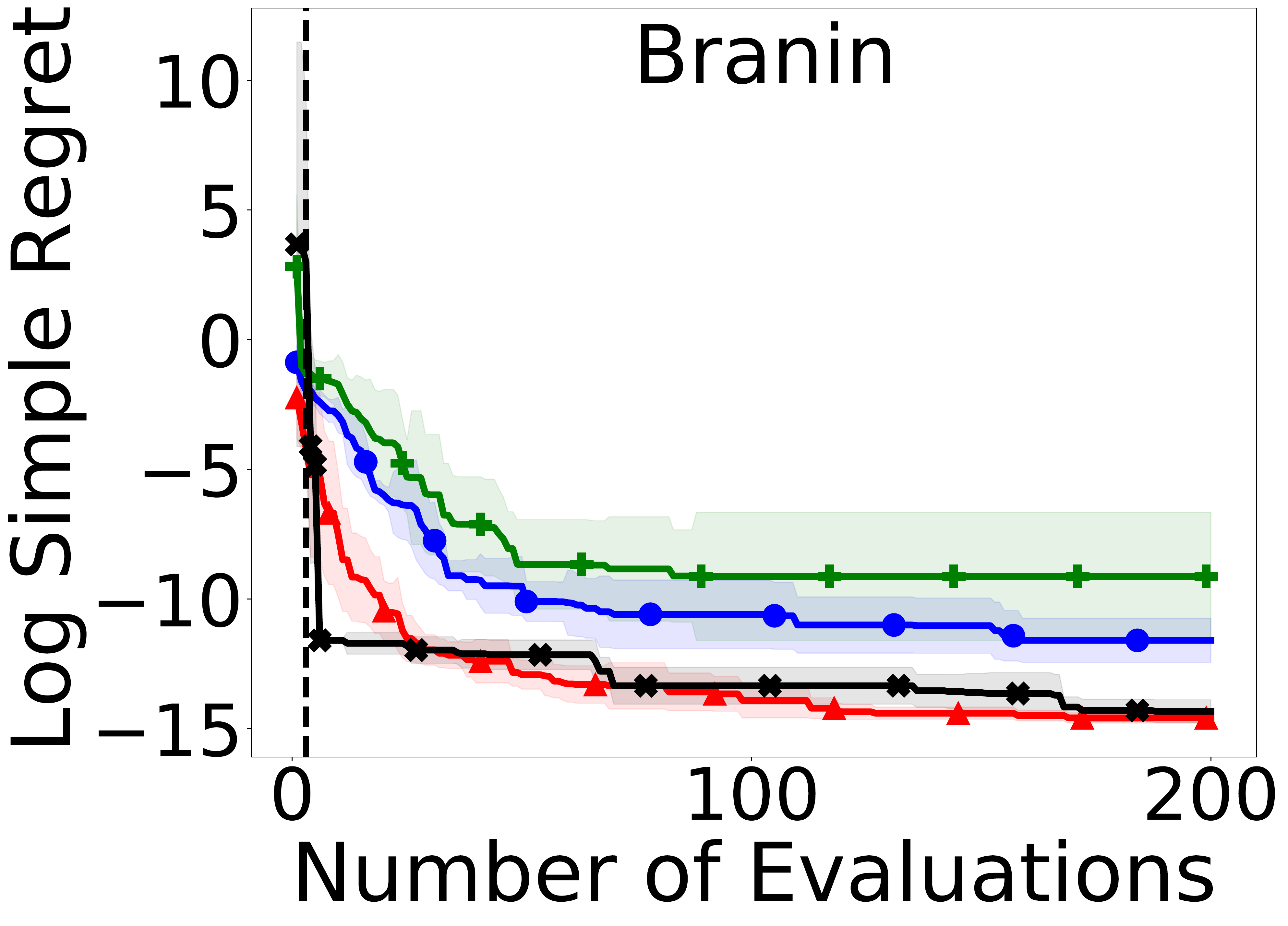}}
        \subfigure{\includegraphics[width=0.42\linewidth]{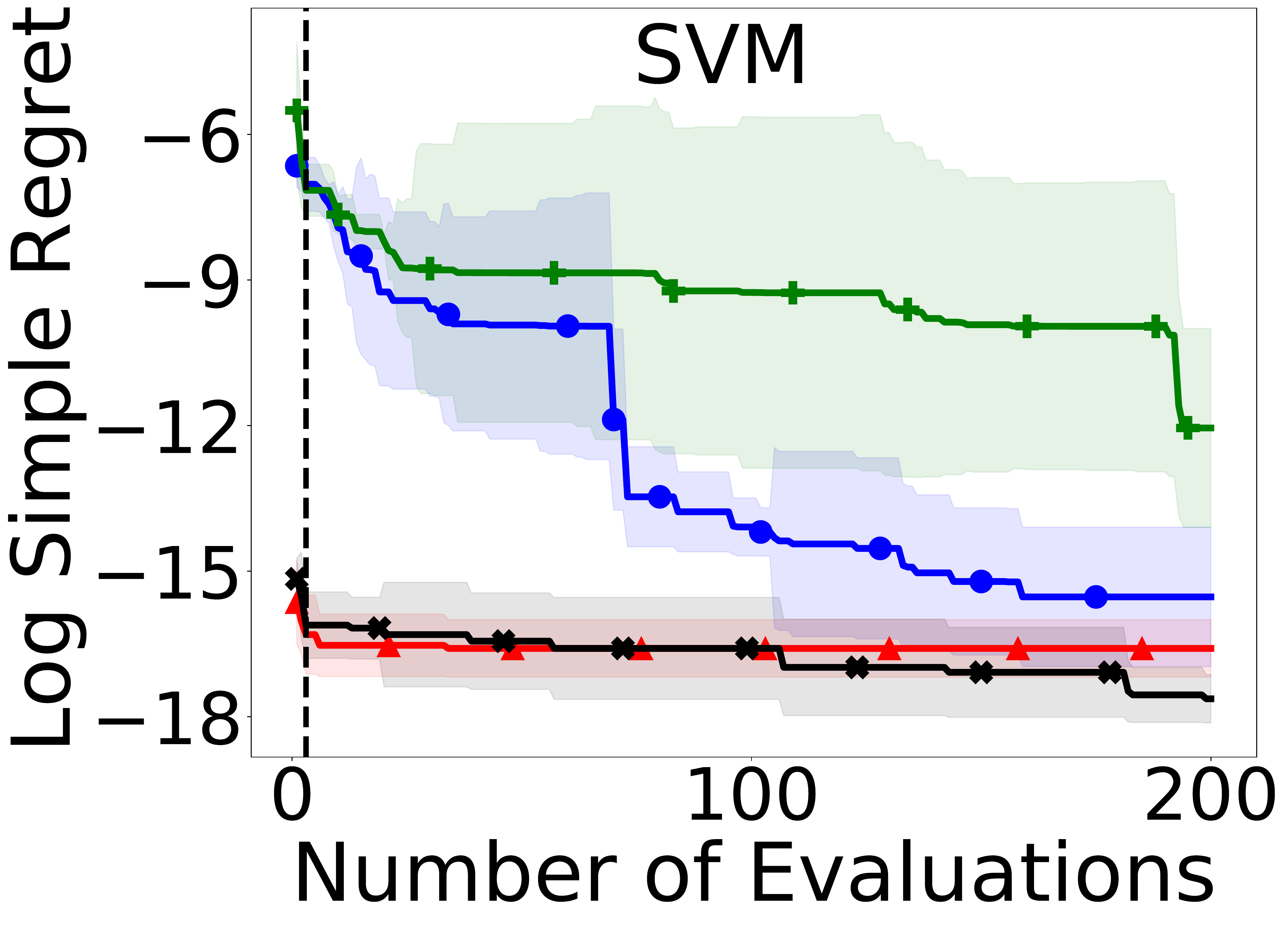}}\\
        \subfigure{\includegraphics[width=0.42\linewidth]{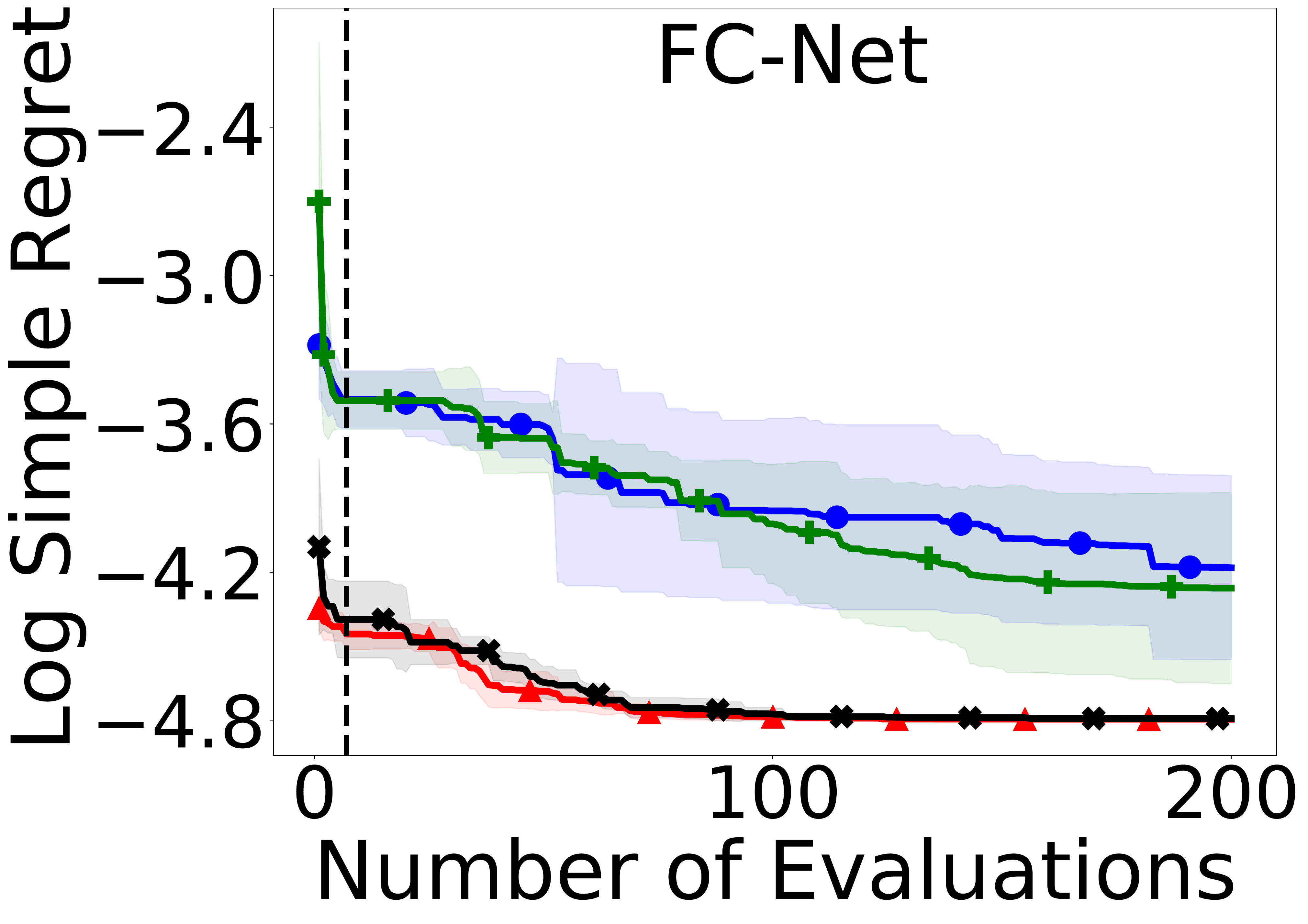}}
        \subfigure{\includegraphics[width=0.42\linewidth]{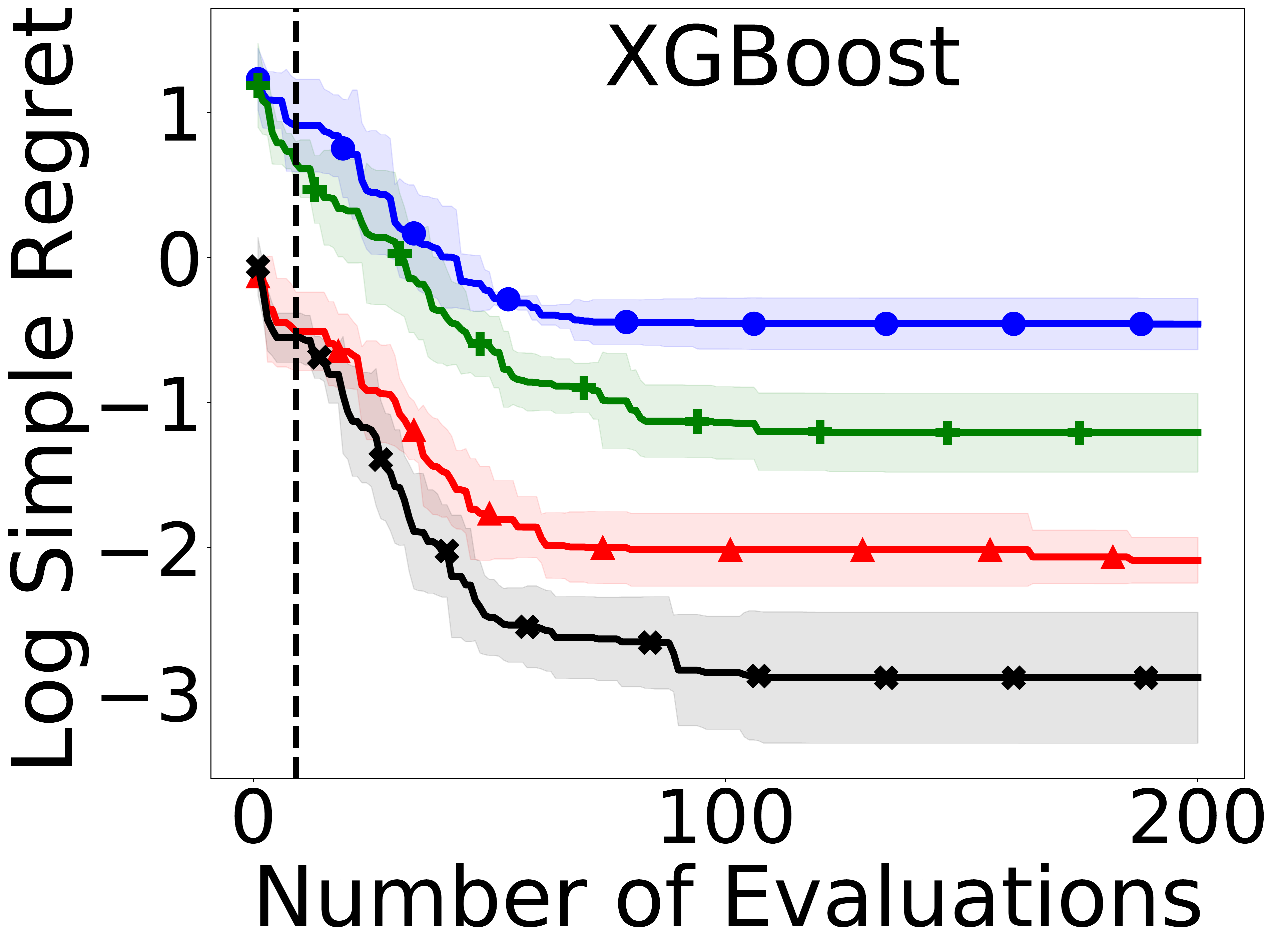}}
    \end{center}
    \caption{Log regret comparison of \name with multivariate and univariate KDE \priorsstr. The line and shaded regions show the mean and standard deviation of the log simple regret after 5 runs. All methods were initialized with $D+1$ random samples, where $D$ is the number of input dimensions, indicated by the vertical dashed line. We run the benchmarks for $200$ iterations.}
    \label{fig:multivariate.regret}
\end{figure*}


In this section we compare the performance of \name with univariate and multivariate priors. For this, we construct synthetic univariate and multivariate priors using Kernel Density Estimation (KDE) with a Gaussian kernel. We build strong and weak versions of the KDE priors. The strong priors are computed using a KDE on the best $10D$ out of $10$,$000$,$000D$ uniformly sampled points, while the weak priors are computed using a KDE on the best $10D$ out of $1$,$000D$ uniformly sampled points. We use the same points for both univariate and multivariate priors. We use scipy's Gaussian KDE implementation, but adapt its Scott's Rule bandwidth to $100n^{-\frac{1}{d}}$, where $d$ is the number of variables in the KDE prior, to make our priors more peaked.

Figure~\ref{fig:multivariate.regret} shows a log regret comparison of \name with univariate and multivariate KDE \priorsstr. We note that in all cases \name achieves similar performance with univariate and multivariate \priorsstr. For the Branin and SVM benchmarks, the weak multivariate \priorstr leads to slightly better results than the weak univariate \priorstr. However, we note that the difference is small, in the order of $10^{-4}$ and $10^{-6}$, respectively. 

Surprisingly, for the XGBoost benchmark, the univariate version for both the weak and strong \priorsstr lead to better results than their respective multivariate counterparts, though, once again, the difference in performance is small, around $0.2$ and $0.03$ for the weak and strong \priorstr, respectively, whereas the XGBoost benchmark can reach values as high as $f(\param) = 700$. Our hypothesis is that this difference comes from the bandwidth estimator ($100n^{-\frac{1}{d}}$), which leads to larger bandwidths, consequently, smoother \priorsstr, when a multivariate prior is constructed.

\begin{figure*}[tb]        
    \begin{center}
        \subfigure{\includegraphics[scale=0.22]{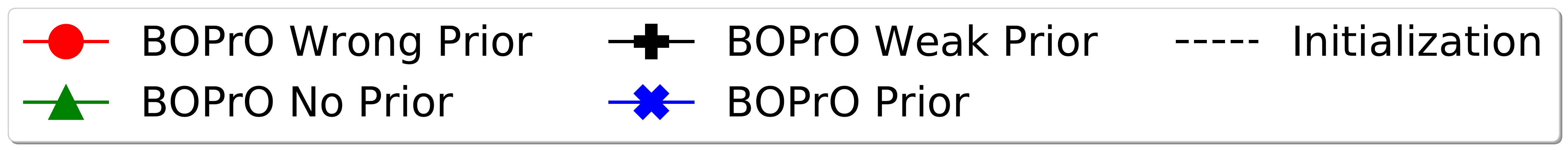}}\\
        \subfigure{\includegraphics[width=0.4\linewidth]{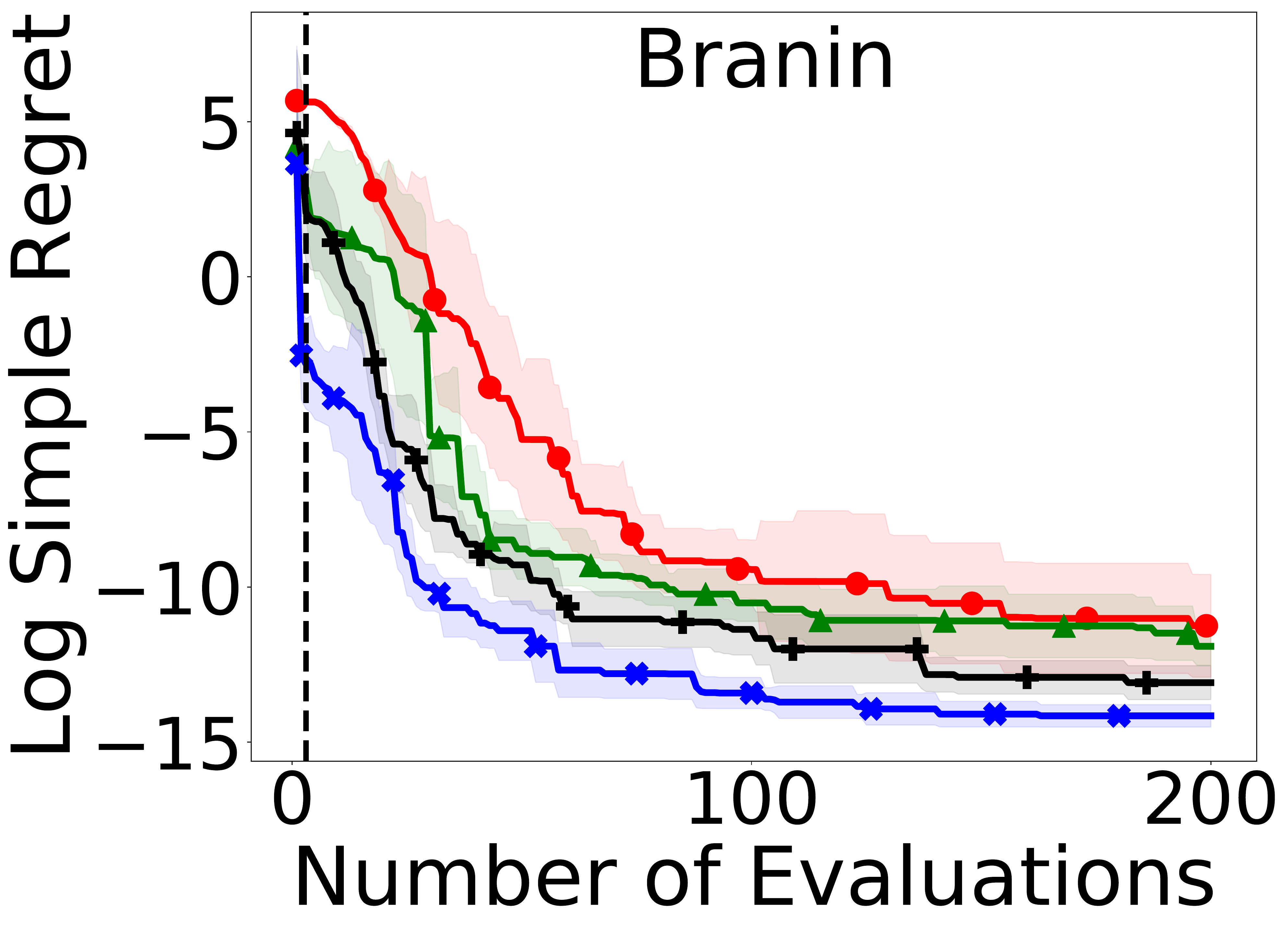}}
        \subfigure{\includegraphics[width=0.4\linewidth]{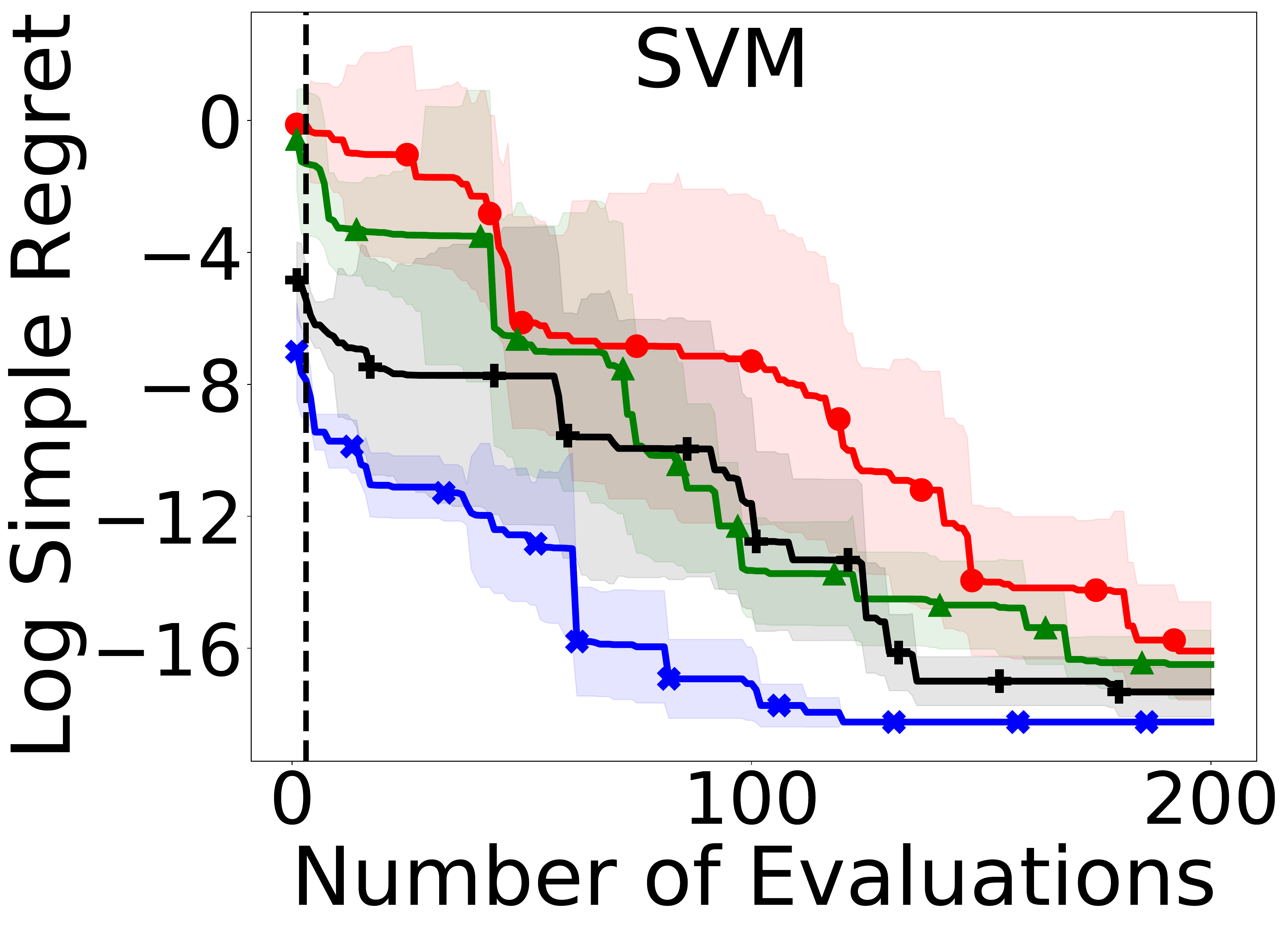}}\\
        \subfigure{\includegraphics[width=0.4\linewidth]{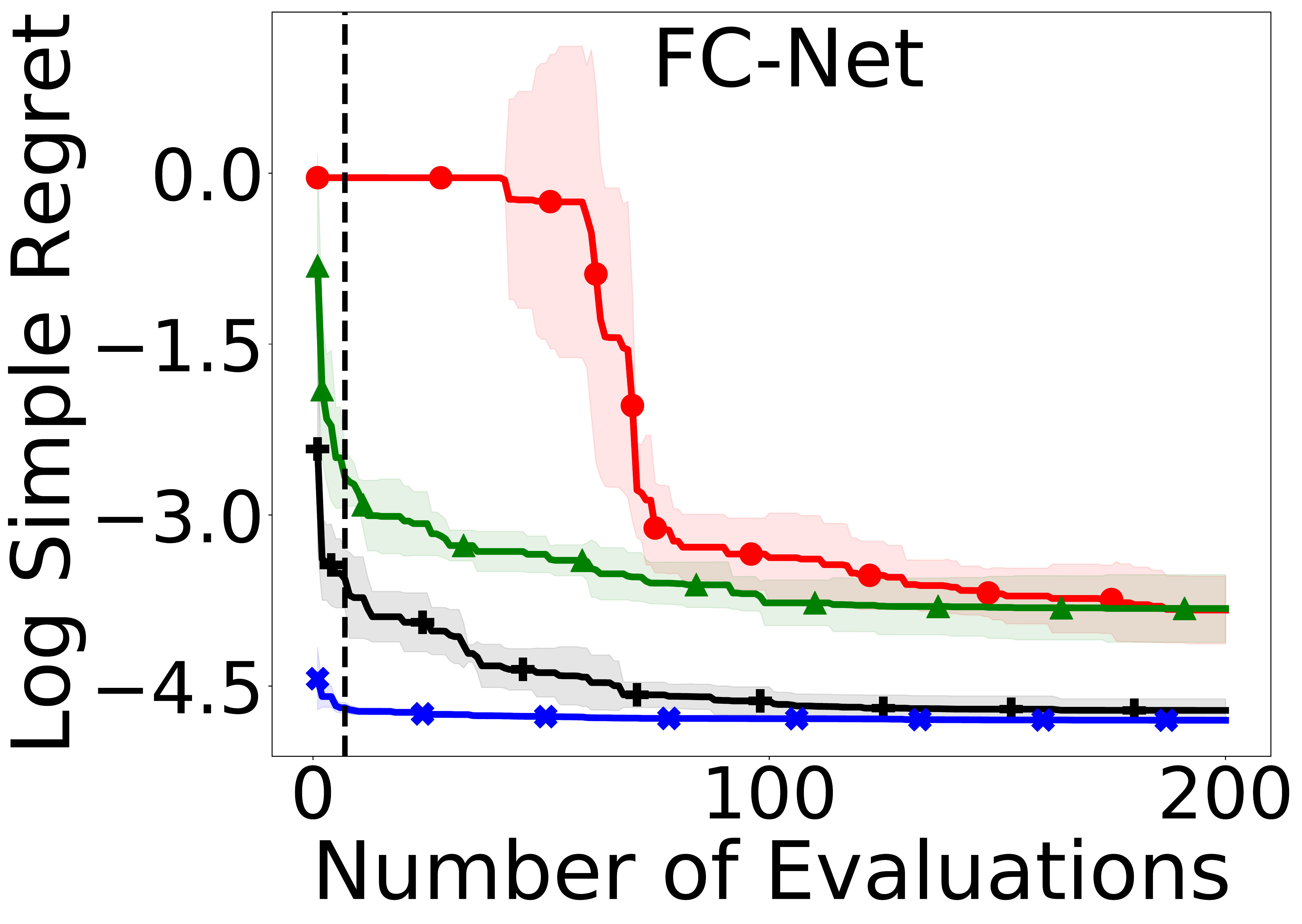}}
        \subfigure{\includegraphics[width=0.4\linewidth]{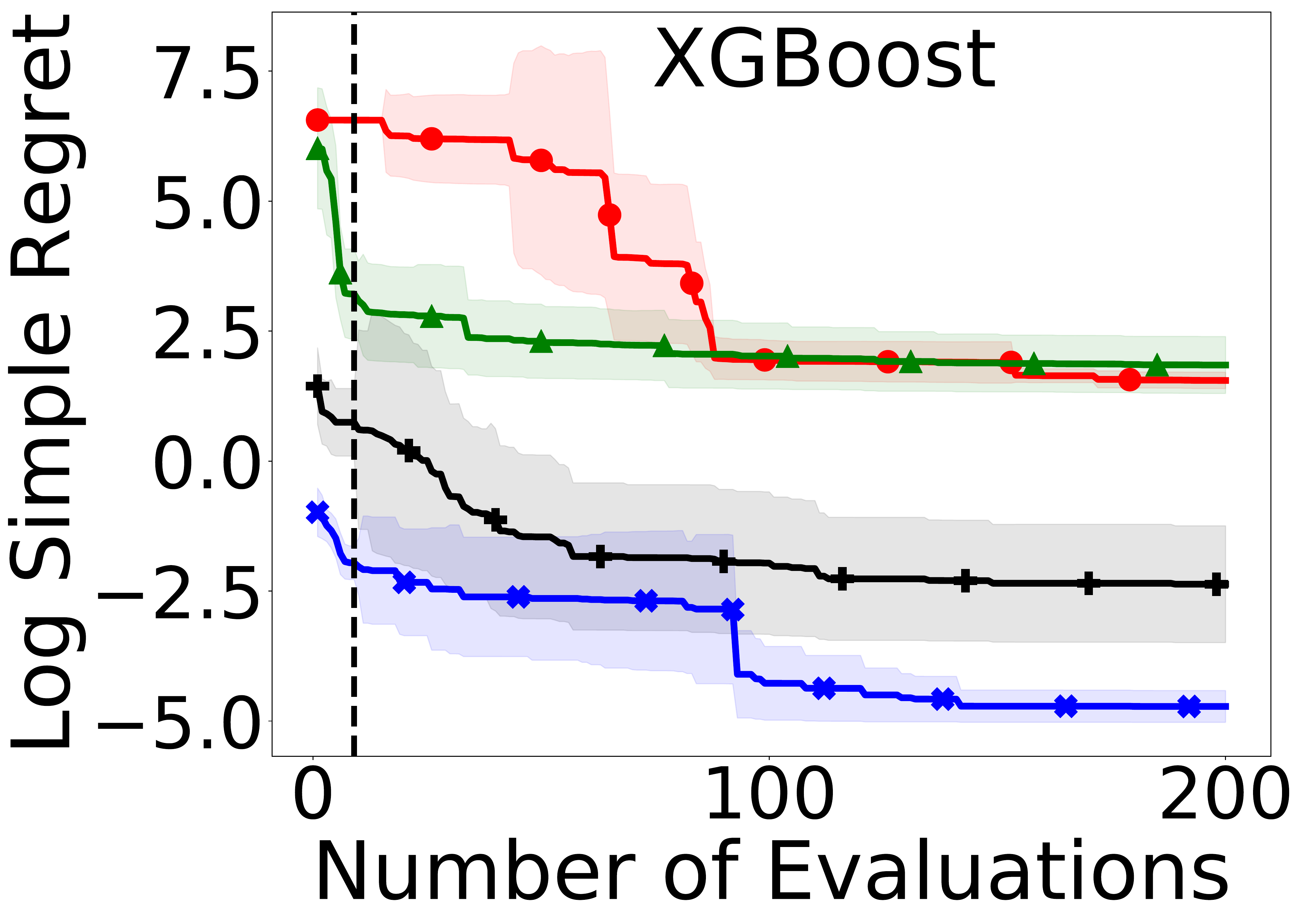}}
    \end{center}
    \caption{Log regret comparison of \name with varying \priorstr quality. The line and shaded regions show the mean and standard deviation of the log simple regret after 5 runs. All methods were initialized with $D+1$ random samples, where $D$ is the number of input dimensions, indicated by the vertical dashed line.  We run the benchmarks for $200$ iterations.}
    \label{fig:misleading.regret}
\end{figure*}

\section{Misleading \Priorstr Comparison} \label{sec:misleading}

Figure~\ref{fig:misleading.regret} shows the effect of injecting a misleading \priorstr in \name. We compare \name with a misleading \priorstr, no \priorstr, a weak \priorstr, and a strong \priorstr. For our misleading prior, we use a Gaussian centered at the worst point out of $10$,$000$,$000D$ uniform random samples. Namely, for each parameter, we inject a prior of the form $\mathcal{N}(x_{w}, \sigma_w^2)$, where $x_{w}$ is the value of the parameter at the point with highest function value out of $10$,$000$,$000D$ uniform random samples and $\sigma_w = 0.01$. For all benchmarks, we note that the misleading \priorstr slows down convergence, as expected, since it pushes the optimization away from the optima in the initial phase. However, \name is still able to forget the misleading \priorstr and achieve similar regret to \name without a \priorstr.

\section{Comparison to Other Baselines} \label{sec:tpe.appendix}


\begin{figure*}[tb]        
    \begin{center}
        \subfigure{\includegraphics[scale=0.23]{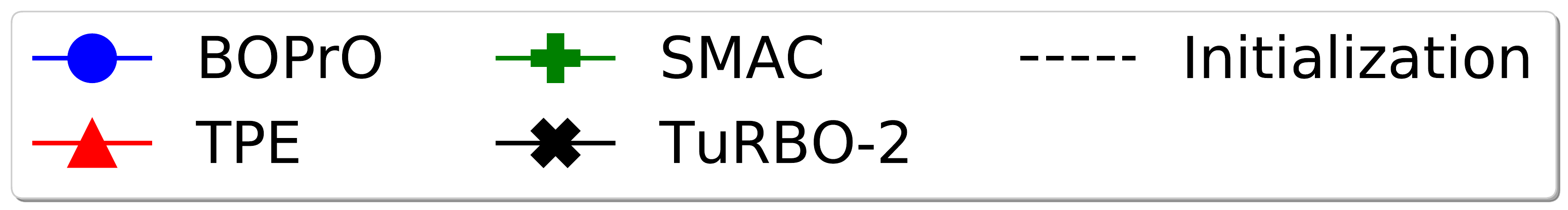}}\\
        \subfigure{\includegraphics[width=0.42\linewidth]{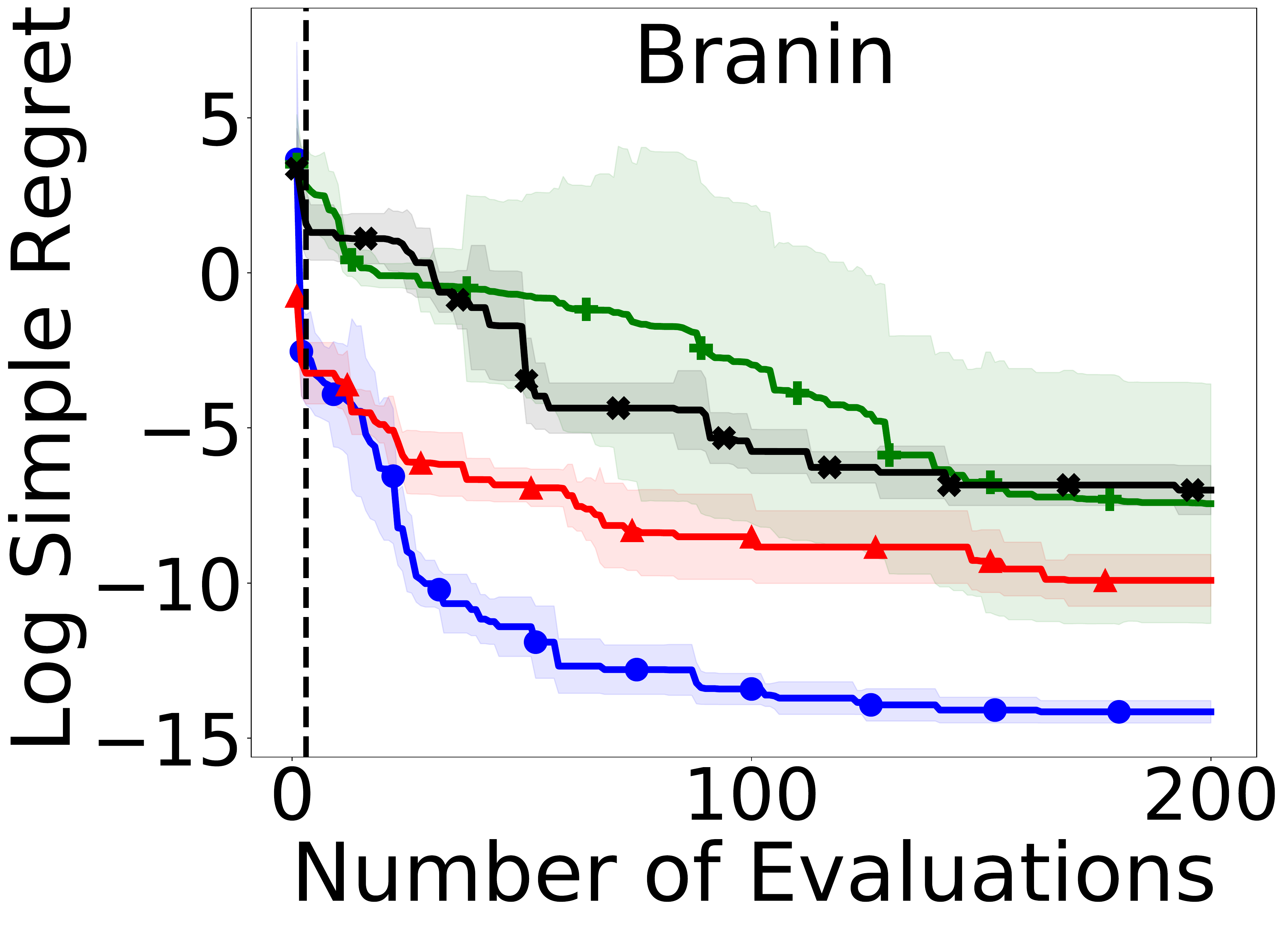}}
        \subfigure{\includegraphics[width=0.42\linewidth]{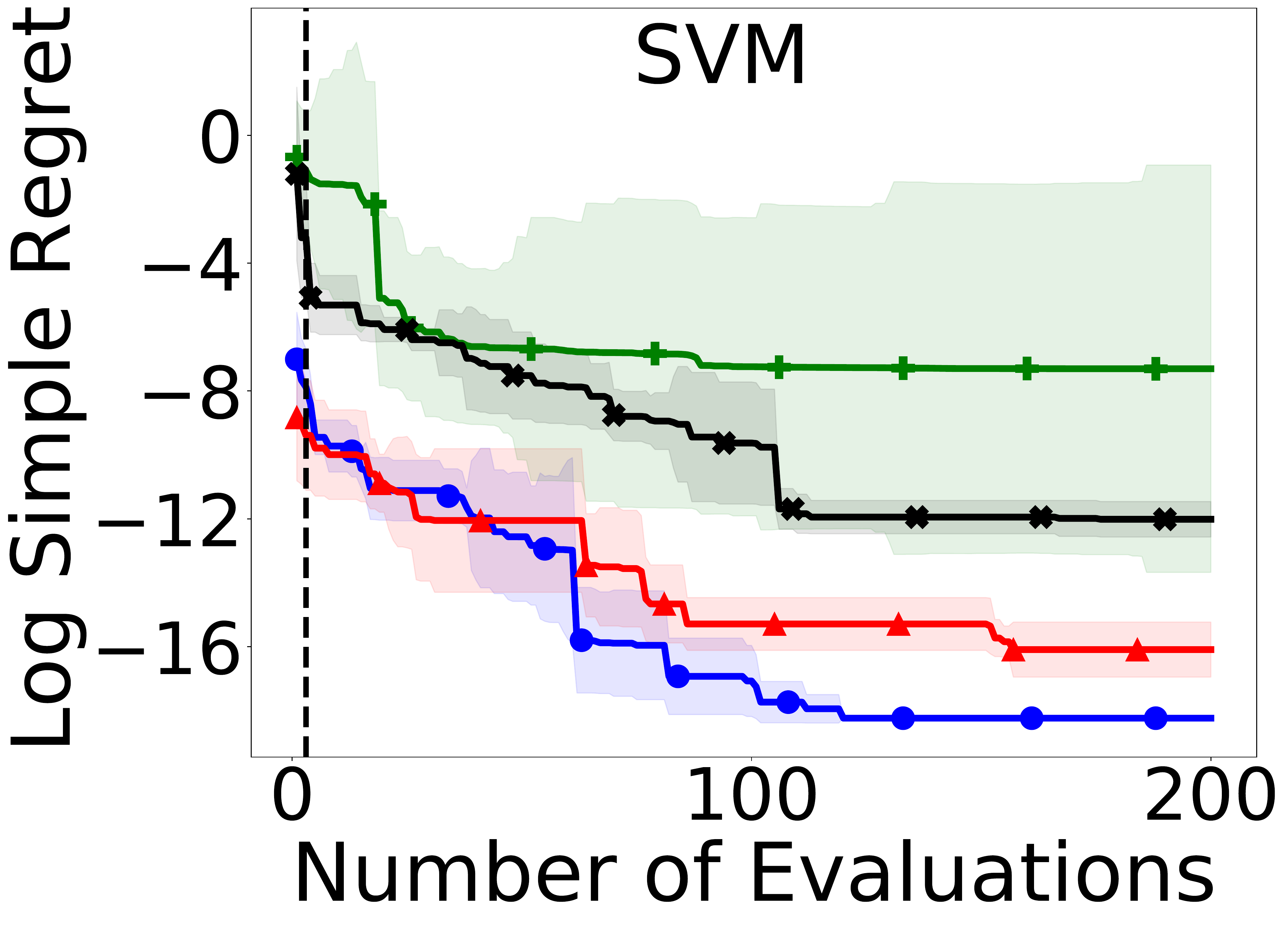}} \\
        \subfigure{\includegraphics[width=0.41\linewidth]{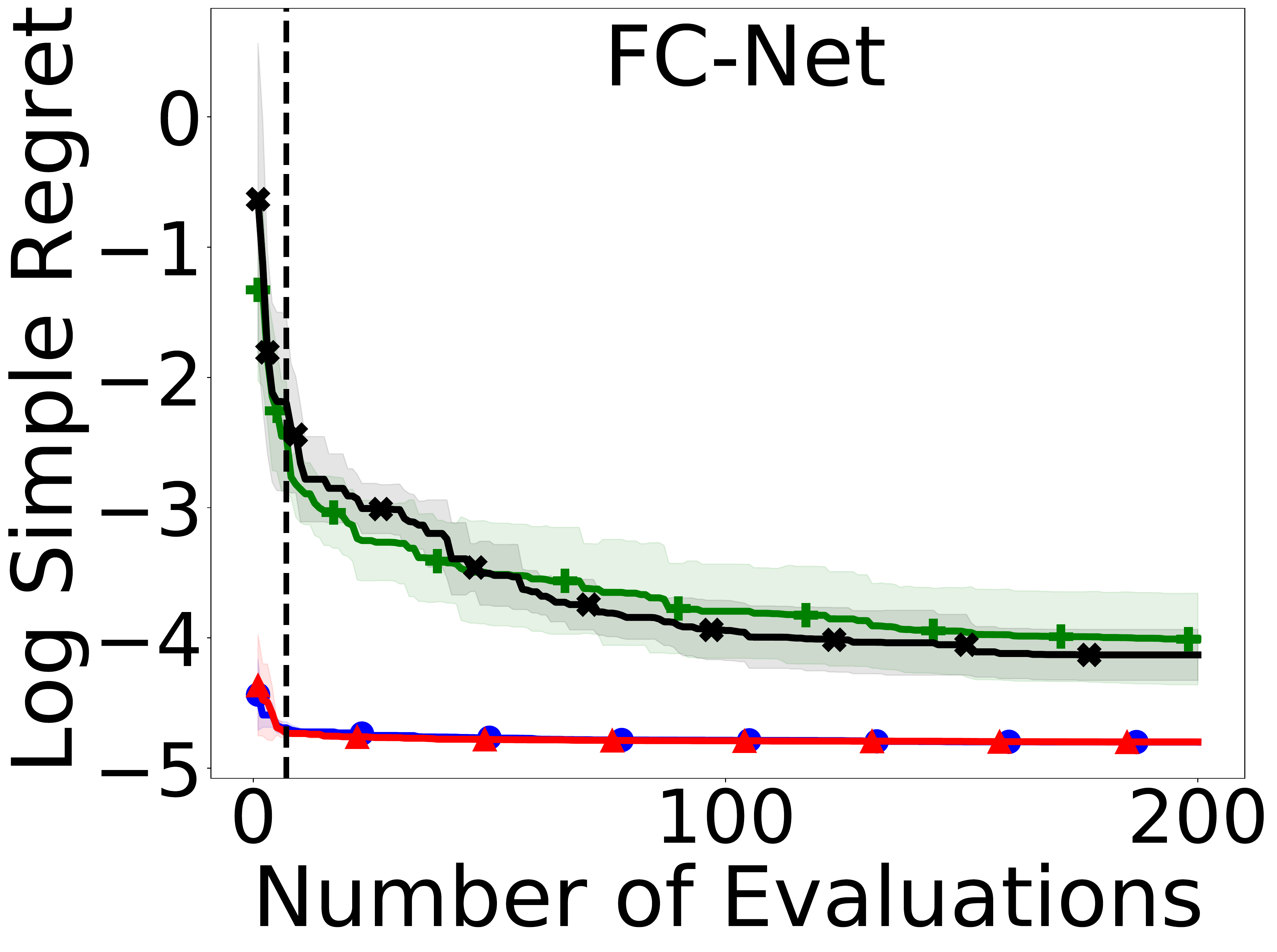}}
        \subfigure{\includegraphics[width=0.43\linewidth]{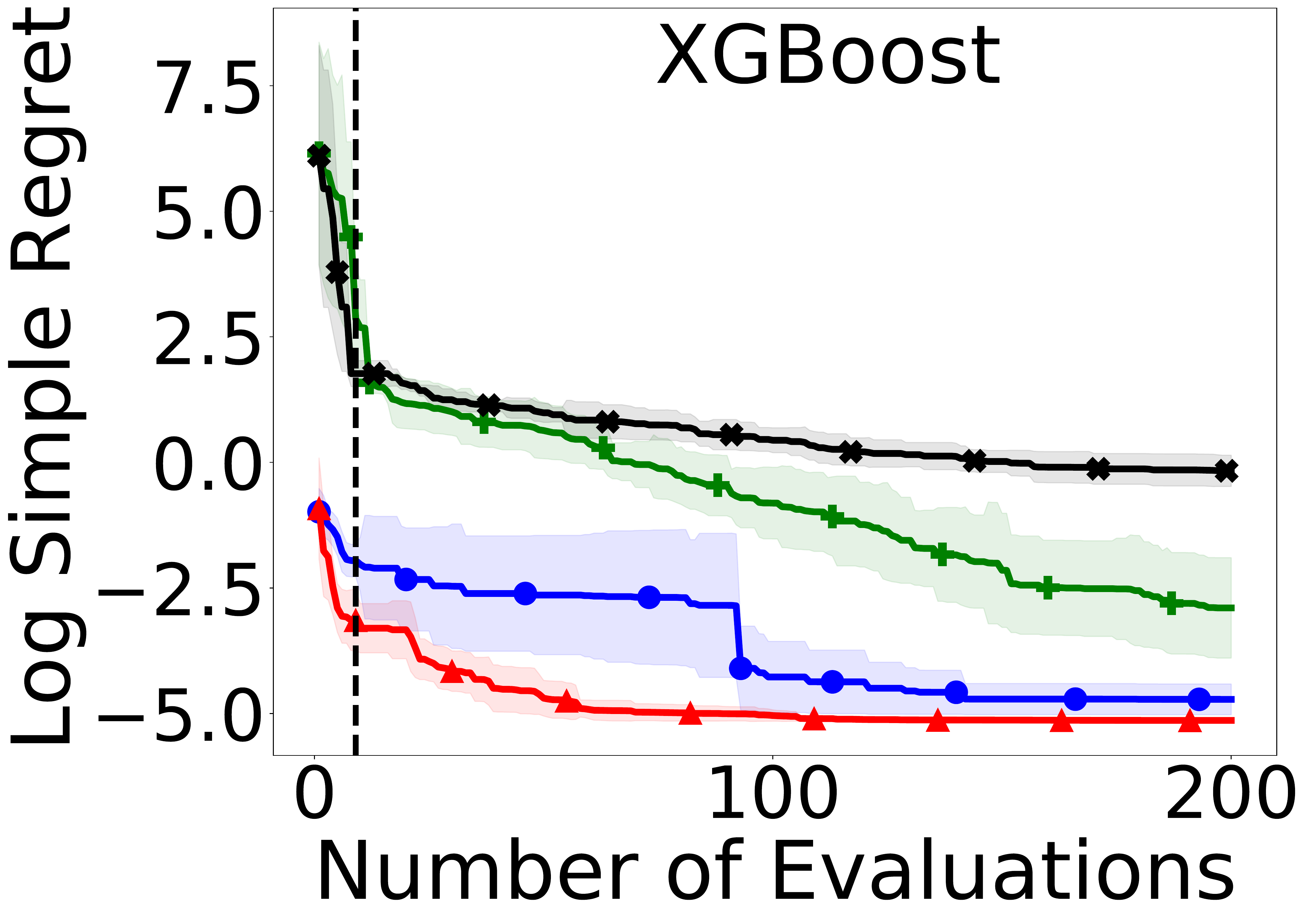}}
    \end{center}
    \caption{Log regret comparison of \name, SMAC, and TPE. The line and shaded regions show the mean and standard deviation of the log simple regret after 5 runs. \name was initialized with $D+1$ random samples, where $D$ is the number of input dimensions, indicated by the vertical dashed line.  We run the benchmarks for $200$ iterations.}
    \label{fig:tpe.regret}
\end{figure*}

\begin{figure*}[tb]        
    \begin{center}
        \subfigure{\includegraphics[scale=0.23]{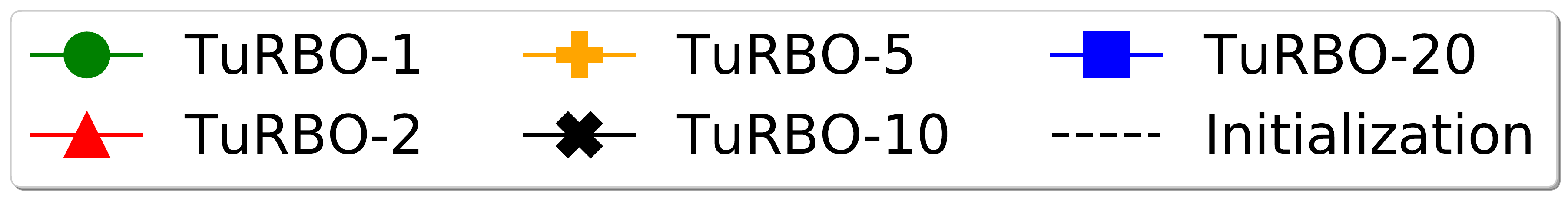}}\\
        \subfigure{\includegraphics[width=0.4\linewidth]{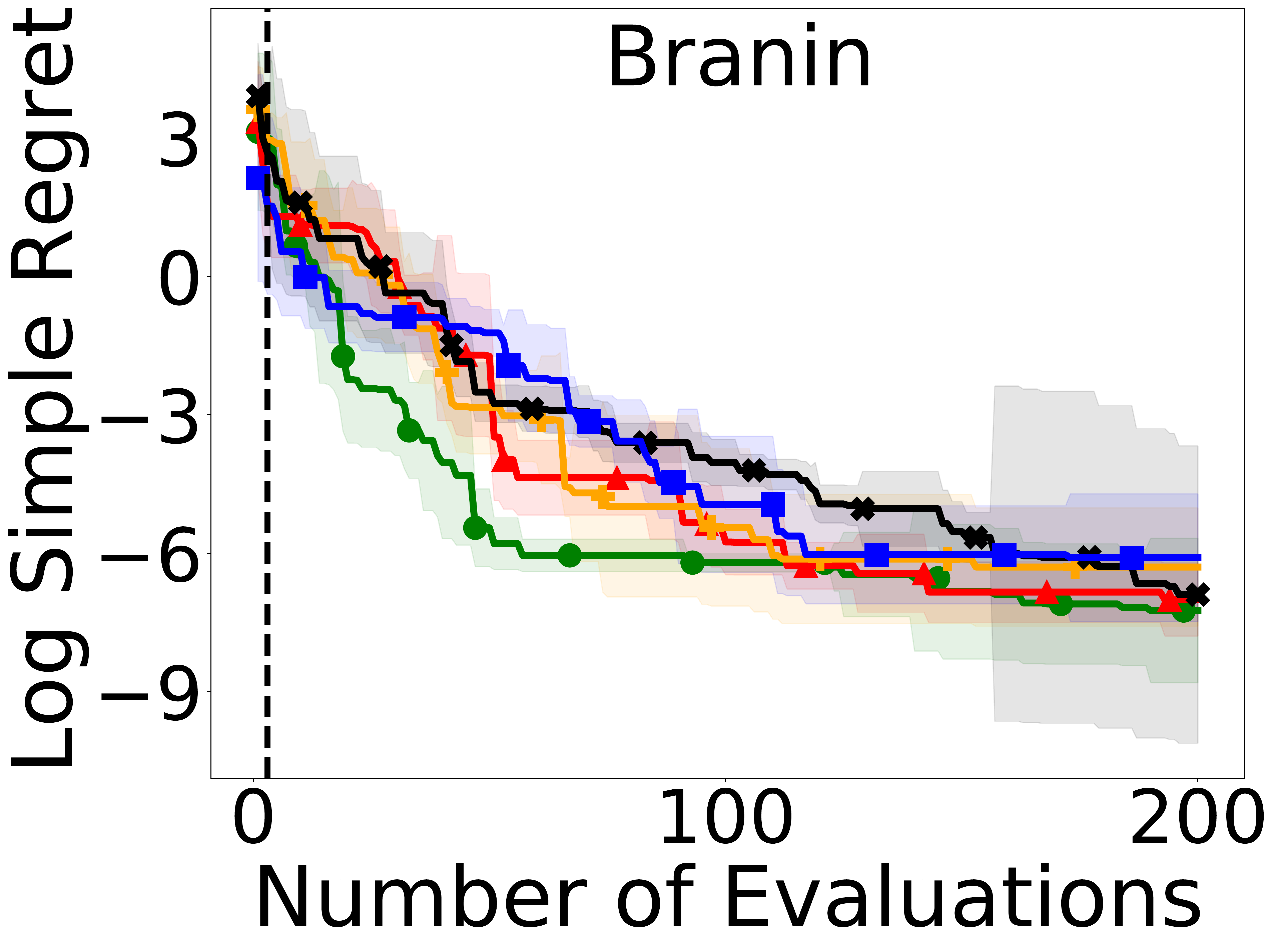}}
        \subfigure{\includegraphics[width=0.415\linewidth]{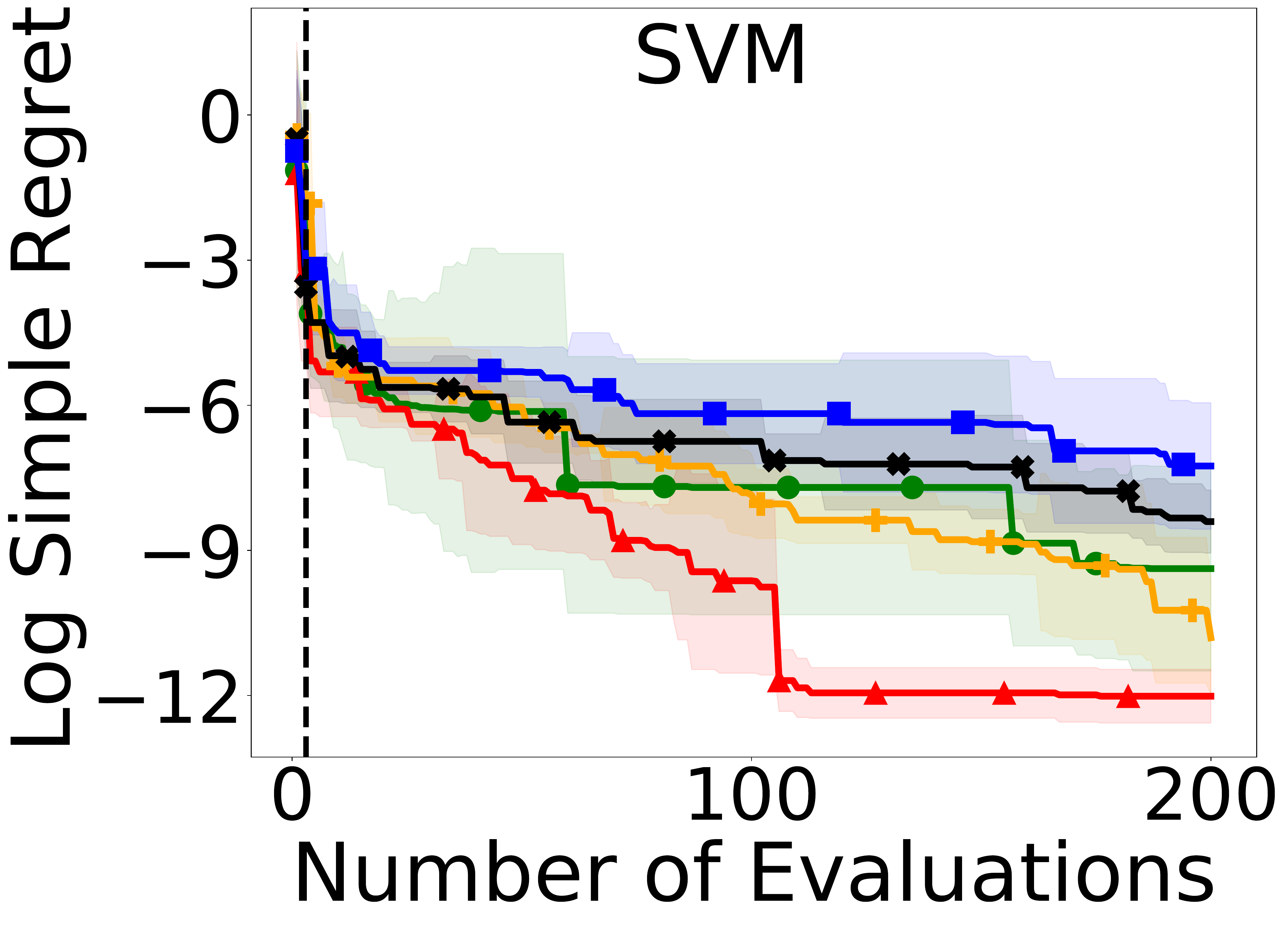}} \\
        \subfigure{\includegraphics[width=0.4\linewidth]{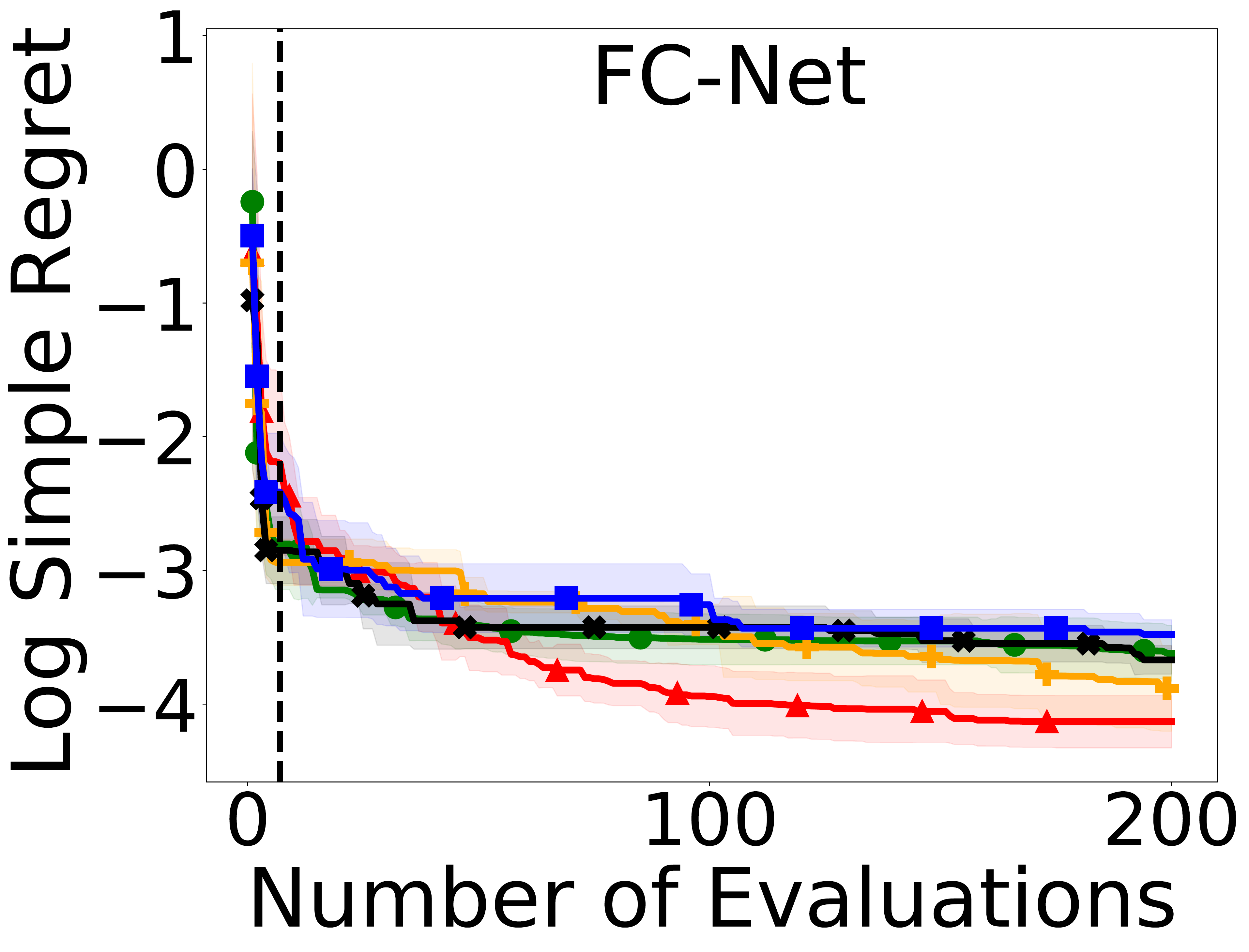}}
        \subfigure{\includegraphics[width=0.4\linewidth]{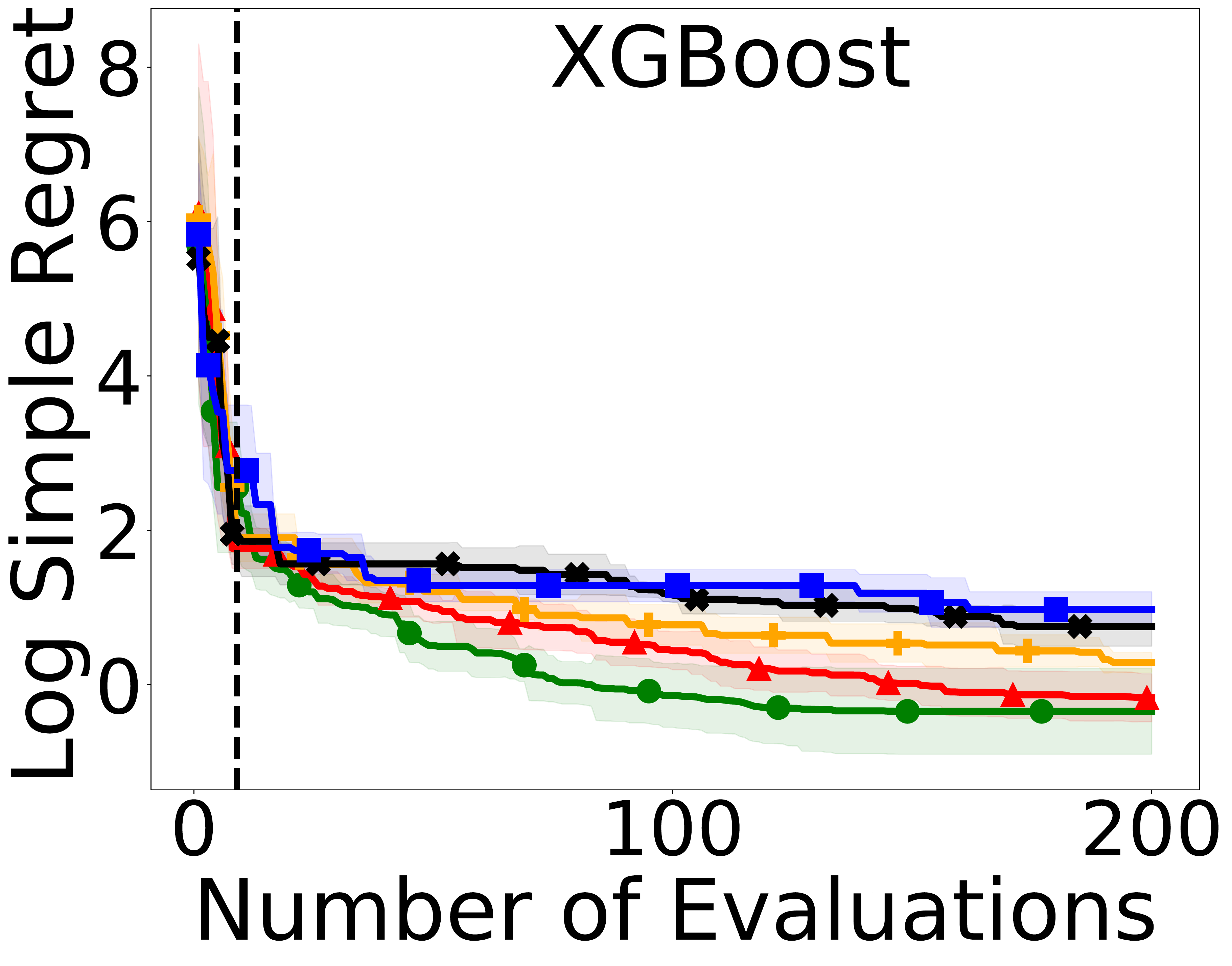}}
    \end{center}
    \caption{Log regret comparison of TuRBO with different number of trust regions. TuRBO-$M$ denotes TuRBO with $M$ trust regions. The line and shaded regions show the mean and standard deviation of the log simple regret after 5 runs. TuRBO was initialized with $D+1$ uniform random samples, where $D$ is the number of input dimensions, indicated by the vertical dashed line. We run the benchmarks for $200$ iterations.}
    \label{fig:turbo.ms}
\end{figure*}

We compare \name to SMAC~\cite{hutter2011sequential}\edit{, TuRBO~\cite{eriksson2019scalable},}{} and TPE~\cite{bergstra2011algorithms} on our four synthetic benchmarks. We use Hyperopt's implementation\footnote{https://github.com/hyperopt/hyperopt} of TPE\edit{, the public implementation of TuRBO\footnote{https://github.com/uber-research/TuRBO},}{} and the SMAC3 Python implementation of SMAC\footnote{https://github.com/automl/SMAC3}. Hyperopt defines \priorsstr as one of a list of supported distributions, including Uniform, Normal, and Lognormal distributions, while SMAC \edit{and TuRBO}{} do not support priors on the locality of an optimum under the form of probability distributions.

For the three Profet benchmarks (SVM, FCNet, and XGBoost), we inject the strong priors defined in Section~\ref{sec:experiments.regret} into both Hyperopt and \name. For Branin, we also inject the strong prior defined in Section~\ref{sec:experiments.regret} into \name, however, we cannot inject this prior into Hyperopt. Our strong prior for Branin takes the form of a Gaussian mixture prior peaked at all three optima and Hyperopt does not support Gaussian mixture priors. Instead, for Hyperopt, we arbitrarily choose one of the optima ($(\pi, 2.275)$) and use a Gaussian prior centered near that optimum. We note that since we compare all approaches based on the log simple regret, both priors are comparable in terms of prior strength, since finding one optimum or all three would lead to the same log regret. Also, we note that using Hyperopt's Gaussian \priorsstr leads to an unbounded search space, which sometimes leads TPE to suggest parameter configurations outside the allowed parameter range. To prevent these values from being evaluated, we convert values outside the parameter range to be equal to the upper or lower range limit, depending on which limit was exceeded. \edit{We do not inject any priors into SMAC and TuRBO, since these methods do not support priors about the locality of an optimum}{}.

Figure~\ref{fig:tpe.regret} shows a log regret comparison between \name, SMAC, \edit{TuRBO-2}{} and TPE on our four synthetic benchmarks. \edit{We use TuRBO-2 since it led to better performance overall compared to other variants, see Fig.~\ref{fig:turbo.ms}.}{}
\name achieves better performance than SMAC \edit{and TuRBO}{} on all four benchmarks. Compared to TPE, \name achieves similar or better performance on three of the four synthetic benchmarks, namely Branin, SVM, and FCNet, and slightly worse performance on XGBoost.
We note, however, that the good performance of TPE on XGBoost may be an artifact of the approach of clipping values to its maximal or minimal values as mentioned above.
In fact, the clipping nudges TPE towards promising configurations in this case, since XGBoost has low function values near the edges of the search space. Overall, the better performance of \name is expected, since \name is able to combine prior knowledge with more sample-efficient surrogates, which leads to better performance.

\section{\Priorstr Baselines Comparison} \label{sec:prior_init.appendix}

\begin{figure*}[tb]        
    \begin{center}
        \subfigure{\includegraphics[scale=0.22]{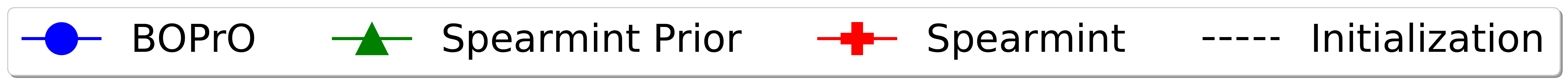}}\\
        \subfigure{\includegraphics[width=0.4\linewidth]{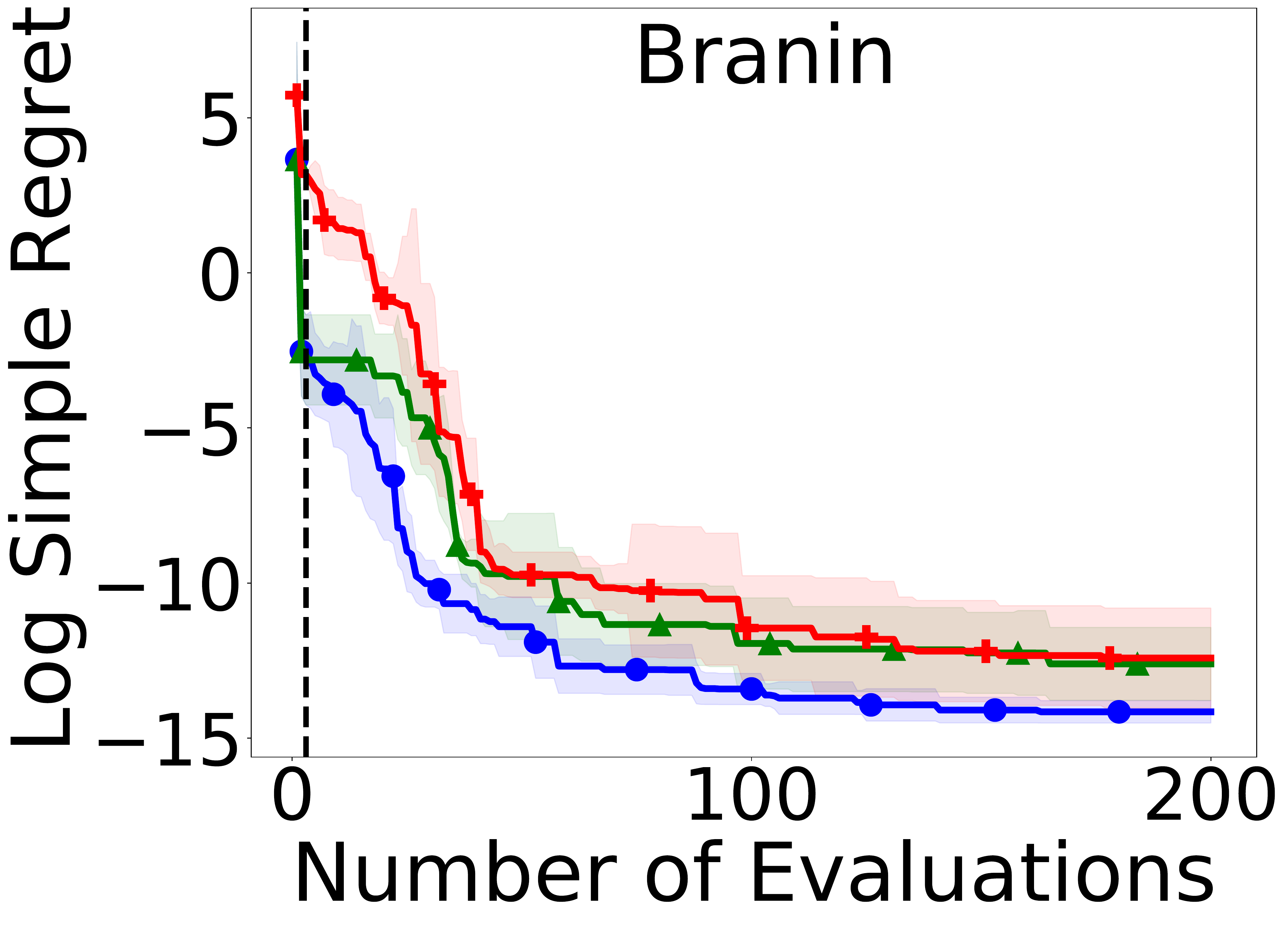}}
        \subfigure{\includegraphics[width=0.4\linewidth]{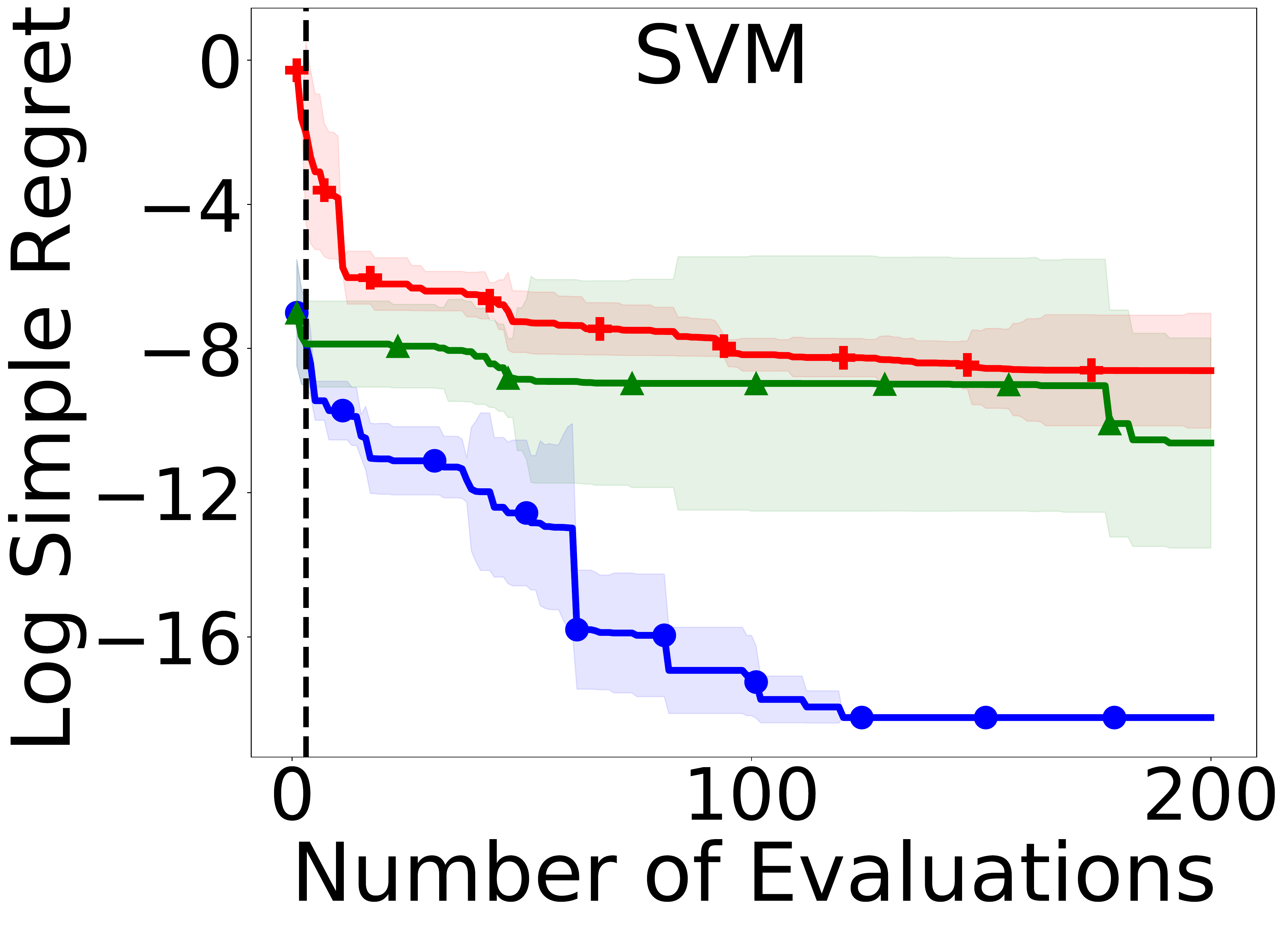}}\\
        \subfigure{\includegraphics[width=0.4\linewidth]{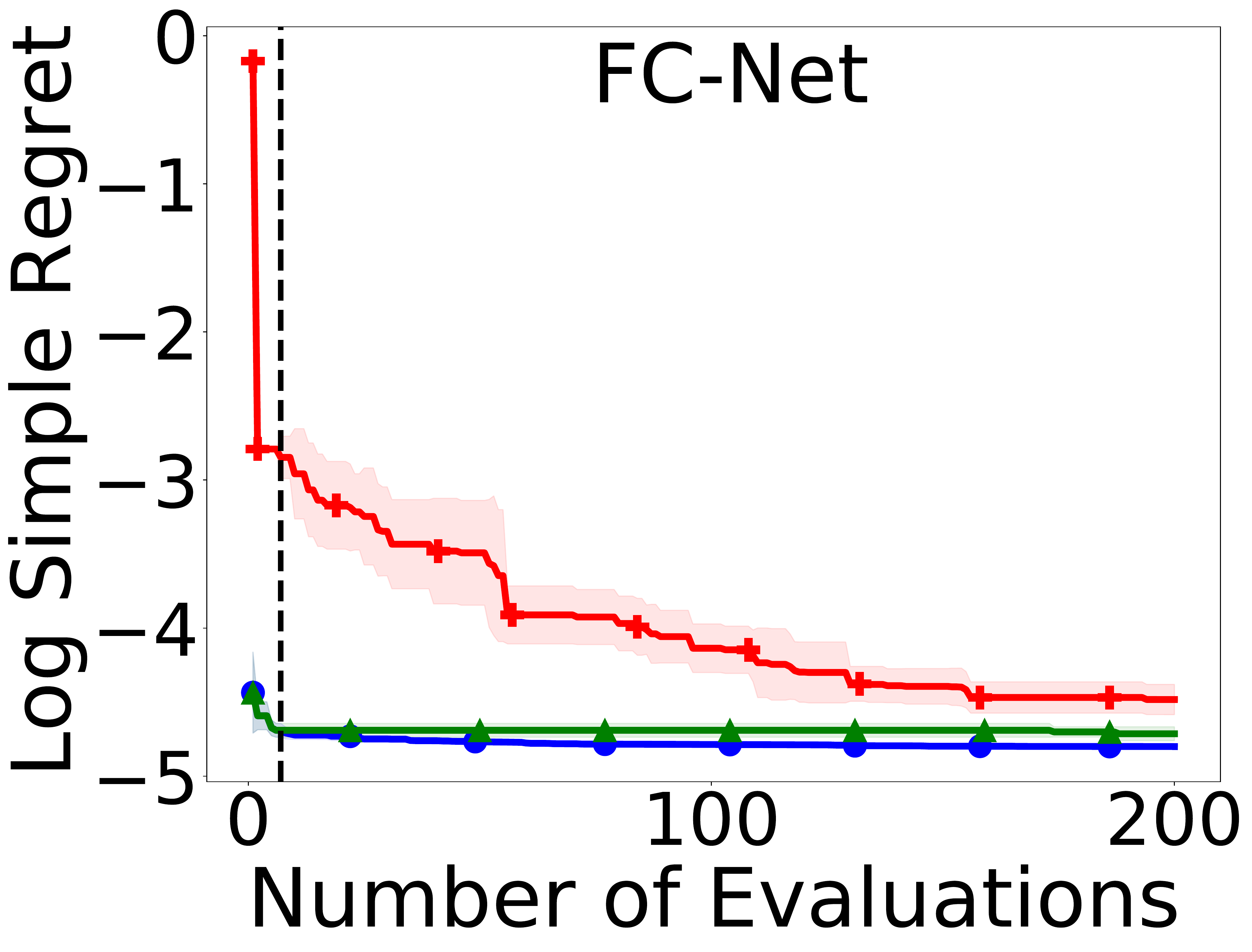}}
        \subfigure{\includegraphics[width=0.4\linewidth]{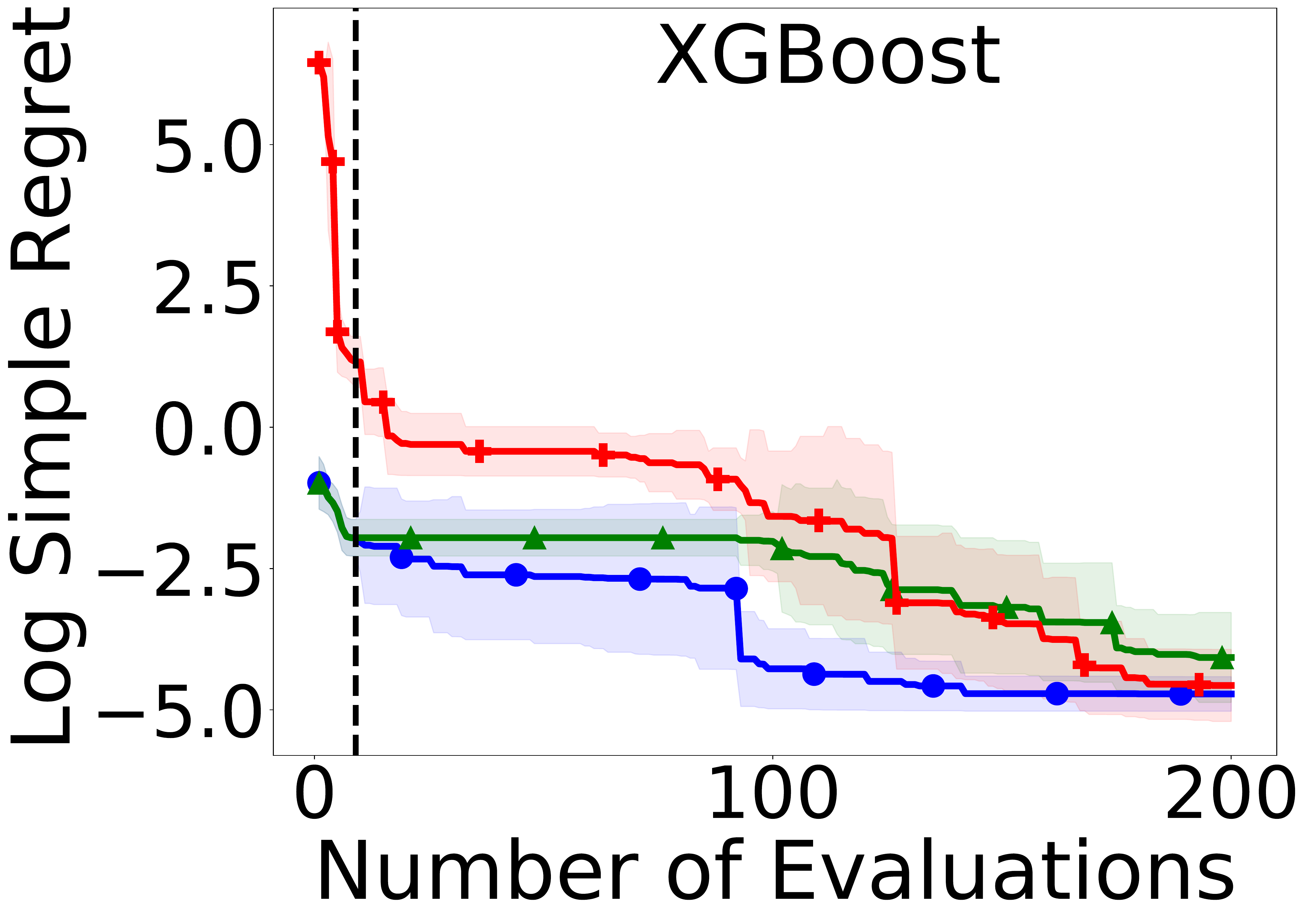}}
    \end{center}
    \caption{Log regret comparison of \name, Spearmint with \priorstr initialization, and Spearmint with default initialization. The line and shaded regions show the mean and standard deviation of the log simple regret after 5 runs. \name and Spearmint \Priorstr were initialized with $D+1$ random samples from the \priorstr, where $D$ is the number of input dimensions, indicated by the vertical dashed line.  We run the benchmarks for $200$ iterations.}
    \label{fig:spearmint.prior}
\end{figure*}

\begin{figure*}[tb]        
    \begin{center}
        \subfigure{\includegraphics[scale=0.13]{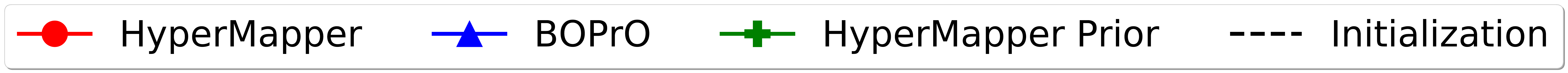}}\\
        \subfigure{\includegraphics[width=0.3\linewidth]{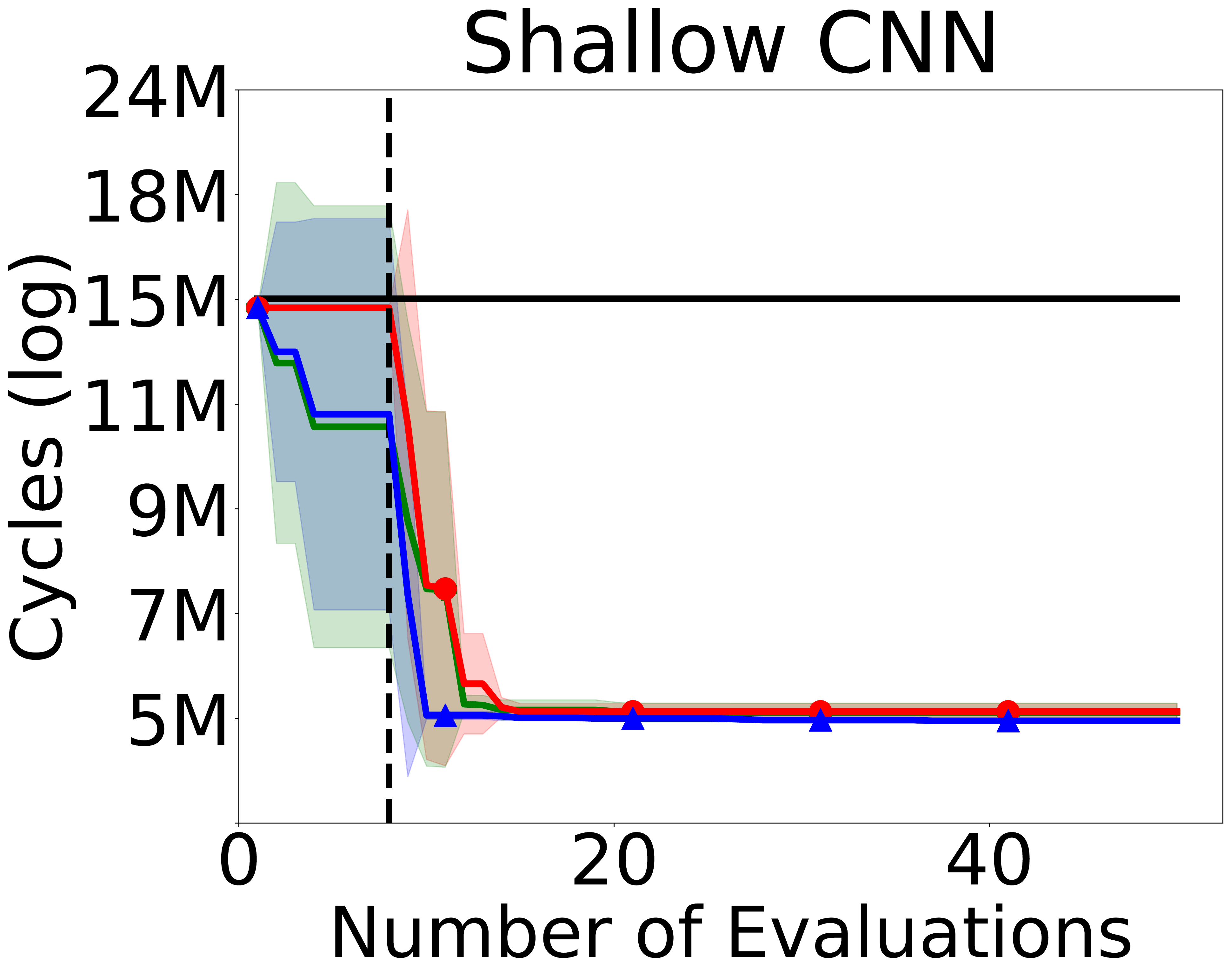}}
        \subfigure{\includegraphics[width=0.3\linewidth]{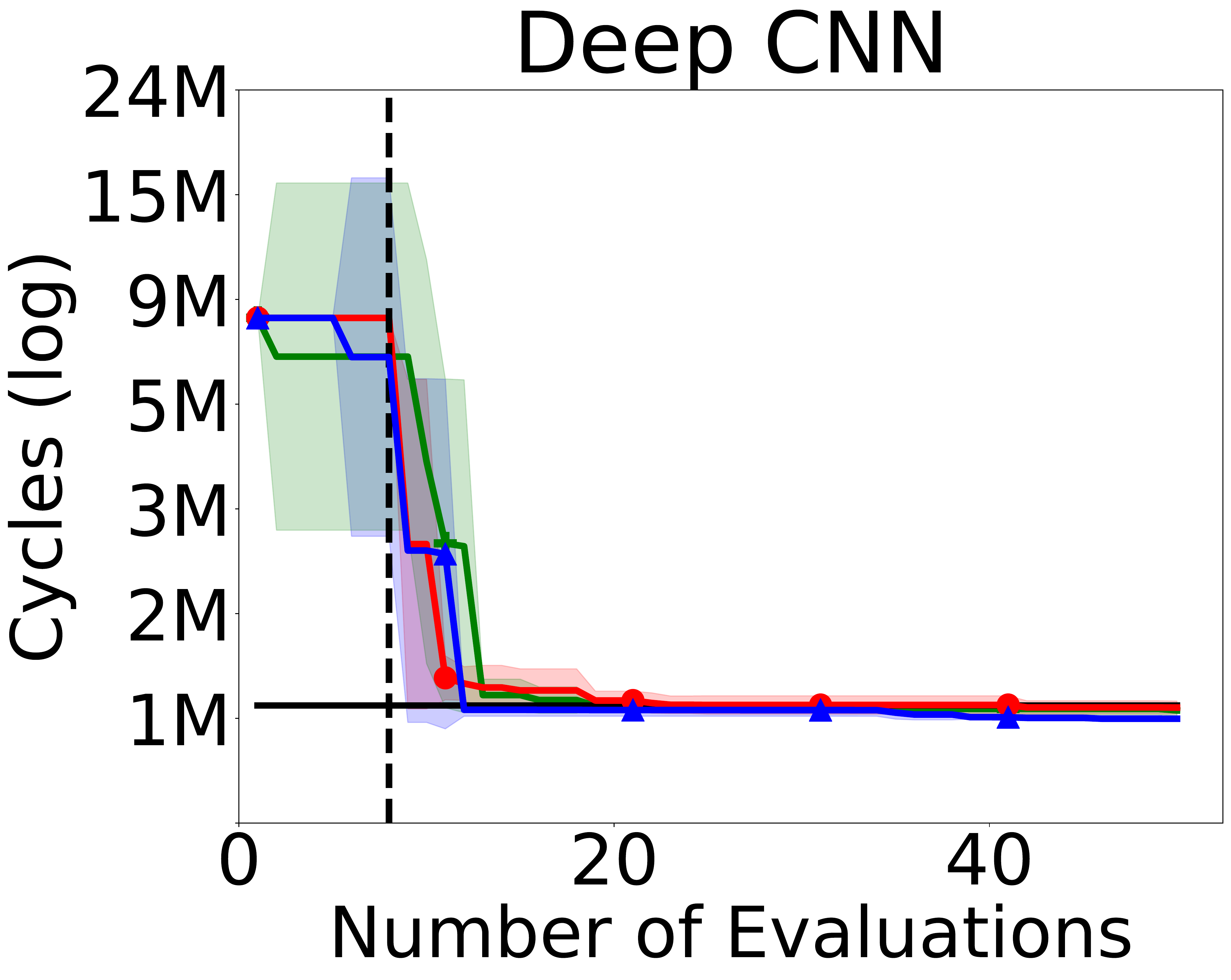}}
        \subfigure{\includegraphics[width=0.31\linewidth]{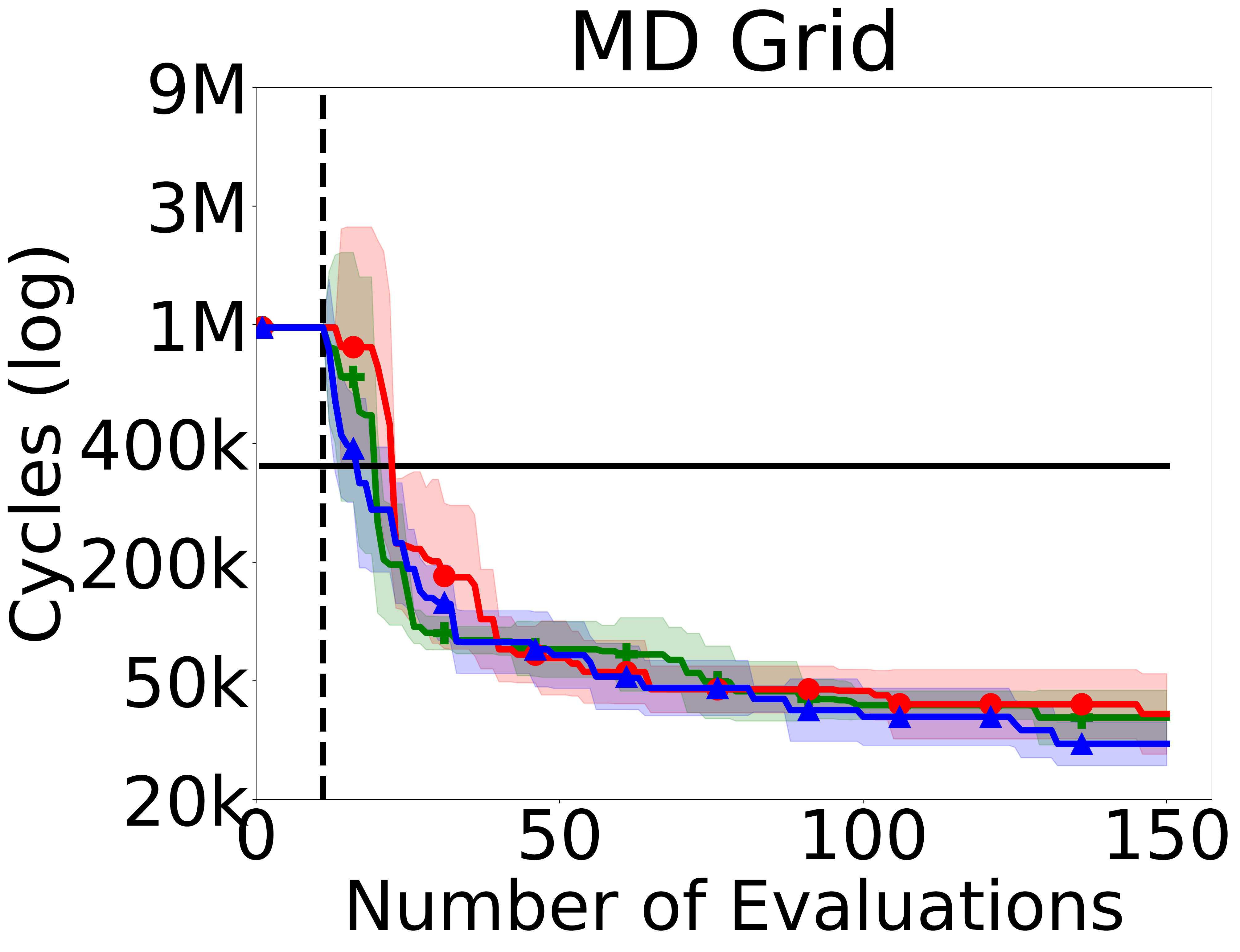}}
    \end{center}
    \caption{Log regret comparison of \name, HyperMapper with \priorstr initialization, and HyperMapper with default initialization. The line and shaded regions show the mean and standard deviation of the log simple regret after 5 runs. \name and HyperMapper \Priorstr were initialized with $D+1$ random samples from the \priorstr, where $D$ is the number of input dimensions, indicated by the vertical dashed line.}
    \label{fig:hm.prior}
\end{figure*}

We show that simply initializing a BO method in the DoE phase by sampling from a prior on the locality of an optimum doesn't necessarily lead to better performance. Instead in \name, it is the pseudo-posterior in Eq.~\eqref{eq:posterior} that drives its stronger performance by combining prior and new observations. To show that, we compare \name with Spearmint and HyperMapper such as in section Sections~\ref{sec:experiments.regret} and ~\ref{sec:experiments.spatial}, respectively, but we initialize all three methods using the same approach. Namely, we initialize all methods with $D+1$ samples from the \priorstr. Our goal is to show that simply initializing Spearmint and HyperMapper with the prior will not lead to the same performance as \name, because, unlike \name, these baselines do not leverage the prior after the DoE initialization phase. We report results on both our  synthetic and real-world benchmarks



Figure~\ref{fig:spearmint.prior} shows the comparison between \name and Spearmint \Priorstr. In most benchmarks, the prior initialization leads to similar final performance. In particular, for XGBoost, the prior leads to improvement in early iterations, but to worse final performance. We note that for FCNet, Spearmint Prior achieves better performance, however, we note that the improved performance is given almost solely from sampling from the \priorstr. There is no improvement for Spearmint \Priorstr{} until around iteration 190. In contrast, in all cases, \name is able to leverage the \priorstr both during initialization and its Bayesian Optimization phase, leading to improved performance. \name still achieves similar or better performance than Spearmint \Priorstr in all benchmarks.

Figure~\ref{fig:hm.prior} shows similar results for our Spatial benchmarks. The \priorstr does not lead HyperMapper to improved final performance. For the Shallow CNN benchmark, the \priorstr leads HyperMapper to improved performance in early iterations, compared to HyperMapper with default initialization, but HyperMapper \Priorstr is still outperformed by \name. Additionally, the \priorstr leads to degraded performance in the Deep CNN benchmark. These results confirm that \name is able to leverage the \priorstr in its pseudo-posterior during optimization, leading to improved performance in almost all benchmarks compared to state-of-the-art BO baselines.

\begin{figure*}[tb]
    \begin{center}
        \subfigure{\includegraphics[width=0.39\textwidth]{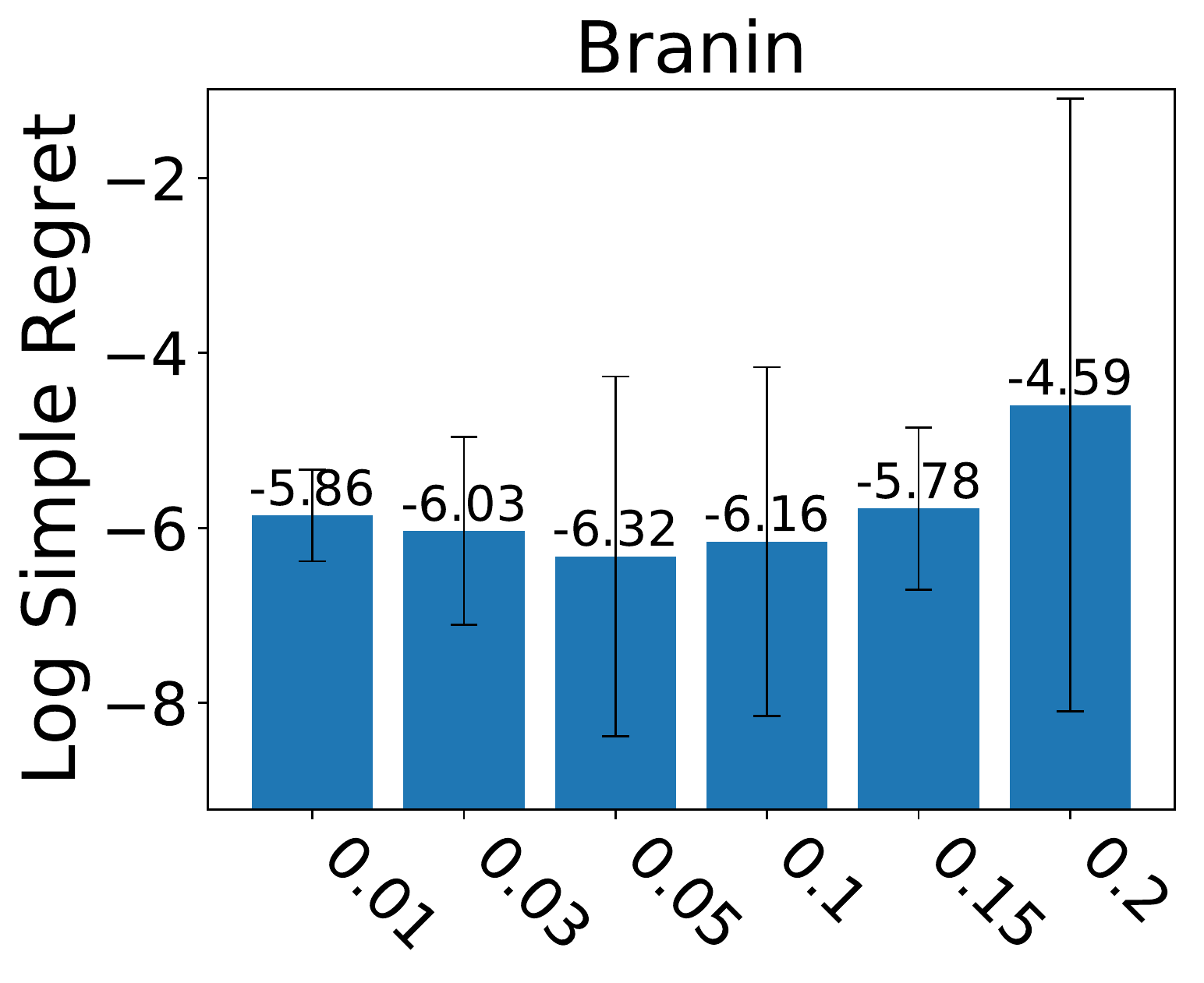}}
        \subfigure{\includegraphics[width=0.4\textwidth]{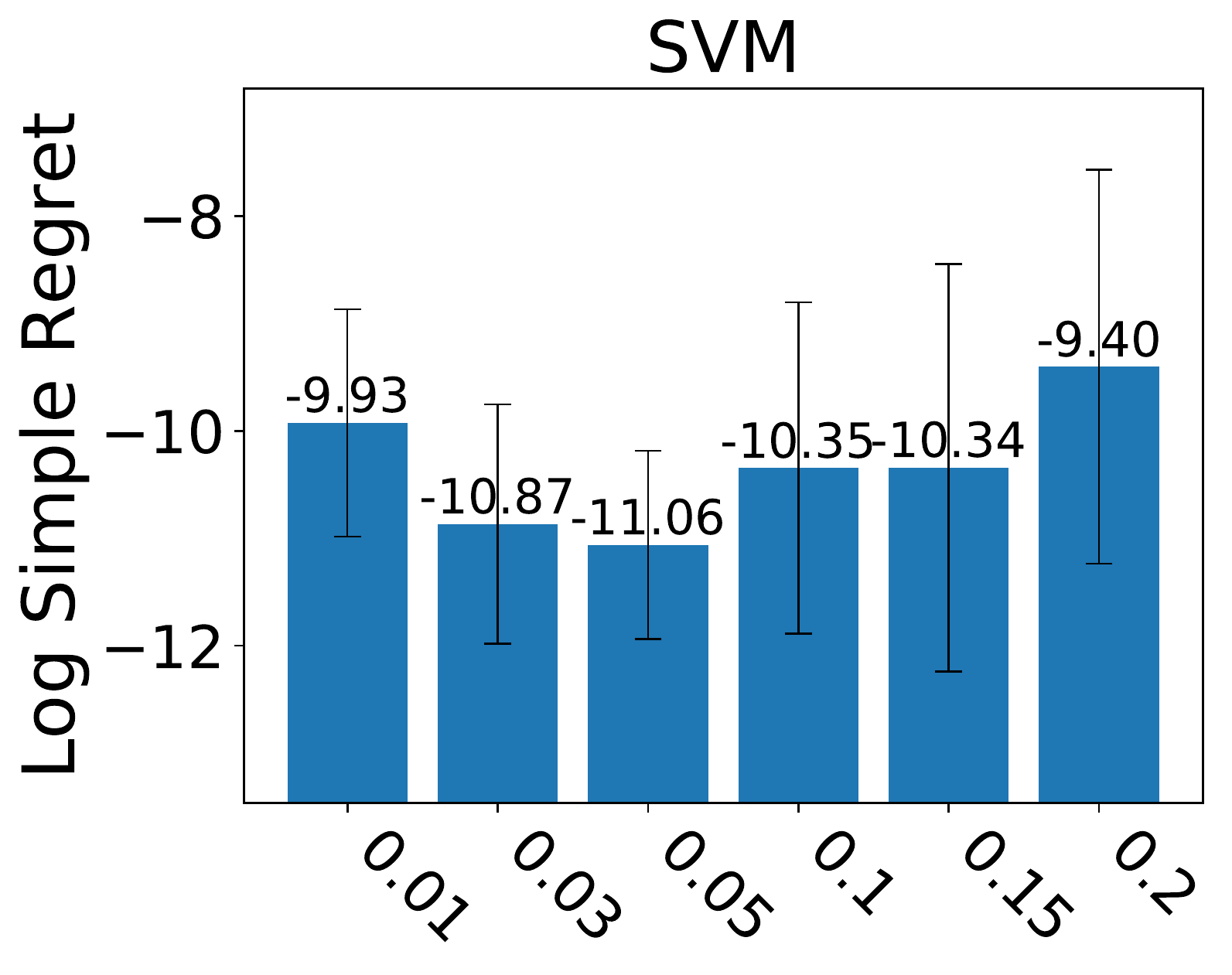}} \\
        \subfigure{\includegraphics[width=0.4\textwidth]{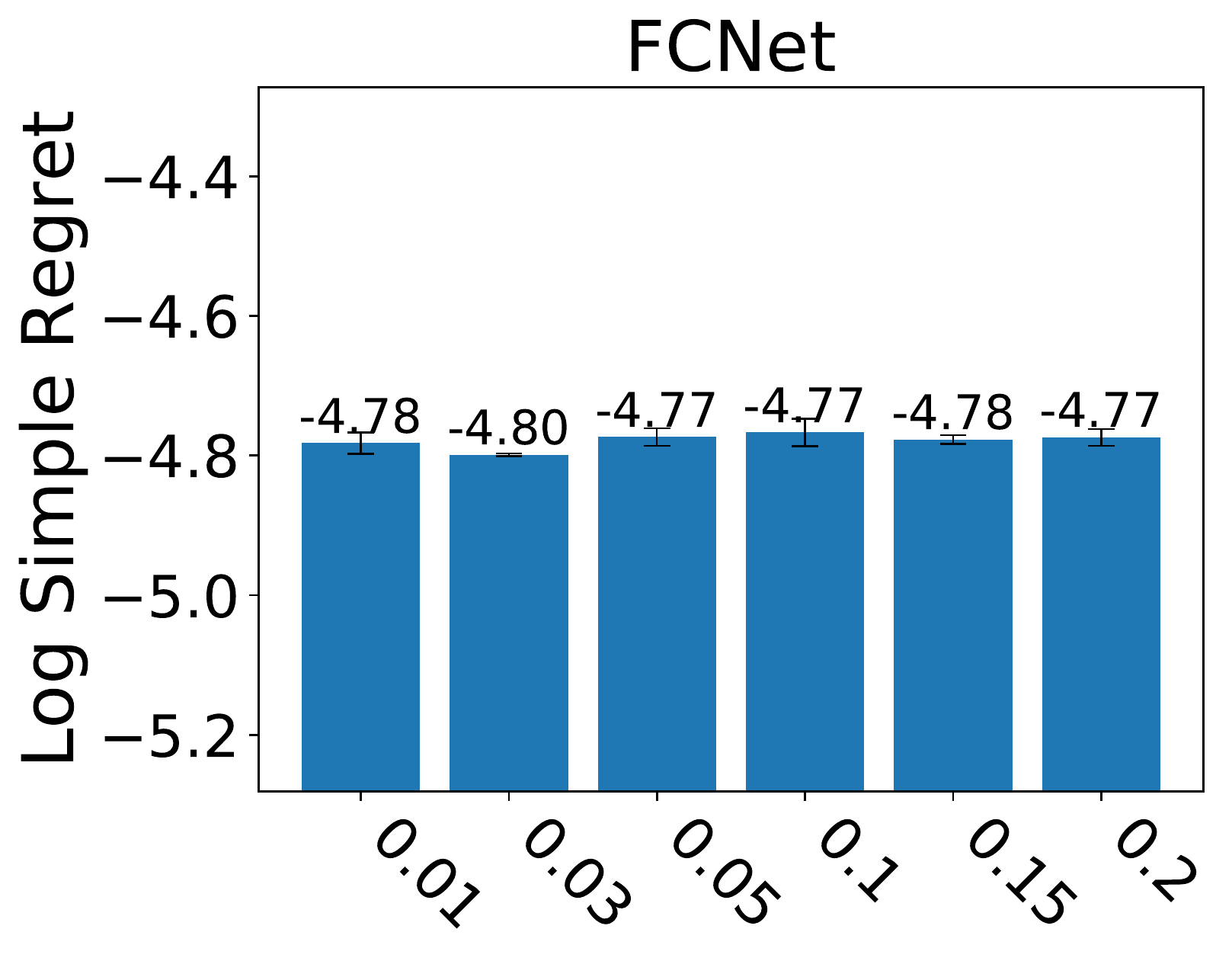}}
        \subfigure{\includegraphics[width=0.4\textwidth]{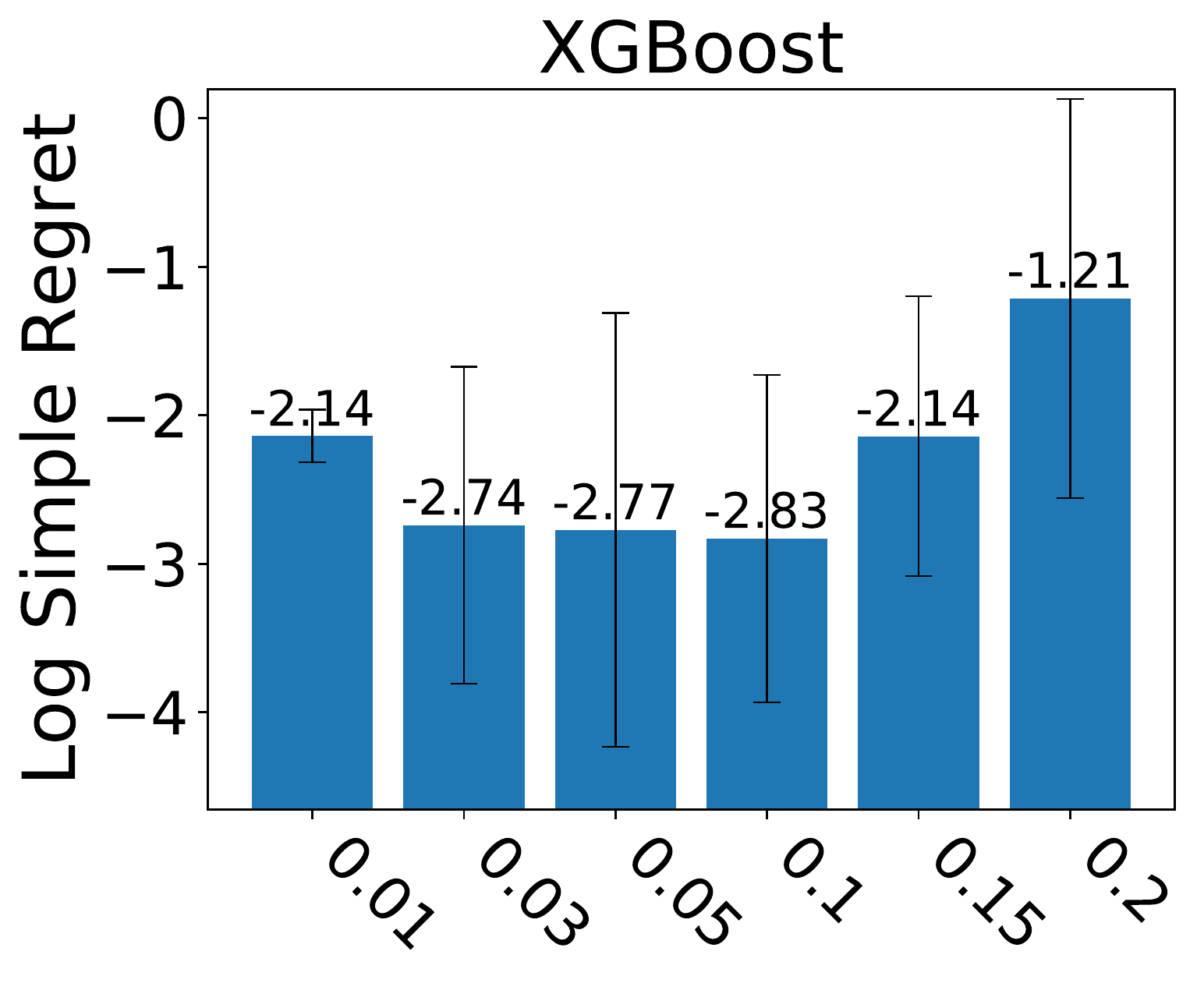}}
    \end{center}
    \caption{Comparison of \name with the strong \priorstr and different values for the $\gamma$ hyperparameter on our four synthetic  benchmarks. We run \name with a budget of $10D$ function evaluations, including $D+1$ randomly sampled DoE configurations.}
    \label{fig:gamma_comparison}
\end{figure*}

\section{$\gamma$-Sensitivity Study} \label{sec:gamma.appendix}

We show the effect of the $\gamma$ hyperparameter introduced in Section~\ref{sec:priorop.model} for the quantile identifying the points considered to be good. To show this, we compare the performance of \name with our strong \priorstr and different $\gamma$ values. For all experiments, we initialize \name with $D+1$ random samples and then run \name until it reaches $10D$ function evaluations. For each $\gamma$ value, we run \name five times and report mean and standard deviation.

Figure~\ref{fig:gamma_comparison} shows the results of our comparison. We first note that values near the lower and higher extremes lead to degraded performance, this is expected, since these values will lead to an excess of either exploitation or exploration. Further, we note that \name achieves similar performance for all values of $\gamma$, however, values around $\gamma = 0.05$ consistently lead to better performance. 

\section{$\beta$-Sensitivity Study} \label{sec:beta.appendix}

\begin{figure*}[tb]
    \begin{center}
        \subfigure{\includegraphics[width=0.4\textwidth]{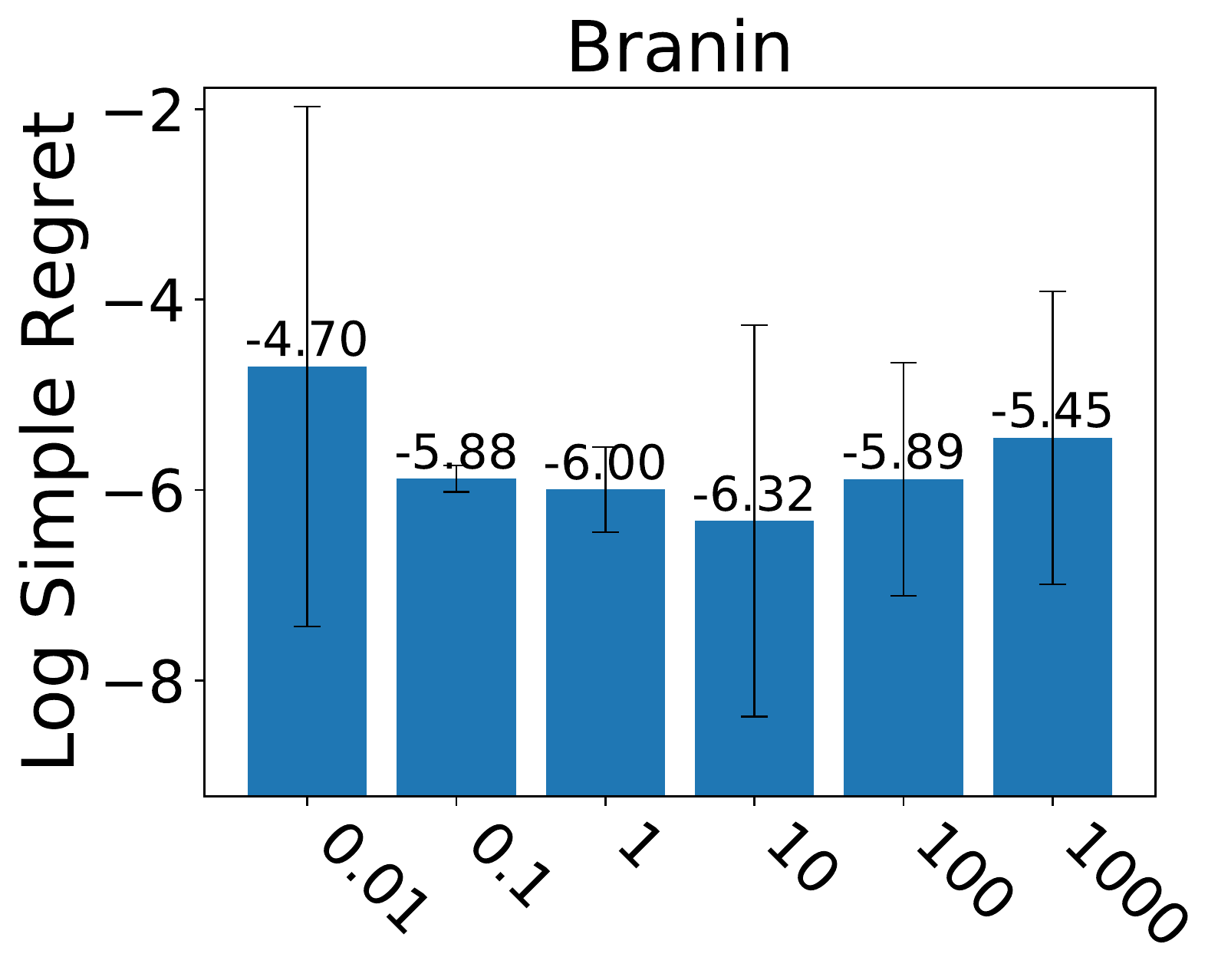}}
        \subfigure{\includegraphics[width=0.41\textwidth]{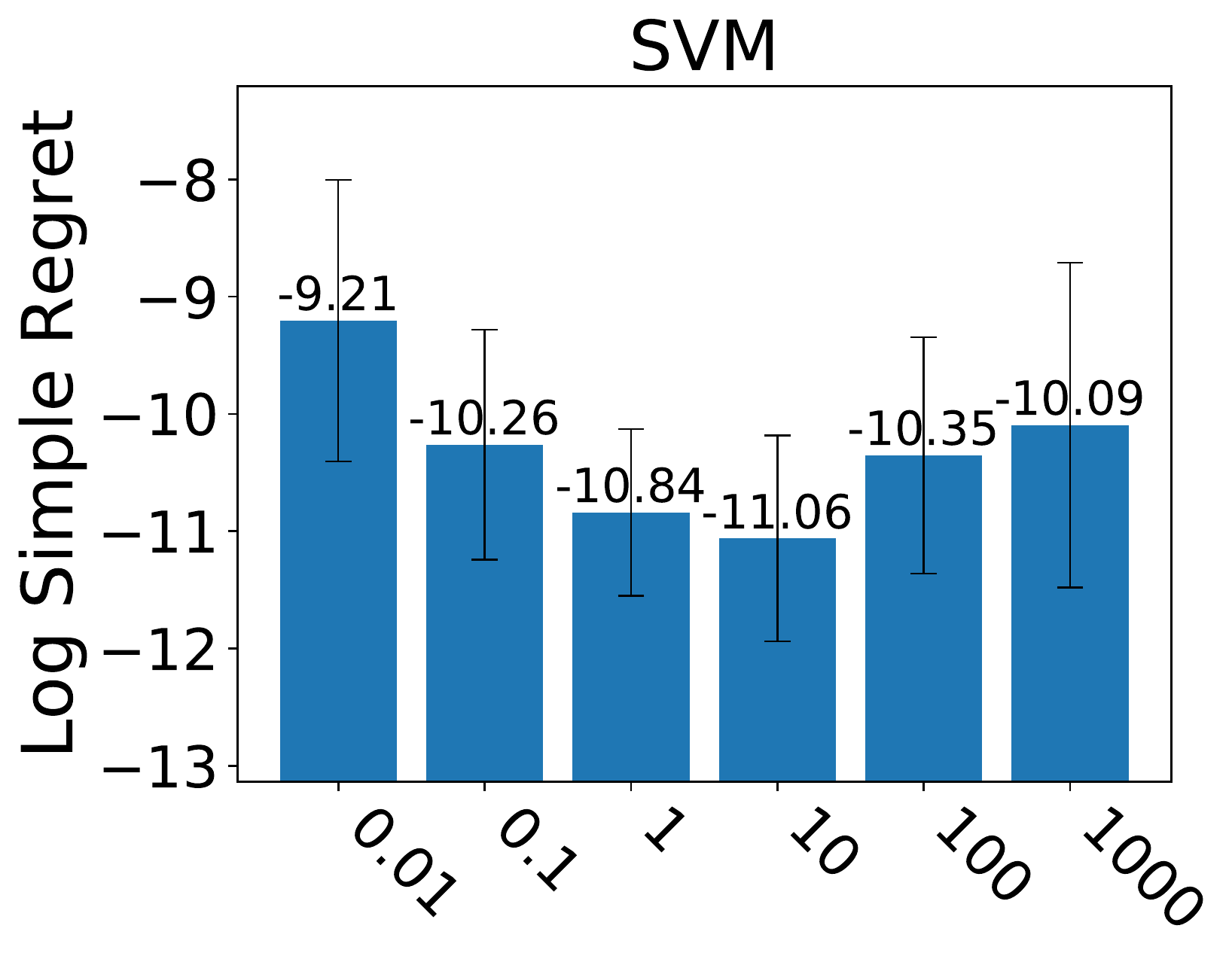}} \\
        \subfigure{\includegraphics[width=0.41\textwidth]{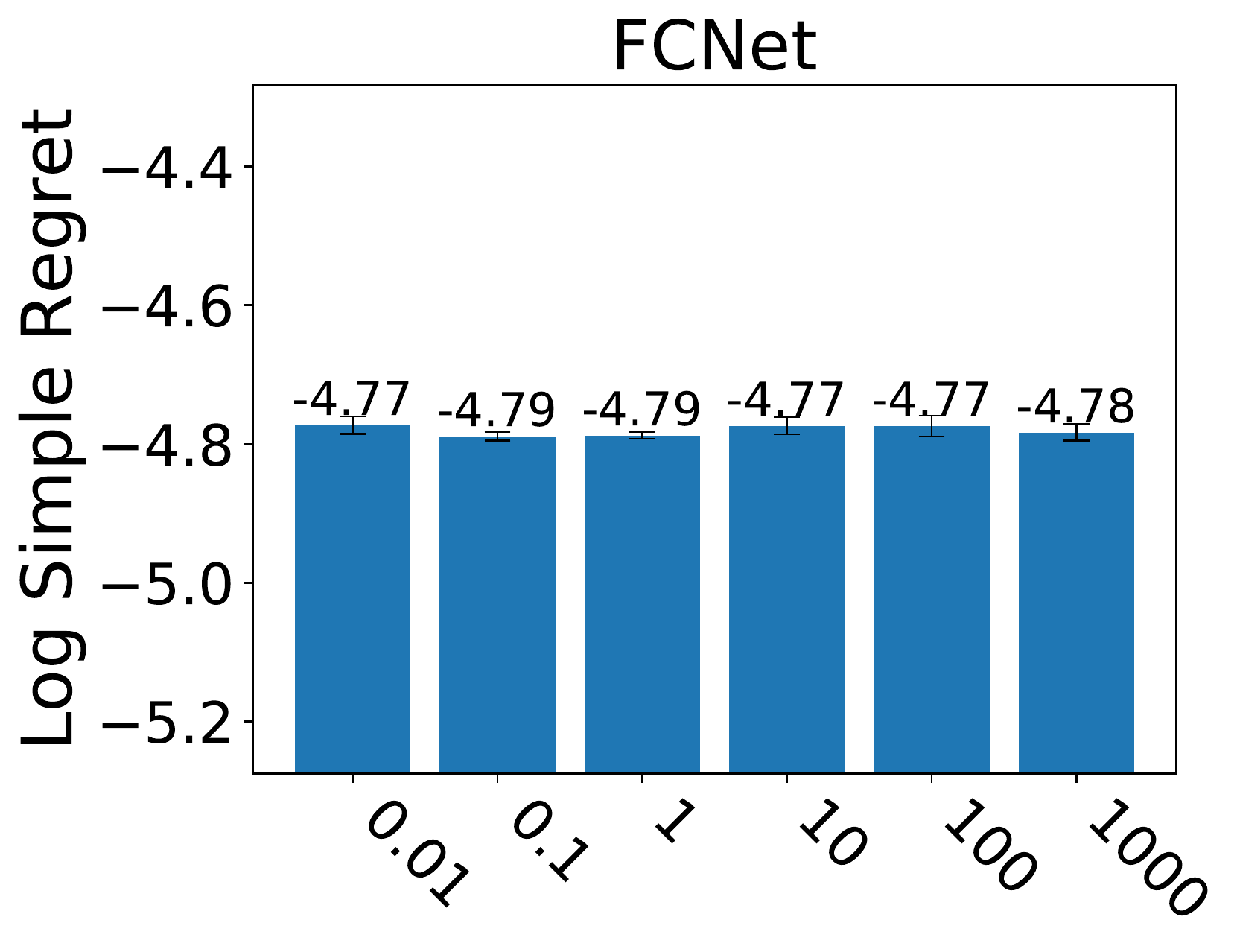}}
        \subfigure{\includegraphics[width=0.4\textwidth]{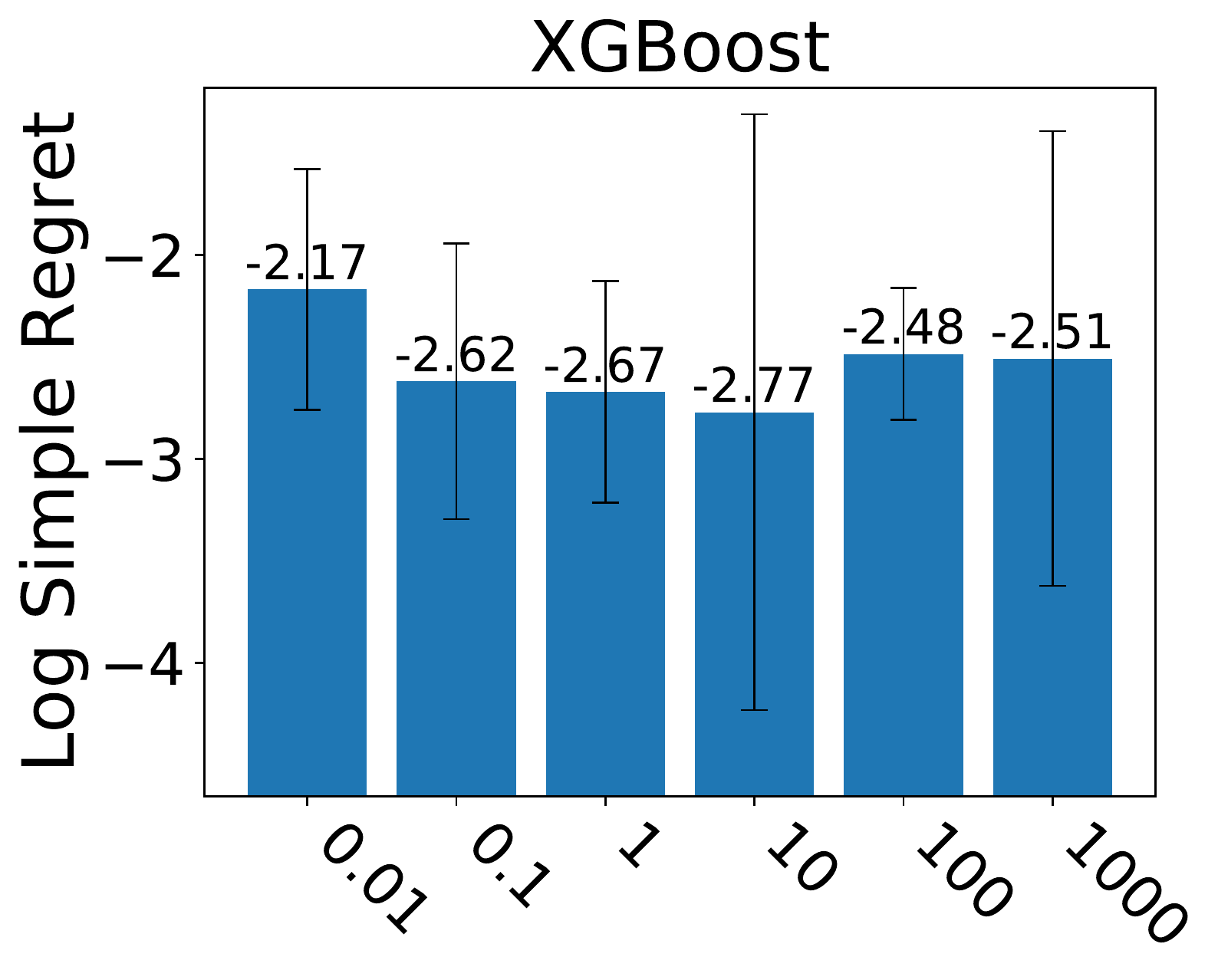}}
    \end{center}
    \caption{Comparison of \name with the strong \priorstr and different values for the $\beta$ hyperparameter on our four synthetic benchmarks. We run \name with a budget of $10D$ function evaluations, including $D+1$ randomly sampled DoE configurations.}
    \label{fig:beta_comparison}
\end{figure*}

We show the effect of the $\beta$ hyperparameter introduced in Section~\ref{sec:priorop.posterior} for controlling the influence of the \priorstr over time. To show the effects of $\beta$, we compare the performance of \name with our strong \priorstr and different $\beta$ values on our four synthetic benchmarks. For all experiments, we initialize \name with $D+1$ random samples and then run \name until it reaches $10D$ function evaluations. For each $\beta$ value, we run \name five times and report mean and standard deviation.

Figure~\ref{fig:beta_comparison} shows the results of our comparison. We note that values of $\beta$ that are too low (near $0.01$) or too high (near $1000$) often lead to lower performance. This shows that putting too much emphasis on the model or the \priorstr will lead to degraded performance, as expected.  Further, we note that $\beta = 10$ lead to the best performance in three out of our four benchmarks. This result is reasonable, as $\beta = 10$ means \name will put more emphasis on the \priorstr in early iterations, when the model is still not accurate, and slowly shift towards putting more emphasis on the model as the model sees more data and becomes more accurate.

\end{document}